\documentclass[10pt]{article} % For LaTeX2e
\usepackage[preprint]{tmlr}

\usepackage{hyperref}
\usepackage{cancel}
\usepackage{amsmath}
\usepackage{cleveref}
\usepackage{algorithmic}
\usepackage{stmaryrd}
\usepackage{booktabs}
\usepackage{array}
\usepackage{float}
\usepackage{mathtools}
\usepackage{tikz}

\newtheorem{theorem}{Theorem}
\newtheorem{lemma}[theorem]{Lemma} 
\newtheorem{proposition}[theorem]{Proposition} 

\newtheorem{corollary}[theorem]{Corollary}
\newtheorem{definition}[theorem]{Definition}

\newtheorem{assumption}[theorem]{Assumption}

\usepackage{stfloats}
\usepackage{caption}
\usepackage{subcaption}
\usepackage{verbatim}

\usepackage{shortcut}
\usepackage{wrapfig}
\usepackage[ruled, linesnumbered]{algorithm2e}

\usepackage{stix}
\usepackage{xspace}
\usepackage{pifont}% http://ctan.org/pkg/pifont
\usepackage{makecell}
\usepackage{nicefrac}
\usepackage{xcolor}

\usepackage{annotate-equations}

\newcommand{\stmethod}{\texttt{STMA}\xspace}
\newcommand{\smethod}{\texttt{SMA}\xspace}
\newcommand{\tmethod}{\texttt{TMA}\xspace}
\newcommand{\method}{\texttt{MA}\xspace}
\newcommand{\filtersize}{f}

\newcommand{\hermconj}{^{\mathsf{H}}}

\definecolor{Green}{RGB}{28,198,133}
\definecolor{Orange}{RGB}{250, 190, 81}
\definecolor{Blue}{RGB}{38, 133, 229}
\definecolor{Red}{RGB}{228, 62, 31}

\newcommand{\cmark}{\color{Green}\ding{51}\color{black}}%
\newcommand{\xmark}{\color{Red}\ding{55}\color{black}}%

\newcommand{\mcrot}[4]{\multicolumn{#1}{#2}{\rlap{\rotatebox{#3}{#4}~}}}

\author{\name Théo Gnassounou* \email theo.gnassounou@inria.fr \\
       \addr Université Paris-Saclay\\
        Inria, CEA\\
       Palaiseau, 91120, France
       \AND
       \name Antoine Collas* \email antoine.collas@inria.fr \\
       \addr Université Paris-Saclay\\
        Inria, CEA\\
       Palaiseau, 91120, France
       \AND
       \name Rémi Flamary \email remi.flamary@polytechnique.edu \\
       \addr CMAP, UMR 7641\\
       Institut Polytechnique de Paris\\
       Palaiseau, 91120, France
       \AND
       \name Karim Lounici \email karim.lounici@polytechnique.edu \\
       \addr CMAP, UMR 7641\\
       Institut Polytechnique de Paris\\
       Palaiseau, 91120, France
       \AND
       \name Alexandre Gramfort \email alexandre.gramfort@inria.fr \\
       \addr Université Paris-Saclay\\
        Inria, CEA\\
       Palaiseau, 91120, France}

\def\month{MM}  % Insert correct month for camera-ready version
\def\year{YYYY} % Insert correct year for camera-ready version
\def\openreview{\url{https://openreview.net/forum?id=XXXX}} % Insert correct link to OpenReview for camera-ready version

\begin{document}
\def\thefootnote{*}\footnotetext{Equal contribution}
    \title{Multi-Source and Test-Time Domain Adaptation on Multivariate Signals using Spatio-Temporal Monge Alignment}
% \author{IEEE Publication Technology Department
% \thanks{}}

\maketitle

\begin{abstract}
  % The usual machine learning methods have difficulty in generalizing results across subjects for tasks using biological signal data such as electroencephalogram (EEG). The reason is that the variability in the signals can be huge between subjects and trials. To overcome this problem, several pre-processing methods have been proposed, as well as specific subject normalizations trained on the data. The downside of these methods is that they do not allow generalization to new subjects without additional training. This is why we propose a new method of Convolutional Monge Mapping Normalization (\texttt{CMMN}), which proposes an smart normalization that does not require subject specific training and allows to have an inference without new training. We first prove that our method outperforms classical normalization and finetuned normalization methods. We then show that our method generalizes very easily to more complicated tasks such as the adaptation between different datasets. Finally, for these last problems that are usually solved by domain adaptation methods that minimize the divergence between embeddings, we present that our method allows to reach similar results for a much lower cost and allows to have even better results when we couple the two methods.

  {Machine learning applications on signals such as computer vision or biomedical
data often face significant challenges due to the variability that exists across
hardware devices or session recordings.
This variability poses a Domain Adaptation (DA) problem, as training and testing data distributions often differ. 
In this work, we propose Spatio-Temporal Monge Alignment (\stmethod{}) to mitigate these variabilities.
This Optimal Transport (OT) based method adapts the cross-power spectrum density (cross-PSD) of multivariate signals by mapping them to the Wasserstein barycenter of source domains (multi-source DA).
Predictions for new domains can be done with a filtering without the need for
retraining a model with source data (test-time DA).
We also study and discuss two special cases of the method, Temporal Monge Alignment (\tmethod{}) and Spatial Monge Alignment (\smethod{}).
Non-asymptotic concentration bounds are derived for the mappings estimation, which reveals a bias-plus-variance error structure with a variance decay rate of $\mathcal{O}(n_\ell^{-1/2})$ with $n_\ell$ the signal length.
This theoretical guarantee demonstrates the efficiency of the proposed computational schema.
Numerical experiments on multivariate biosignals and image data show that \stmethod{} leads to significant and consistent performance gains between datasets acquired with very different settings.
Notably, \stmethod{} is a pre-processing step complementary to state-of-the-art deep learning methods. }

\end{abstract}

\section{Introduction}

Machine learning approaches have led to impressive results across many applications, from computer vision and biology to audio and language processing. However, these approaches have known limitations in the presence of distribution shifts between training and evaluation datasets.
Indeed, performance drops in the presence of distribution shifts have been observed in different fields such as computer vision \cite{ganin2016domainadversarial},
clinical data \cite{Harutyunyan_2019}, and tabular data
\cite{gardner2024benchmarking}.

\sloppy
\paragraph{Domain Adaptation (DA) from single to multi-domain}
This problem of data shift is well-known in machine learning and has been investigated in the
DA community \cite{sugiyama07a,bendavid2007da, quinonero2008dataset,
redko2022survey}.
Given source domain data with access to labels, DA methods aim to learn a model that can adequately predict on a target domain where no label is available, assuming the existence of a distribution shift between the two domains. 
Traditionally, DA methods try to adapt the source to be closer to the target using reweighting methods \cite{sugiyama07a, shimodaira_improving_2000}, mapping estimation \cite{sun_correlation_2016, courty_optimal_2016} or dimension reduction \cite{pan2011tca}. 
Inspired by the successes of deep learning in computer vision, modern methods aim to reduce the shift between the embeddings of domains learned by a feature extractor.
To achieve this, most methods attempt to minimize a divergence between the features of the source and target data. 
Several divergences have been proposed in the literature:
correlation distance \cite{sun_deep_2016}, adversarial method \cite{ganin2016domainadversarial}, maximum mean discrepancy distance \cite{long2015learning} or
Optimal Transport (OT) \cite{shen2018wasserstein,damodaran2018deepjdot}.

With the rise of portable devices and open data, the problem of DA has evolved from a single-source setting where only one source is known to a multi-source setting where each source domain can have a different shift.
This leads to a multi-source DA problem
\cite{sun_survey_nodate}, where the variability across multiple domains can help training better predictors.
For instance, one can learn domain-specific batch
normalization \cite{li2016revisiting,kobler-etal:22}, compute weights to give more importance
to some domains \cite{turrisi_multi-source_2022}, or use moment matching to
align the features \cite{peng_moment_2019}.
Another method to deal with multiple sources is to create an intermediate domain between sources and target. \cite{montesuma2021wasserstein} propose to compute a Wasserstein barycenter of the sources and then project this barycenter to the target using classical OT methods.

An additional challenge with traditional DA methods \cite{damodaran2018deepjdot, ganin2016domainadversarial, sun_deep_2016} is that they require using both the source and target domains to train the predictor.
This means one needs access to the source data to train a model for new domains.
However, the source data may not always be available due to privacy concerns or memory limitations.
Test-time DA is a branch of DA that aims to adapt a predictor to the target data without access to the source data.
In \cite{liang2021really}, the authors propose SHOT, which trains the classifier on source data and matches the target features to the fixed classifier using Information Maximization (IM). IM aims to produce predictions that are individually confident and globally diverse. \cite{ahmed2021unsupervised} enhance SHOT with multi-source information, allowing the selection of the optimal combination of sources by learning the weights of each source model.
\cite{yang_generalized_2021} uses the local structure clustering technique on the target feature to adapt the model. 
Note that these methods are often complex and require high computational resources since they require training a new estimator for each new target domain.

\paragraph{DA for multivariate signals and biomedical data} 
This paper focuses on multi-source and test-time DA for multivariate signals, which are common in applications where data is collected from multiple devices.
For instance, in the biomedical field, signals such as Electroencephalogram (EEG), Electrocardiogram (ECG), or Electromyogram (EMG) are often multivariate and can exhibit a shift between subjects, sessions, or devices.
Several applications of DA in biosignals have been proposed for pairs of datasets \cite{Jeon2019dabci, eldele_adast_2022}.
However, in clinical datasets, we often have access to multiple subjects or hospital data, which can be considered as separate domains. 
In this paper, we are interested in two biosignal applications with their own specificities: sleep staging and Brain-Computer Interface (BCI).
Sleep staging is the process of categorizing sleep into five stages based on the electrical activity of the brain, \ie EEG \cite{STEVENS200445}. 
BCI motor imagery classifies mental visualization of motor movements using EEG signals.
In both applications, datasets can exhibit diverse population distributions because of variations in factors such as age, gender, diseases, or individual physiological responses to stimuli \cite{zhang2018NSRR}. 
All these variabilities introduce data shifts between different subjects or datasets (domains) that limit performances of traditional supervised learning methods \cite{wimpff2024calibrationfree, eldele_adast_2022}.
% The above-mentioned often suggest reducing shifts during neural network training through custom loss and architecture. However, implementing these methods can be complex and requires high computational resources, which may not be feasible for hospitals and specialists. Moreover, these methods lack interpretability, which is essential for biological applications.

To take into account those shifts, or more precisely to try to mitigate them, simple normalization techniques have proven useful for BCI and sleep staging, leading to
easy computation and application at test-time.
For instance, researchers proposed Riemannian Procrustes Analysis (RPA) \cite{Rodriguez2019RPA} to reduce
the shift in BCI.
RPA matches data distributions of source and target domains
through recentering, stretching, and rotation, while operating on a Riemanian manifold of covariance matrices. 
When using deep learning, people typically only use the recentering part proposed in RPA and recenter the signal using Euclidean alignment \cite{he_transfer_2019, junqueira2024systematic} or Riemannian Alignment (RA) \cite{xu2020bcicross, wimpff2024calibrationfree}. 
Alternatively, in sleep staging, a study by \cite{gnassounou2023convolutional} utilized a convolutional Monge mapping normalization method to align each signal to a Wasserstein barycenter. 
This approach improved sleep staging significantly, compared to classical normalization techniques, such as the unit standard deviation scaling \cite{chambon2018deep, apicella2023effects}.

\begin{figure}[t]
    \centering
    \includegraphics[width=0.9\textwidth]{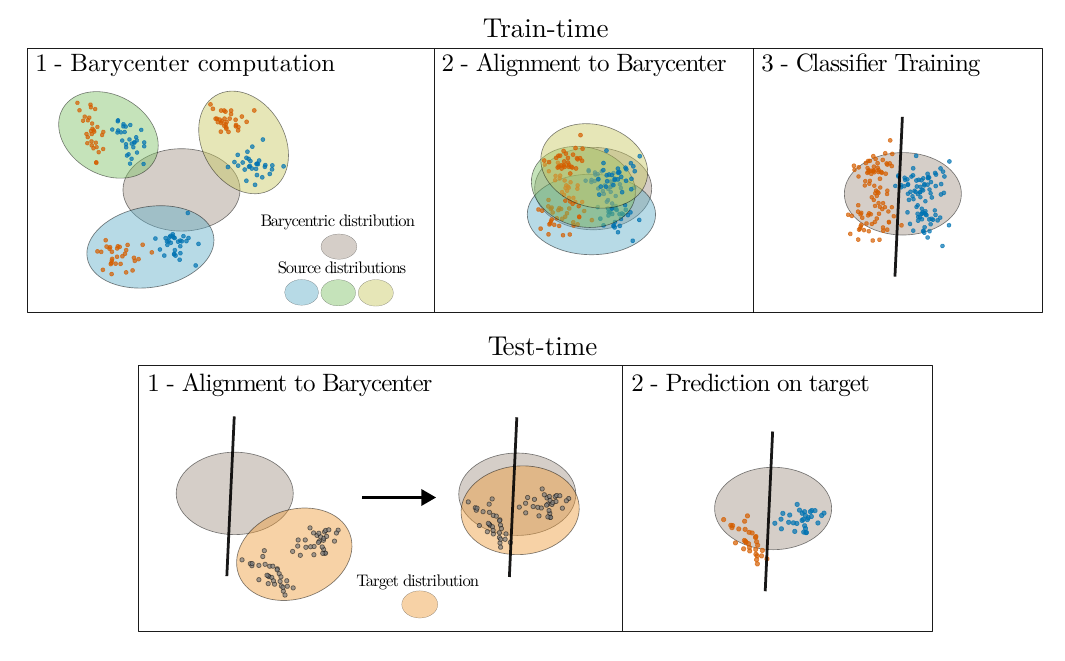}
    \caption{Illustration of Monge Alignment. At train-time the barycenter (grey ellipse) is estimated
from 3 source distributions (colored ellipses). The predictor is learned on normalized data. At test-time the same barycenter
is used to align the unlabeled target distribution (orange ellipse) and predict.}
    \label{fig:concept_figure}
\end{figure}

\paragraph{Contributions}
This paper proposes a general method for multi-source and test-time
DA for multivariate signals.
% This method, denoted as Monge Alignment (\method{}) and illustrated in \autoref{fig:concept_figure}, uses Optimal Transport (OT) modeling of stationary signals to estimate a mapping that aligns the different domains onto a barycenter of the domain before learning a predictor on the aligned data where the shifts have been attenuated (\autoref{fig:concept_figure} top).
{
This method denoted as Monge Alignment (\method{}), uses OT modeling of stationary signals to estimate a mapping.
The latter aligns the different domains onto a barycenter of the domains before
learning a predictor, attenuating the shift (\autoref{fig:concept_figure} top).
}
At test time, one only needs
to align the target domain on the barycenter and predict with the trained model
(\autoref{fig:concept_figure} bottom).
This method can be seen as a special case of \cite{montesuma2021wasserstein}, where the authors propose to align
signals using the Wasserstein barycenter, but where the modeling of multivariate signals allows for more efficient barycenter computation and mapping estimation.
It is also a generalization of the preliminary work of \cite{gnassounou2023convolutional} where a Convolutional Monge
Mapping Normalization was proposed to deal very efficiently with temporal shifts, but the different signals in the observations are assumed to be independent.
The novel contributions of this paper are:
\begin{itemize}
    \item A general method for multi-source and test-time DA for multivariate
    signals, using OT between Gaussian signals, denoted as Spatio-Temporal Monge
    Alignment (\stmethod{}) {as well as Temporal and Spatial Monge Alignments
    (\tmethod{} and \smethod{} respectively) tailored for specific shifts assumptions.}
    \item Efficient algorithms for estimations of the mapping and the barycenter of the signals using Fast Fourier Transform (FFT) and block-diagonalization of the covariances matrices.
    \item {Non-asymptotic concentration bounds of \stmethod{}, \tmethod{} and \smethod{} estimation} which present the interactions between the number of samples, the number of domains, the dimensionality of the signals, and the size of the filters.
    \item Novel numerical experiments on two biosignal tasks, Sleep staging, and BCI motor imagery, that show the benefits of the proposed method over state-of-the-art methods and illustrate the relative importance of spatial and temporal filtering.
\end{itemize}
In addition to the contributions discussed above, the proposed multi-source method is predictor-agnostic.
It can be used for shallow or deep learning methods, as it acts as a pre-processing step.
The estimated parameters can be very efficiently
computed for new domains without the need for the source data or refitting of the learned model at test-time. \autoref{tab:comparision_method} summarizes the specificities of the proposed method compared to other existing DA methods.

\paragraph{Paper outline}
After quickly recalling OT between Gaussian distributions, we
introduce the spatio-temporal Monge mapping for Gaussian multivariate signals in \autoref{sec:math}.
In \autoref{sec:CMMN}, we detail the Spatio-Temporal Monge Alignment (\stmethod{}) method for multi-source and test-time DA and its special cases using only temporal (\tmethod{}) or spatial (\smethod{}) alignments.
{This section also provides non-asymptotic concentration bounds  which reveals a bias-plus-variance error structure.}
% This section provides the first statistical results regarding estimating those OT barycenters and mappings.
In \autoref{sec:exp}, we apply \method{} to two real-life biosignal tasks using
EEG data: sleep stage classification and BCI motor imagery and provide an illustrative example on images (2D signals).
\begin{table}[t]
    \centering
    \footnotesize
    % \scalebox{1}{
    % \begin{tabular}{|c|c|c|c|c|c|c|}
    % \hline
    %    \textbf{Methods}  & \textbf{Multi-source} & \textbf{Test-time }& \makecell{\textbf{No need} \\ \textbf{to refit}} & \makecell{\textbf{Method} \\ \textbf{agnostic}} & \makecell{\textbf{Temporal} \\ \textbf{shift}} & \makecell{\textbf{Spatial} \\ \textbf{shift}} \\
    % \hline\hline
    %     \makecell{\textbf{Divergence minimization} \\ \cite{damodaran2018deepjdot}} & \xmark & \xmark & \xmark & \xmark & \cmark & \cmark \\
    %     \makecell{\textbf{Moment Matching} \\ \cite{peng_moment_2019}} & \cmark & \xmark & \xmark & \xmark & \cmark & \cmark \\
    %     \makecell{\textbf{SHOT} \\ \cite{liang2021really}} & \xmark & \cmark & \xmark & \xmark & \cmark & \cmark \\
    %     \makecell{\textbf{DECISION} \\ \cite{ahmed2021unsupervised}}& \cmark & \cmark & \xmark & \xmark & \cmark & \cmark \\
    %     \makecell{\textbf{AdaBN} \\ \cite{li2016revisiting}} & \cmark & \cmark & \cmark & \xmark & \cmark & \cmark \\
    %     \makecell{\textbf{Riemanian Alignment} \\ \cite{wimpff2024calibrationfree}} & \cmark & \cmark & \cmark & \cmark & \xmark & \cmark \\ \hline
    %     \makecell{\textbf{Temporal \method{}} \\ \cite{gnassounou2023convolutional}} & \cmark & \cmark & \cmark & \cmark & \cmark & \xmark \\
    %     \makecell{\textbf{Spatial \method{}} \\ (ours) } & \cmark & \cmark & \cmark & \cmark & \xmark & \cmark \\
    %    \makecell{\textbf{Spatio-Temp \method{}} \\ (ours)} & \cmark & \cmark & \cmark & \cmark & \cmark & \cmark \\
    % \hline
    % \end{tabular}
    % }

% \vspace{-5mm}
       \scalebox{1}{
    \begin{tabular}{|l|c|c|c|c|c|c|}
  %  \hline
    \mcrot{1}{l}{0}{\textbf{Methods}} 
   
         & %\rot{90}{\makecell{\textbf{Multi}  \\ source}} 
       \mcrot{1}{l}{60}{\textbf{Multi-source}}
       & %\rot{90}{\makecell{\textbf{Source}  \\free }}
       \mcrot{1}{l}{60}{\textbf{Test-time}}
       & %\rot{90}{\makecell{\textbf{No need} \\ \textbf{to refit}}} 
       \mcrot{1}{l}{60}{\textbf{No need to refit}}
       & %\rot{90}{\makecell{\textbf{Method} \\ \textbf{agnostic}} }
        \mcrot{1}{l}{60}{\textbf{Method agnostic}}
       & %\rot{90}{\makecell{\textbf{Temporal} \\ \textbf{shift}}} 
       \mcrot{1}{l}{60}{\textbf{Temporal shift}}
       & %\rot{90}{\makecell{\textbf{Spatial} \\ \textbf{shift}}} 
       \mcrot{1}{l}{60}{\textbf{Spatial shift}} \\
    \hline
        {\textbf{Divergence minimization} \cite{sun_deep_2016}} & \xmark & \xmark & \xmark & \xmark & \cmark & \cmark \\
        {\textbf{Moment Matching}  \cite{peng_moment_2019}} & \cmark & \xmark & \xmark & \xmark & \cmark & \cmark \\
        {\textbf{SHOT}  \cite{liang2021really}} & \xmark & \cmark & \xmark & \xmark & \cmark & \cmark \\
        {\textbf{DECISION} \cite{ahmed2021unsupervised}}& \cmark & \cmark & \xmark & \xmark & \cmark & \cmark \\
        {\textbf{AdaBN}  \cite{li2016revisiting}} & \cmark & \cmark & \cmark & \xmark & \cmark & \cmark \\
        {\textbf{Riemanian Alignment} \cite{wimpff2024calibrationfree}} & \cmark & \cmark & \cmark & \cmark & \xmark & \cmark \\ \hline
        {\textbf{Temporal \method{}}  \cite{gnassounou2023convolutional}} & \cmark & \cmark & \cmark & \cmark & \cmark & \xmark \\
        {\textbf{Spatial \method{}}  (ours) } & \cmark & \cmark & \cmark & \cmark & \xmark & \cmark \\
       {\textbf{Spatio-Temp \method{}}  (ours)} & \cmark & \cmark & \cmark & \cmark & \cmark & \cmark \\
    \hline
    \end{tabular}
    }
    \caption{Comparison of DA methods with various specificities.
    ``Multi-source indicates the ability to handle multiple sources,
    ``Test-time'' means that the method does not require source data at test-time,
    ``No need to refit'' signifies whether refitting is required for a new domain at test-time,
    ``Method agnostic'' refers to independence from a specific type of predictors (\eg deep learning),
    while ``Temporal shift'' and ``Spatial shift'' indicate the capability to handle temporal and spatial shifts. Only \stmethod{} handles all the specificities.}
    \label{tab:comparision_method}
\end{table}

\paragraph*{Notations}
\noindent Vectors are denoted by small cap boldface letters (e.g., $\bx$), matrices by large cap boldface letters (e.g., $\bX$). The element-wise product is $\odot$, and the element-wise power of $n$ is $\cdot^{\odot n}$. $\intset{K}$ denotes $\{1, \ldots, K\}$. 
The absolute value is $|.|$.
The discrete convolution operator is $*$. 
{$\cS_{n}$ and $\cS_{n}^{++}$ denote the sets of symmetric and symmetric positive definite matrices of size $n \times n$.
$\cH_{n}$ and $\cH_{n}^{++}$ denote the sets of hermitian and hermitian positive definite matrices of size $n \times n$.
$\cC_{n}$ is the set of real-valued circulant matrices of size $n \times n$.}
{
We denote by $\lambda_{\max}(\bA)$ the maximum eigenvalue of the matrix $\bA$. We also define the effective rank by $\br(\bA) = \frac{\text{tr}(\bA)}{\lambda_{\max}(\bA)}$ where $\text{tr}(\bA)$ is the trace of $\bA$.
}
$\bbN^*$ is the integer set excluding $0$, and $\bbR^{+*}$ is the set of strictly positive real values. $\bX_k$ and $\bX_t$ relate to source domains $k \in \intset{n_d}$ and the target domain, respectively. $\Vectr: \bbR^{n_c \times n_\ell} \to \bbR^{n_c n_\ell}$ concatenates rows of a time series into a vector, and $\Vectr^{-1}: \bbR^{n_c n_\ell} \to \bbR^{n_c \times n_\ell}$ is the reciprocal. $\ceil{.}$ and $\floor{.}$ denote the smallest integer greater than or equal to, and the greatest integer less than or equal to, respectively. $(\bx)_{l}$ refers to the $l^\text{th}$ element of $\bx$, and $(\bX)_{lm}$ denotes the element of $\bX$ at the $l^\text{th}$ row and $m^\text{th}$ column. 
% $\bX^*$ is the conjugate of $\bX$ and
$\bX\hermconj$ is the conjugate transpose of $\bX$. $\otimes$ is the Kronecker operator.

\section{Signal adaptation with Optimal Transport}
\label{sec:OT}
In this section, we first briefly introduce OT
between centered Gaussian distributions. Then, we describe how \cite{gnassounou2023convolutional} has used this mapping between distributions to align univariate signals on a barycenter.
Finally, we introduce approximations on circulant matrices, which explain how the structure of covariance matrices is efficiently used.

\subsection{Optimal Transport between Gaussian distributions}

\subsubsection{Monge mapping for Gaussian distributions} Let two Gaussian
distributions $\mu_s = \mathcal{N}(\mathbf{0}, \bSigma_s)$ and $\mu_t = \mathcal{N}(\mathbf{0}, \bSigma_t)$, where $\bSigma_s$ and $\bSigma_t$ are symmetric positive definite matrices.
OT between Gaussian distributions is one of the rare cases where a closed-form solution exists.
% The OT cost, also called the Bures-Wasserstein distance when using a quadratic ground metric, is

%
% \begin{equation}
%     \mathcal{W}^2_2(\mu_s, \mu_t) = \text{Tr}\left(\bSigma_s + \bSigma_t - 2 \left(\bSigma_t^{\frac{1}{2}} \bSigma_s \bSigma_t^{\frac{1}{2}}\right)^{\frac{1}{2}}\right) \; .
%     \label{eq:bures_wass}
% \end{equation}
% \begin{equation}
%     \mathcal{W}^2_2(\mu_s, \mu_t) = \Vert \bm_s - \bm_t \Vert_2^2 + \text{Tr}\left(\bSigma_s + \bSigma_t - 2 \left(\bSigma_t^{\frac{1}{2}} \bSigma_s \bSigma_t^{\frac{1}{2}}\right)^{\frac{1}{2}}\right) \; ,
%     \label{eq:bures_wass}
% \end{equation}
%
% and is also called the Bures metric \cite{Forrester_2015} between positive definite matrices.
The OT mapping, also called  Monge mapping, is the following affine function \cite{Forrester_2015, bhatia2019bures,peyre_computational_2020}:
\begin{equation}
    m(\bx) = \bA \bx \quad \text{with} \quad  \bA = \bSigma_s^{-\frac{1}{2}}\left( \bSigma_s^{\frac{1}{2}} \bSigma_t\bSigma_s^{\frac{1}{2}} \right)^{\frac{1}{2}}  \bSigma_s^{-\frac{1}{2}} = \bA^\trans\;.
    \label{eq:Monge_map}
\end{equation}
In practical applications, one can estimate the covariance matrices of the two distributions to get an estimation of the Wasserstein distance and its associated mapping \cite{flamary2020concentration}.
% Interestingly, in this case, the concentration of the estimators is in $O(N^{-1/2})$, where $N$ is the number of observations, for the divergence estimation  (\cite{nadjahi2021fast}) and for the mapping estimation  (\cite{flamary2020concentration}). This is particularly interesting because OT in the general case is known to be very sensitive to the curse of dimensionality with usual concentrations in $O(N^{-1/D})$ where $D$ is the dimensionality of the data (\cite{fournier2015rate}).

\subsubsection{Wasserstein barycenter between Gaussian distributions}
\label{sec:bary}
The Wasserstein barycenter $\overline\mu$ {defines a notion of averaging of probability distributions $\{\mu_k\}_{1\leq k \leq n_d}$ which is the solution of a convex program involving Wasserstein distances:}
\begin{equation}
    \overline\mu  \triangleq \underset{\mu}{\text{arg min}} \; \frac{1}{n_d} \sum_{k=1}^{n_d} \mathcal{W}_2^2(\mu, \mu_k) \; .
    \label{eq:bw_bary}
\end{equation}
Interestingly, when $\mu_k$ are Gaussian distributions, the barycenter is still a Gaussian distribution $\overline\mu =
\mathcal{N}(\mathbf{0}, \overline{\bSigma})$ \cite{agueh2011bary}. 
% Its mean$\overline\bm=\frac{1}{n_d}\sum_i\bm_i$ is the average of the means of the Gaussians,
There is, in general, no closed-form expression for computing the covariance matrix 
$\overline{\bSigma}$.
{But the first order optimality condition of this convex problem shows that $\overline{\bSigma}$ is the unique positive definite fixed point of the map
\begin{equation}
    \overline{\bSigma} = \Psi\left(\overline{\bSigma}, \left\{\bSigma_k\right\}_{k=1}^{n_d}\right) \quad \text{where} \quad \Psi\left(\bA, \left\{\bB_k\right\}_{k=1}^{n_d}\right) = \frac{1}{n_d}\sum_{k=1}^{n_d} \left( \bA^{\frac{1}{2}} \bB_k \bA^{\frac{1}{2}} \right)^{\frac{1}{2}}.
    \label{eq:bw_fixed_point}
\end{equation}
Iterating this fixed-point map, \ie
\begin{equation}
    \overline{\bSigma}^{(\ell+1)} = \Psi\left(\overline{\bSigma}^{(\ell)}, \left\{\bSigma_k\right\}_{k=1}^{n_d}\right),
    \label{eq:bw_fixed_point_iterations}
\end{equation}
converges in practice to the solution $\overline{\bSigma}$~\cite{mroueh2019wasserstein}.
}
% Similarly to the distance estimation and mapping estimation, statistical estimation of the barycenter 
% from sampled distribution has been shown to have a concentration in $O(N^{-1/2})$ \cite{kroshnin2021statistical}.

\subsection{OT mapping of stationary signals using circulant matrices}
\label{sec:approx_circ}
OT emerges as a natural tool for addressing distribution shift and has found significant application in DA, as evidenced by works such as \cite{courty2014domain}, \cite{damodaran2018deepjdot}, and \cite{turrisi_multi-source_2022}.
Another line of methods leverages the Gaussian assumption to computationally simplify the estimation of Wasserstein mapping and barycenter, as illustrated in \cite{mroueh2019wasserstein} and \cite{flamary2020concentration}.
In the context of this work, the focus is on the analysis of time-series data.
Under certain judicious assumptions, the computational efficiency of Monge mapping and barycenter computation is further enhanced, as proposed by \cite{gnassounou2023convolutional}.

\subsubsection{Circulant matrices for stationary and periodic signals}
\begin{figure}[t]
    \centering
    \begin{subfigure}[b]{0.47\textwidth}
        \centering
        \includegraphics[width=0.5\linewidth]{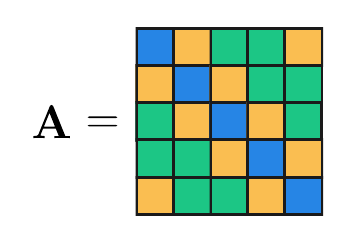}
        \caption{$\bA\in {\cC_5 \cap \cS_5^{++}}$
        with coefficients 
        $\color{Blue}\mdblksquare\color{black},
        \color{Orange}\mdblksquare\color{black},
        \color{Green}\mdblksquare \color{black}\in \bbR$}
        \label{fig:circ}
    \end{subfigure}
    \hfill
    \begin{subfigure}[b]{0.47\textwidth}
        \centering
        \includegraphics[width=0.65\linewidth]{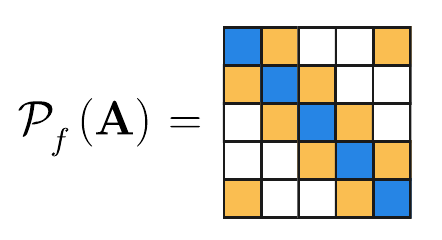}
        \caption{$\cP_{\filtersize}(\bA) \in {\cE_{3,5} \cap \cS_5^{++}}$ approximates $\bA$}
        \label{fig:approx_mapping}
     \end{subfigure}
     \caption{
        A Monge mapping $\bA$ (cf.~\autoref{eq:Monge_map}) that is a $5 \times 5$ symmetric positive definite and circulant matrix, \ie belonging to {$\cC_5 \cap \cS_5^{++}$}, and its approximation in {$\cE_{3,5} \cap \cS_5^{++}$} as defined in~\autoref{eq:def_approx}.
        In this case, for $\bx \in \bbR^{n_\ell}$, $\cP_{\filtersize}(\bA) \bx = \bh * \bx$ with $\bh = [
       \color{Orange}\mdblksquare\color{Blue}
       \mdblksquare\color{Orange}\mdblksquare\color{black}]^\trans$.
       }
     \label{fig:circ_approx}
\end{figure}

A common assumption in signal processing is that if the signal is long enough, then the signal can be considered periodic and stationary, and the covariance matrix becomes a symmetric positive definite {(\ie $\cS_{n_\ell}^{++}$)} and circulant {(\ie $\cC_{n_\ell}$)} matrix \ie a symmetric positive definite matrix with cyclically shifted rows.
\autoref{fig:circ} presents an example of such a matrix with an univariate signal of length $n_\ell=5$.
The utility of this assumption lies in its diagonalization through the Fourier basis, simplifying analyses and computations on such time series.
Indeed, for $\bA \in {\cC_{n_\ell} \cap \cS_{n_\ell}^{++}}$ and denoting the power spectrum density (PSD) 
% of $\bA$ 
by $\bq \in (\bbR^{+*})^{{n_\ell}}$, we have the decomposition{~\cite[Theorem 3.1]{gray06}}
\begin{equation}
    \label{eq:def_PSD}
    \bA = \bF_{n_\ell} \diag(\bq) \bF_{n_\ell}\hermconj \;,
\end{equation}
with the Fourier basis of elements
\begin{equation}
    \label{eq:fourier_basis}
    (\bF_{n_\ell})_{lm} \triangleq \frac{1}{\sqrt{n_\ell}}\exp\left(-2i \pi \frac{(l-1) (m-1)}{n_\ell}\right) \;,
\end{equation}
for $l,m \in \intset{n_\ell}$ and $n_\ell \in \bbN^*$.

\subsubsection{Convolutional Monge Mapping and filter size}

In the paper by \cite{gnassounou2023convolutional}, the authors use the circularity structure to diagonalize all the covariance matrices with the Fourier basis.
This leads to a simplification of the Monge mapping \autoref{eq:Monge_map}.
Let us consider two centered Gaussian distributions: a source distribution $\cN(\mathbf{0}, \bSigma_s)$ and a target distribution $\cN(\mathbf{0}, \bSigma_t)$, both with covariance matrices in {$\cC_{n_\ell} \cap \cS_{n_\ell}^{++}$}, and a realization $\bx \in \bbR^{n_\ell}$ of the source distribution.
Using Fourier-based eigen-factorization of the covariance matrices, we can map between these two Gaussian distributions, characterized by their power spectral densities (PSDs) $\bq_s \triangleq \diag(\bF_{n_\ell}\hermconj \bSigma_s \bF_{n_\ell})$ and $\bq_t \triangleq \diag(\bF_{n_\ell}\hermconj \bSigma_t\bF_{n_\ell})${, with the Monge mapping from \autoref{eq:Monge_map}.}
This mapping $\bA$ belongs to {$\cC_{n_\ell} \cap \cS_{n_\ell}^{++}$} and hence can be expressed as the following convolution~\cite[Chapter 3]{gray06}
\begin{equation}
    m(\bx) = \bh * \bx\;, \quad \text{with} \quad \bh \triangleq \frac{1}{\sqrt{n_\ell}} \bF_{n_\ell}\hermconj \left(\bq_t^{\odot\frac{1}{2}}\odot \bq_s^{\odot-\frac{1}{2}} \right) \in \bbR^{n_\ell}\;.
    \label{eq:Monge_map_conv}
\end{equation}

\noindent
Utilizing the same factorization, \cite{gnassounou2023convolutional} introduce a novel closed-form expression for barycenter computation.
Subsequently, the authors propose the Convolutional Monge Mapping Normalization (\texttt{CMMN}) method, wherein a barycenter is computed, and all domains are aligned to this central point, constituting a test-time DA approach. Notably, this method exhibits significant performance gains in practical applications.

A nuanced aspect of this technique lies in choosing the filter size in \autoref{eq:Monge_map_conv}.
% Choosing the filter size in \autoref{eq:Monge_map_conv} may have a significant impact.
{In practice, $\bh$ is computed with a size $f$ instead of $n_\ell$.}
Opting for a filter size equivalent to the signal size (\ie $\filtersize=n_\ell$) results in a perfect mapping of domains to the barycenter, potentially eliminating class-discriminative information. Selecting $\filtersize=1$ scales the entire signal, equivalent to a standard z-score operation.
% The parameter $\filtersize$ becomes instrumental in determining the extent to which signals should be aligned.
% {In addition, the longer the signal the harder the estimation problem becomes and conversely.}
Despite its significance, the impact of $\filtersize$ is not studied by \cite{gnassounou2023convolutional}.
Here, we explore the approximations made to incorporate the filter size $\filtersize$ into the theoretical framework.

\subsubsection{Approximation of circulant matrices}
% Matrices in $\cC_{n_\ell}$, and their associated filter, have $\ceil{n_\ell/2}$ parameters that is, of the order of the signal length.
% To reduce this parameter number, we propose approximating matrices in $\cC_{n_\ell}$ by linear mappings with only $\ceil{\filtersize/2} \ll \ceil{n_\ell/2}$ parameters.
% First, we define the set of symmetric circulant matrices with $\ceil{\filtersize/2}$ parameters as
% \begin{equation}
%     \cE_{\filtersize, n_\ell} \triangleq \left\{\bA \in \cC_{n_\ell} \mid (\bA)_{1l}=0, \; l \in \left\llbracket \ceil{\filtersize/2}+1 , n_\ell - \floor{\filtersize/2} \right\rrbracket\right\} \;.
% \end{equation}
{
Matrices in $\cC_{n_\ell}$, and their associated filter, have $n_\ell$ parameters, \ie the signal length.
To reduce this parameter number, we propose approximating matrices in $\cC_{n_\ell}$ by linear mappings with only $\filtersize \ll n_\ell$ parameters.
First, we define the set of circulant matrices with $\filtersize$ parameters as
\begin{equation}
    \cE_{\filtersize, n_\ell} \triangleq \left\{\bA \in \cC_{n_\ell} \mid (\bA)_{1l}=0, \; l \in \left\llbracket \ceil{\filtersize/2}+1 , n_\ell - \floor{\filtersize/2} \right\rrbracket\right\} \;.
\end{equation}
}
Then, we approximate every $\bA\in\cC_{n_\ell}$ by its nearest element in $\cE_{\filtersize, n_\ell}$, \ie
\begin{equation}
    \label{eq:def_approx}
    \cP_{\filtersize}(\bA) \triangleq \argmin_{\bGamma \in \cE_{\filtersize, n_\ell}} \, \Vert \bGamma - \bA \Vert_\text{Fro}
\end{equation}
where $\Vert . \Vert_\text{Fro}$ is the Frobenius norm.
Thus, $\cP_{\filtersize}(\bA)$ is the orthogonal projection of $\bA$ onto $\cE_{\filtersize, n_\ell}$ and
% its elements are $(\cP_{\filtersize}(\bA))_{1l} = (\bA)_{1l}$ for $l \in \intset{\ceil{\filtersize/2}}$.
{its elements are $(\cP_{\filtersize}(\bA))_{1l} = (\bA)_{1l}$ for $l \in \intset{\ceil{\filtersize/2}} \cup \llbracket n_\ell - \floor{\filtersize/2} + 1, n_\ell \rrbracket$.}
An example in $\cE_{3,5}$ is presented in \autoref{fig:approx_mapping}.
% With this approximation, we get that $\cP_{\filtersize}(\bA)$ applies a convolution with the symmetric filter $\bh\in\bbR^\filtersize$ of elements $(\bh)_{\floor{\filtersize/2}+l} = (\cP_{\filtersize}(\bA))_{1l} = (\bA)_{1l}$ for $l \in \intset{\ceil{\filtersize/2}}$,
{With this approximation, we get that $\cP_{\filtersize}(\bA)$ applies a convolution with the filter $\bh\in\bbR^\filtersize$ that contains the non-zero elements of the first row of $\bA$,}
\begin{equation}
    \label{eq:def_conv}
    \cP_{\filtersize}(\bA)\; \bx = \bh * \bx \enspace .
\end{equation}
{Since $\cP_{\filtersize}(\bA)\in \cC_{n_\ell}$, it admits the decomposition $\cP_{\filtersize}(\bA) = \bF_{n_\ell} \diag(\bq) \bF_{n_\ell}\hermconj$ with $\bq\in\bbC^{n_\ell}$ that will be interpreted latter as a cross-PSD.
An important property of $\bh$ is its computation from a sub-sampling\footnote{{In the following, we assume that the signal length ($n_\ell$) is a multiple of the filter size ($\filtersize$).}} of $\bq$, denoted $\bp \in {\bbC^f}$ and of elements
\begin{equation}
    (\bp)_l \triangleq (\bq)_{\frac{(l-1)n_\ell}{\filtersize} + 1}
\end{equation}
for $l \in \intset{\filtersize}$.}
We denote by $g_{\filtersize}$ the function that does this operation, \ie
\begin{equation}
    \label{eq:sub-sampling}
    g_{\filtersize}(\bq) \triangleq \bp \; .
\end{equation}
The following proposition bridges the gap between filtering, $\cE_{\filtersize, n_\ell}$ and $g_f$.
This way, we can sub-sample a given {cross-PSD} to get a smaller one of size $\filtersize$.
The associated filter is computed with an inverse Fourier transform and the mapping is achieved with a convolution.
Overall, the number of parameters is {$\filtersize$} and the complexity to compute the filter and the mapping are $\mathcal{O}(f\log(f))$ and $\mathcal{O}(n_\ell \log(n_\ell))$ respectively thanks to the Fast Fourier Transform (FFT).
\begin{proposition}[$\cE_{\filtersize, n_\ell}$ and filtering]
    \label{prop:filtering}
    Let $\bF_{n_\ell} \in \bbC^{n_\ell\times n_\ell}$ and $\bF_{\filtersize} \in \bbC^{\filtersize\times \filtersize}$ be the Fourier bases.
    Given $\bA = \bF_{n_\ell} \diag(\bq) \bF_{n_\ell}\hermconj \in \cE_{\filtersize,n_\ell}$ {with $\bq \in \bbC^{n_\ell}$}, and {$\bh \in \bbR^{\filtersize}$ the filter containing the non-zero elements of the first row of $\bA$}, then
    for every $\bx \in \bbR^{n_\ell}$, we have
    \begin{equation*}
        \bA \bx = \bh * \bx
    \end{equation*}
    where $*$ is the convolution operator and $\bh$ is the inverse Fourier transform of $\bp = g_{\filtersize}(\bq) \in {\bbC^f}$, \ie
    \begin{equation*}
        \bh = \frac{1}{\sqrt{\filtersize}} \bF_{\filtersize}\hermconj \bp \;.
    \end{equation*}
\end{proposition}

\section{Spatio-Temporal Monge mapping and barycenter for Gaussian signals}

This section introduces a new method for mapping multivariate signals
%overcoming the limitations of existing approaches
by seamlessly incorporating the filter size parameter and using the previously presented approximation of circulant matrices.
We first describe the structure of the spatio-temporal covariance matrix.
Then, we define the $\filtersize$-Monge mapping and compute its closed-form expression for the spatio-temporal structure as well as the associated Wasserstein barycenter.
Next, we establish two mappings and their barycenters when only the temporal or spatial structure is considered.
In the following, we consider multivariate signals $\bX \triangleq [\bx_1, \dots, \bx_{n_c}]^\trans \in \bbR^{n_c \times n_\ell}$ with $n_c$ channels of length $n_\ell$, and assume that the vectorized signal follows a centered Gaussian distribution, \ie $\Vectr(\bX) \sim \mathcal{N}(\mathbf{0}, \bSigma)$ where $\bSigma \in \cS_{n_c n_\ell}^{++}$ is the spatio-temporal covariance matrix.

\subsection{Spatio-temporal covariance matrix}
\label{sec:math}

First, the spatio-temporal covariance matrix is expressed as
\begin{equation}
    \bSigma = 
    \bbE\left[ \Vectr(\bX)\Vectr(\bX)^\trans\right]
    =
    \begin{pmatrix}
        \bSigma_{1, 1} & \dots & \bSigma_{1, n_c} \\
        \dots & \dots & \dots \\
        \bSigma_{n_c, 1} & \dots & \bSigma_{n_c, n_c} 
    \end{pmatrix} \in \cS_{n_c n_\ell}^{++}
    \;,
    \label{eq:cov}
\end{equation}
where $\bSigma_{l,m}$ corresponds to the temporal covariance matrix between the signals of the \textit{l}$^{th}$ and the \textit{m}$^{th}$ channels.
We add a simple assumption on the $\bSigma_{l,m}$ that will lead to a simple Monge mapping later in this section.
\begin{assumption}
    %Let a centered Gaussian distribution $\cN(\mathbf{0}, \bSigma)$ with $\bSigma \in \cS_{n_c n_\ell}^{++}$.
    For every $l,m\in \intset{n_c}^2$, we assume that $\bSigma_{l,m}$, defined in~\autoref{eq:cov}, belongs to $\cC_{n_\ell}$, \ie
    \begin{equation*}
        \bSigma_{l,m} = \bF_{n_\ell} \diag(\bq_{l,m}) \bF_{n_\ell}\hermconj,
    \end{equation*}
    {where $\bq_{l,m} \in \bbC^{n_\ell}$ is such that the cross-PSD $\bQ_\ell \triangleq \left[(\bq_{l,m})_\ell\right]_{l,m=1,1}^{n_c,n_c} \in \cH_{n_c}^{++}$, for $\ell \in \intset{n_\ell}$.}
    % where $\bq_{l,m} \in \bbR^{n_\ell}$ is the cross-PSD between the signals of the \textit{l}$^{th}$ channel and the \textit{m}$^{th}$ channel such that the matrices $\bQ_\ell \triangleq ((\bq_{l,m})_\ell)_{lm}$ belong to $\cS_{n_\ell}^{++}$.
    % For $l=m$, $\bq_{l,l}$ boils down to the PSD. 
    \label{assu:spat_temp}
\end{assumption}
\begin{figure}[t]
    \centering
    \includegraphics[width=\textwidth]{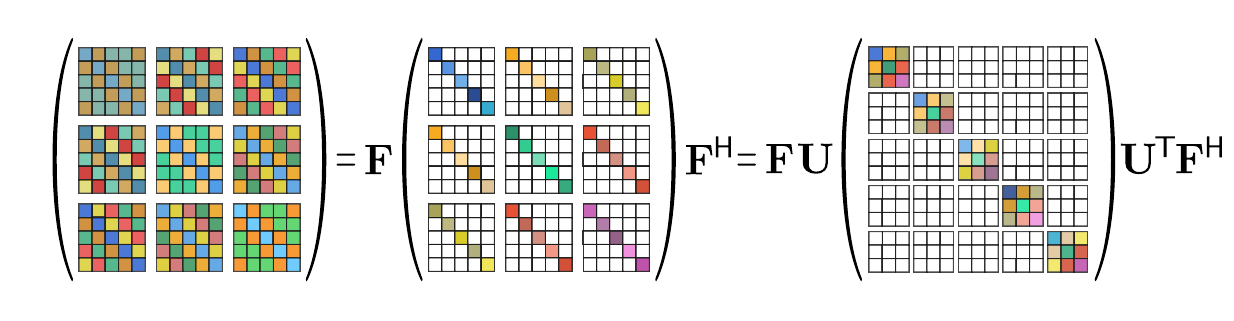}
    \caption{
        Block diagonalization of the covariance matrix $\bSigma \in \cS_{n_c n_\ell}^{++}$ following the Assumption~\ref{assu:spat_temp} with $n_\ell=5$ and $n_c = 3$.
        $\bF = \diag(\bF_{n_\ell}, \dots, \bF_{n_\ell}) \in \bbC^{n_c n_\ell}$ and $\bU$ is a permutation matrix.
    }
    \label{fig:block_cov}
\end{figure}
\noindent
Under the Assumption~\ref{assu:spat_temp}, $\bSigma$ is re-written
\begin{equation}
    \bSigma = 
    \bF
    \begin{pmatrix}
        \diag(\bq_{1, 1})& \dots & \diag(\bq_{1, n_c}) \\
        \dots & \dots & \dots \\
        \diag(\bq_{n_c, 1}) & \dots & \diag(\bq_{n_c, n_c})
    \end{pmatrix}
    \bF\hermconj\;,
\end{equation}
with $\bF \triangleq \diag(\bF_{n_\ell}, \dots, \bF_{n_\ell}) \in \bbC^{n_c n_\ell \times n_c n_\ell}$.
Hence, there exists a permutation matrix $\bU \in \bbR^{n_c n_\ell \times n_c n_\ell}$ that block-diagonalizes $\bSigma$, \ie
\begin{equation}
    \bSigma = \bF \bU \bQ \bU^\trans \bF\hermconj\;,
    \label{eq:blockdiag_T}
\end{equation}
with $\bQ \triangleq \diag(\bQ_1, \dots, \bQ_{n_\ell}) \in {\cH_{n_c n_\ell}^{++}}$ and {$\bQ_\ell$ defined in Assumption~\ref{assu:spat_temp}.}
% \begin{equation}
%     \bQ_l = \begin{pmatrix}    
%         (\bq_{1, 1})_l & \dots & (\bq_{1, n_c})_l \\
%         \dots & \dots & \dots \\
%         (\bq_{n_c, 1})_l & \dots & (\bq_{n_c, n_c})_l \\
%     \end{pmatrix} \in \cS_{n_c}^{++}\;.
% \end{equation}
An example of this decomposition for $n_\ell=5$ and $n_c=3$ is given in \autoref{fig:block_cov}.
Notably, $\bQ$ is the {cross-PSD} and can be sub-sampled by extending the function $g_{\filtersize}$ to block-diagonal matrices, \ie
\begin{equation}
    g_{\filtersize}(\bQ) \triangleq \diag\left(\bQ_1, \bQ_{\frac{n_\ell}{\filtersize} + 1}, \dots, \bQ_{\frac{(\filtersize-1)n_\ell}{\filtersize} + 1}\right) \in {\cH_{n_c \filtersize}^{++}}.
\end{equation}
Thus, we can sub-sample the cross-PSDs of the blocks of $\bSigma$,
\begin{equation}
    \begin{pmatrix}
        \diag(\bp_{1, 1})& \dots & \diag(\bp_{1, n_c}) \\
        \dots & \dots & \dots \\
        \diag(\bp_{n_c, 1}) & \dots & \diag(\bp_{n_c, n_c})
    \end{pmatrix}
    = \bV g_{\filtersize}(\bQ) \bV^\trans \in {\cH_{n_c \filtersize}^{++}}
    \label{eq:blockdiag_{\filtersize}}
\end{equation}
with $\bp_{l,m} = g_{\filtersize}(\bq_{l,m}) \in {\bbC^f}$ and $\bV \in \bbR^{n_c \filtersize \times n_c \filtersize}$ a permutation matrix.

\subsection[Spatio-Temporal Monge mapping and barycenter]{Spatio-Temporal $\filtersize$-Monge mapping and barycenter}
The Assumption~\ref{assu:spat_temp} implies a specific structure on the Monge mapping from~\autoref{eq:Monge_map}.
Indeed, given $\cN(\mathbf{0}, \bSigma_s)$ and $\cN(\mathbf{0}, \bSigma_t)$ source and target centered Gaussian distributions respectively and following the Assumption~\ref{assu:spat_temp},
the Monge mapping, introduced in the~\autoref{eq:Monge_map}, has the structure
\begin{equation*}
    \bA = 
    \begin{pmatrix}
        \bA_{1, 1} & \dots & \bA_{1, n_c} \\
        \dots & \dots & \dots \\
        \bA_{n_c, 1} & \dots & \bA_{n_c, n_c} 
    \end{pmatrix} \in \cS_{n_c n_\ell}^{++}
    \quad \text{with} \quad
    \bA_{l,m}\in \cC_{n_\ell}
\end{equation*}
for every $l,m \in \intset{n_c}$.
This motivates the definition of the $\filtersize$-Monge mapping which leverages the approximation of circulant matrices defined in \autoref{sec:approx_circ}.  
\begin{samepage}
\begin{definition}
    \label{def:approx_mapping}
    Given $\cN(\mathbf{0}, \bSigma_s)$ and $\cN(\mathbf{0}, \bSigma_t)$ source and target centered Gaussian distributions respectively and following the Assumption~\ref{assu:spat_temp},
    the $\filtersize$-Monge mapping on $\bX \in \bbR^{n_c \times n_\ell}$ is defined as
    \begin{equation*}
        m_{\filtersize} (\bX) = \Vectr^{-1}\left(\widetilde{\bA} \Vectr\left(\bX\right)\right)
    \end{equation*}
    where
    \begin{equation*}
        \widetilde{\bA} =
        \begin{pmatrix}
            \cP_{\filtersize}(\bA_{1, 1}) & \dots & \cP_{\filtersize}(\bA_{1, n_c}) \\
            \dots & \dots & \dots \\
            \cP_{\filtersize}(\bA_{n_c, 1})  & \dots & \cP_{\filtersize}(\bA_{n_c, n_c}) 
        \end{pmatrix} \quad \text{and} \quad \bA = \bSigma_s^{-\frac{1}{2}}\left( \bSigma_s^{\frac{1}{2}} \bSigma_t\bSigma_s^{\frac{1}{2}} \right)^{\frac{1}{2}}  \bSigma_s^{-\frac{1}{2}}
    \end{equation*}
    is the Monge mapping from \autoref{eq:Monge_map}
    and $\cP_{\filtersize}$ is the function which sub-samples the {cross-PSDs} and defined in~\autoref{eq:def_approx}.
\end{definition}
\end{samepage}

\begin{figure}[t]
    \begin{subfigure}{.5\textwidth}
        \includegraphics[width=1\linewidth]{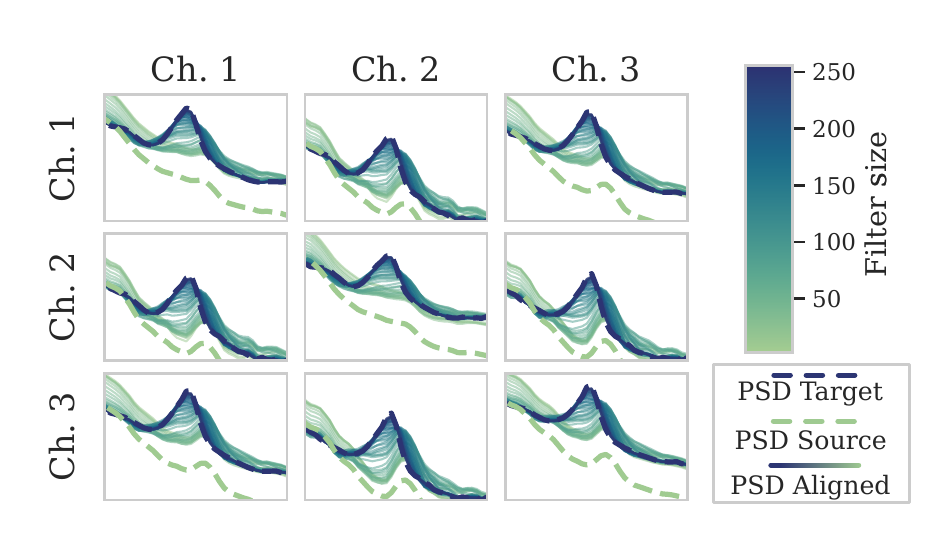}
        \caption{Cross-PSD alignment}
    \end{subfigure}
    \centering
    \begin{subfigure}{.4\textwidth}
        \includegraphics[width=1\linewidth]{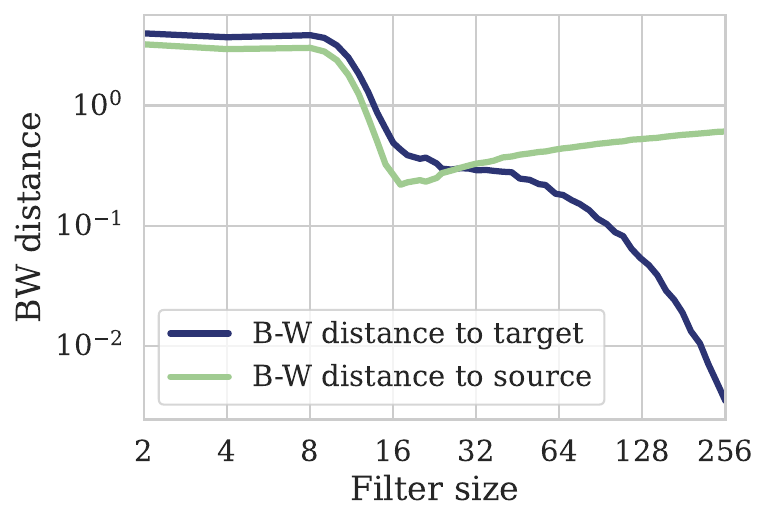}
        \caption{Bures-Wasserstein distance}
    \end{subfigure}
    \caption{Illustration of the Spatio-Temporal Monge mapping on sleep data. (a) Cross-PSD alignment from source cross-PSD (green dotted line) to target cross-PSD (blue dotted line) for different filter sizes. (b) Bures-Wasserstein distance between the aligned signal and the source signal (green line) and the target line (blue line).}
    \label{fig:psd_mapping}
\end{figure}

\noindent
{We now explore the properties of the $\filtersize$-Monge mapping.
The next proposition establishes the spatio-temporal mapping, showing that it is a sum of convolutions and thus fast to compute.}
\begin{proposition}[Spatio-Temporal mapping]
    \label{prop:spatio_temp_map}
    Let $\cN(\mathbf{0}, \bSigma_s)$ and $\cN(\mathbf{0}, \bSigma_t)$ be source and target centered Gaussian distributions respectively and following the Assumption~\ref{assu:spat_temp}, \ie for $d\in\{s,t\}$,
    $\bSigma_d = \bF \bU \bQ_d \bU^\trans \bF\hermconj$
    as defined in~\autoref{eq:blockdiag_T}.
    % having the structure defined in \autoref{eq:cov} with block matrices belonging to $\cE_{\filtersize, n_\ell}$.
    Let $\filtersize\leq n_\ell$, $\bP_d = g_{\filtersize}(\bQ_d)$ and
    \begin{equation*}
        \begin{pmatrix}
            \diag(\bp_{1, 1}) & \dots & \diag(\bp_{1, n_c}) \\
            \dots & \dots & \dots \\
            \diag(\bp_{n_c, 1})  & \dots & \diag(\bp_{n_c, n_c}) 
        \end{pmatrix}
        \triangleq \bV \bP_s^{-\frac{1}{2}} \left(\bP_s^{\frac{1}{2}} \bP_t \bP_s^{\frac{1}{2}}\right)^{\frac{1}{2}} \bP_s^{-\frac{1}{2}} \bV^\trans \in {\cH_{n_cf}^{++}}
    \end{equation*}
    with $\bV\in\bbR^{n_c \filtersize \times n_c \filtersize}$ the permutation matrix from \autoref{eq:blockdiag_{\filtersize}}.
    \sloppy
    Given a signal $\bX = [\bx_1, \dots, \bx_{n_c}]^\trans \in \bbR^{n_c \times n_\ell}$ such that $\Vectr(\bX) \sim \cN(\mathbf{0}, \bSigma_s)$, the $\filtersize$-Monge mapping introduced in the Definition~\ref{def:approx_mapping} is a sum of convolutions
    \begin{equation*}
        m_{\filtersize}\left(\bX\right) =  \left[\sum_{j=1}^{n_c} \bh_{1,j} * \bx_j, \dots, \sum_{j=1}^{n_c} \bh_{n_c,j} * \bx_j \right]^\trans
    \end{equation*}
    \sloppy
    where
    $\bh_{i,j} = \frac{1}{\sqrt{\filtersize}} \bF_{\filtersize}\hermconj \bp_{i,j}  \in \bbR^{\filtersize}$.
    % $\bh_{i,j} = \frac{1}{\sqrt{\filtersize}} \bS_F\bF_{\filtersize}\hermconj \bp_{i,j}  \in \bbR^{\filtersize}$.
\end{proposition}

\noindent
The Wasserstein barycenter of multiple Gaussian distributions is the solution of a fixed-point equation as seen in \autoref{sec:bary}.
It is possible to exploit the structure of the spatio-temporal covariances to reduce the computational complexity of the barycenter by dealing with block diagonal matrices.
\begin{lemma}[Spatio-Temporal barycenter]
    \label{lemma:spatio_temp_bary}
    Let $n_d$ centered Gaussian distributions $\cN(\mathbf{0}, \bSigma_k)$ with  $\bSigma_k = \bF \bU \bQ_k \bU^\trans \bF\hermconj$ following the Assumption~\ref{assu:spat_temp}. The Wasserstein barycenter of the $n_d$ distributions is the unique centered Gaussian distribution $\cN(\mathbf{0}, \overline{\bSigma})$ satisfying
    % \begin{equation*}
    %     \overline{\bSigma} = \bF \bU \overline{\bQ} \bU^\trans \bF\hermconj \quad \text{and} \quad \overline\bQ = \frac{1}{n_d}\sum_{k=1}^{n_d} \left(\overline\bQ^{\frac{1}{2}} \bQ_k \overline\bQ^{\frac{1}{2}}\right)^{\frac{1}{2}} \; .
    % \end{equation*}
    \begin{equation*}
        \overline{\bSigma} = \bF \bU \overline{\bQ} \bU^\trans \bF\hermconj \quad \text{and} \quad \overline\bQ = \Psi\left(\overline\bQ, \left\{\bQ_k\right\}_{k=1}^{n_d} \right) \; ,
    \end{equation*}
    where $\Psi$ is defined in \autoref{eq:bw_fixed_point}.
\end{lemma}

\subsection[Special cases: Temporal and Spatial Monge mappings]{Special cases: Temporal and Spatial $f$-Monge mappings}

\subsubsection[Pure temporal Monge mapping and barycenter]{Pure temporal $\filtersize$-Monge mapping and barycenter}
\label{sec:temp}
In some other scenarios, it may be possible that spatial correlation is less relevant in the learning process.
In this context, we assume the signals are uncorrelated between the different channels so that the covariance matrix for each channel is circulant.
{This setting leads to the mapping and barycenter formulas studied in \cite{gnassounou2023convolutional}, here considering $n_c$ uncorrelated sensors. To detail this, we introduce a new
structure for $\bSigma$ in the following assumption.}
\begin{assumption}
    \label{assu:temp}
%    Let a centered Gaussian distribution $\cN(\mathbf{0}, \bSigma)$ with $\bSigma \in \cS_{n_c n_\ell}^{++}$.
    We assume that the different channels are uncorrelated and that the covariance matrix for each channel is circulant, \ie
    \begin{equation*}
        \bSigma = \bF \diag(\bq_{1}, \dots, \bq_{n_c}) \bF\hermconj \in \cS_{n_c n_\ell}^{++},
    \end{equation*}
    where $\bq_{c} \in (\bbR^{+*})^{n_\ell}$ are PSDs.
\end{assumption}

\noindent
This enables a novel reformulation of the $\filtersize$-Monge mapping, specifically tailored to accommodate data that are only temporally correlated.

\begin{proposition}[Temporal mapping]
    \label{prop:temp_map}
    \sloppy
    Let $\cN(\mathbf{0}, \bSigma_s)$ and $\cN(\mathbf{0}, \bSigma_t)$ be source and target centered Gaussian distributions respectively and following the Assumption~\ref{assu:temp}, \ie for $d \in \{s,t\}$, $\bSigma_d = \bF \diag(\bq_{1,d}, \dots, \bq_{n_c,d}) \bF\hermconj$.
    Let $\filtersize\leq n_\ell$, and $\bp_{c,d} = g_{\filtersize}(\bq_{c,d})$.
    Given a signal $\bX = [\bx_1, \dots, \bx_{n_c}]^\trans \in \bbR^{n_c \times n_\ell}$ such that $\Vectr(\bX) \sim \cN(\mathbf{0}, \bSigma_s)$,
    the $\filtersize$-Monge mapping introduced in Definition~\ref{def:approx_mapping} is the convolution
    \begin{equation*}
        m_{\filtersize}(\bX) = \left[\bh_1 * \bx_1, \dots, \bh_{n_c} * \bx_{n_c} \right]^\top
    \end{equation*}
    \sloppy where $\bh_c = \frac{1}{\sqrt{\filtersize}} \bF_{\filtersize}\hermconj \left({\bp_{c,t}}^{\odot\frac{1}{2}}\odot {\bp_{c,s}}^{\odot-\frac{1}{2}} \right) \in \bbR^{\filtersize}$.
    % \sloppy where $\bh_i = \frac{1}{\sqrt{\filtersize}} \bS_F\bF_{\filtersize}\hermconj \left({\bp_{i,t}}^{\odot\frac{1}{2}}\odot {\bp_{i,s}}^{\odot-\frac{1}{2}} \right) \in \bbR^{\filtersize}$.
\end{proposition}

\noindent
Furthermore, in this case we recover a closed-form solution to the Wasserstein barycenter fixed-point equation, which enables its computation without the need for iterative methods. This leads to significantly improved computational efficiency.

\begin{lemma}[Temporal barycenter]
    \label{lemma:temp_bary}
    \sloppy Let $n_d$ centered Gaussian distributions $\cN(\mathbf{0}, \bSigma_k)$ with $\bSigma_k = \bF \diag(\bq_{1,k}, \dots, \bq_{n_c,k}) \bF\hermconj$ following the Assumption~\ref{assu:temp}. The Wasserstein barycenter of the $n_d$ distributions is the centered Gaussian distribution $\cN(\mathbf{0}, \overline{\bSigma})$ with
    \begin{equation*}
        \overline{\bSigma} = \bF \diag(\overline\bq_1, \dots, \overline\bq_{n_c}) \bF\hermconj \quad \text{and} \quad \overline\bq_c = \left( \frac{1}{n_d} \sum^{n_d}_{k=1} \bq_{c,k}^{\odot\frac{1}{2}}\right)^{\odot 2}\; .
    \end{equation*}
\end{lemma}

\subsubsection[Pure spatial Monge mapping and barycenter]{Pure spatial Monge mapping and barycenter}

Conversely, in some other alternative scenarios, the relevance of temporal correlation in the learning process may be diminished. 
Here, we consider only correlations across different channels, that is one set $\filtersize=1$. Adopting a correlation matrix between channels $\bXi = \bbE(\bX\bX^\trans) \in \cS_{n_c}^{++}$, this assumption gives rise to a distinctive structure for $\bSigma \in \cS_{n_c n_\ell}^{++}$.
\begin{assumption}
%    Let a centered Gaussian distribution $\cN(\mathbf{0}, \bSigma)$ with $\bSigma \in \cS_{n_c n_\ell}^{++}$.
    The random signal has a spherical covariance matrix $\bSigma \in \cS_{n_c n_\ell}^{++}$ w.r.t. time.
    Formally for every $l,m\in \intset{n_c}$, $\bSigma_{l,m}$ defined in~\autoref{eq:cov} belongs to $\cE_{1,n_\ell}$, \ie
    \begin{equation*}
        \bSigma_{l,m} = \sigma_{l,m} \bI_{n_\ell} \quad \text{with} \quad \sigma_{l,m} > 0 \;.
    \end{equation*}
    This implies that the overall covariance matrix is given by a Kronecker product
    \begin{equation*}
        \bSigma = \bXi \otimes \bI_{n_\ell}
        \quad \text{with} \quad 
        \bXi = 
        \begin{pmatrix}
            \sigma_{1,1} & \dots & \sigma_{1,n_c} \\
            \dots & \dots & \dots \\
            \sigma_{n_c,1} & \dots & \sigma_{n_c,n_c}
        \end{pmatrix} \in \cS_{n_c}^{++}\;.
    \end{equation*}
    \label{assu:spat}
\end{assumption}
This new structure leads to the following Monge mapping,  specifically designed to accommodate data which are only spatially correlated.

\begin{proposition}[Spatial mapping]
    \label{prop:spat_map}
    Let $\cN(\mathbf{0}, \bSigma_s)$ and $\cN(\mathbf{0}, \bSigma_t)$ be source and target centered Gaussian distributions respectively and following the Assumption~\ref{assu:spat}, \ie for $d\in\{s,t\}$, $\bSigma_d = \bXi_d \otimes \bI_{n_\ell}$.
    Given a signal $\bX = [\bx_1, \dots, \bx_{n_c}]^\trans \in \bbR^{n_c \times n_\ell}$ such that $\Vectr(\bX) \sim \cN(\mathbf{0}, \bSigma_s)$,
    the $\filtersize$-Monge mapping introduced in Definition~\ref{def:approx_mapping} is
    \begin{equation*}
        m_{\filtersize}(\bX) = \bXi_s^{-\frac{1}{2}}\left( \bXi_s^{\frac{1}{2}} \bXi_t\bXi_s^{\frac{1}{2}} \right)^{\frac{1}{2}}  \bXi_s^{-\frac{1}{2}} \bX \;.
    \end{equation*}
\end{proposition}

\noindent
It is interesting to note that when covariance matrices commute, this boils down
to the \texttt{CORAL} mapping proposed by \cite{sun_correlation_2016}, as already observed by \cite{mroueh2019wasserstein}. Also note that Monge mappings are restricted to semi-definite mappings
which is not the case of CORAL.
The purely spatial barycenter still requires a fixed-point algorithm, albeit with reduced complexity since the size of the covariance matrices is $n_c \times n_c$.
\begin{lemma}[Spatial barycenter]
    \label{lemma:spat_bary}
    \sloppy Let $n_d$ centered Gaussian distributions $\cN(\mathbf{0}, \bSigma_k)$ with $\bSigma_k = \bXi_k \otimes \bI_{n_\ell}$ following the Assumption~\ref{assu:spat}. The Wasserstein barycenter of the $n_d$ distributions is the unique centered Gaussian distribution $\cN(\mathbf{0}, \overline{\bSigma})$ satisfying
    \begin{equation*}
        \overline{\bSigma} = \overline{\bXi} \otimes \bI_{n_\ell} \quad \text{and} \quad \overline{\bXi} = \Psi\left(\overline\bXi, \left\{\bXi_k\right\}_{k=1}^{n_d} \right)\; ,
    \end{equation*}
    where $\Psi$ is defined in \autoref{eq:bw_fixed_point}.
\end{lemma}

\section{Multi-source Signal Adaptation with Monge Alignment}
\label{sec:CMMN}

% {
%     In this section, we introduce the Monge Alignment (\method{}) method that
%     leverages the Monge mappings and barycenters developed in the previous
%     sections to perform DA (See \autoref{fig:concept_figure}).
%     Indeed, \method{} pushes forward the distributions of the different domains to the barycenter of source domains.
%     By doing so, we effectively reduce the discrepancy between the source and target domains, enabling better knowledge transfer and improved generalization of any predictor trained on these adapted source domains.
%     Notably, the proposed approach is test-time, \ie adapts new target domains during testing.
%     Hence, it eliminates the need to train a new model at test-time, which is a typical requirement of traditional DA methods~\cite{ganin2016domainadversarial, sun_deep_2016, damodaran2018deepjdot}.

In this section, we introduce the Monge Alignment (\method{}) method that
leverages the Monge mappings and barycenters developed in the previous
sections to perform DA (see \autoref{fig:concept_figure}).
Indeed, \method{} pushes forward the distributions of the different domains to the barycenter of source domains.
By doing so, we effectively reduce the discrepancy between the source and target domains, improving generalization of any predictor trained on these adapted source domains.
Notably, the proposed approach can be applied at test-time, by performing
adaptation to new target domains without the need to retrain a predictor.
% Hence, it eliminates the need to train a new model at test-time, which is a typical requirement of traditional DA methods~\cite{ganin2016domainadversarial, sun_deep_2016, damodaran2018deepjdot}.
First, we describe the general algorithm \method{} at train and test times.
It can be implemented in three specific ways, depending on whether we consider temporal shifts (\tmethod{}), spatial shifts (\smethod{}), or both (\stmethod{}).
Second, the computational aspects of these methods are presented with the computational complexities and the choice of filter size.
Third, the statistical estimation errors made by these algorithms are assessed.
In the following, given $n_d$ labeled source domains, we denote by $\bX_k \in \bbR^{n_c \times n_\ell}$ the data of each domain $k \in \intset{n_d}$, with corresponding labels $\by_k$.
Also, we denote by $\bX_t \in \bbR^{n_c \times n_\ell}$, the target data.

\subsection{Monge Alignment algorithm}

At train-time, \method{} adapts each source domain to their barycenter with the $\filtersize$-Monge mapping from Definition~\ref{def:approx_mapping} 
and then a predictor is trained.
% Then, a predictor is trained on the adapted data.
At test-time, \method{} adapts each target domain to the previously learned barycenter using the $\filtersize$-Monge mapping.
Remarkably, this adaption is performed without accessing the source data.
Finally, inference is achieved on the adapted target data using the trained predictor.
These two phases are presented in \autoref{alg:train_time} and \autoref{alg:test_time}.

% The \autoref{fig:bary_32} and \autoref{fig:bary_256} display an example of barycenter alignment for \stmethod{} method with two values of filter size.
% The bigger the filter size, the more the signal PSD is aligned to the barycenter. The choice of the $\filtersize$ allows us to choose how much we want to align the covariance to the barycenter and, at the same time, allows us to reduce the time complexity (see \autoref{tab:complexity}).

% The proposed method is a test-time DA approach with two phases.
% During the first phase, we train the predictor on normalized data at train-time.
% Then, in the second phase, we adapt the target data to the source domains without accessing the source data at test-time, and employ the predictor learned during train-time.

% The proposed method is a test-time DA method that involves two phases.
% During the first phase, the predictor is trained on normalized data (Train-time). 
% Then, during the second phase (test-time), the target data is adapted to the
% source domains without access to the source data and the predictor learned at
% train time is used. 

\begin{figure*}[t]
\begin{minipage}{.45\textwidth}
    % \RestyleAlgo{boxruled}
      \begin{algorithm}[H]
          \DontPrintSemicolon
          \KwIn{Filter size $\filtersize$, source data $\{\bX_k\}_{k=1}^{n_d}$, source labels $\{\by_k\}_{k=1}^{n_d}$}
       %   Choose alignment: spatial, temporal, spatio-temporal\;
          $\widehat \bSigma_k \leftarrow $ Compute cov. matrix from $\bX_k,
          \forall k$\;
          $\widehat{\overline\bSigma}$ $\leftarrow$ Learn barycenter
          \;
          $\widehat{m}_{\filtersize, k}$ $\leftarrow$ Learn $\filtersize$-Monge maps, $\forall k$\;
     %     $\{ \bX_k\}_{k=1}^{n_d}$  $\leftarrow$ Apply $\filtersize$-Monge maps
        %  $\forall k$\;
          $\widehat h \!\! \leftarrow$ \!\! Train \! on \! adapted \! data \!\! $\{ \widehat{m}_{\filtersize, k}(\bX_k)\}_{k=1}^{n_d}$\;
      \Return Trained predictor $\widehat h$, barycenter $\widehat{\overline\bSigma}$
      
      \caption{\method{} at Train-Time} 
      \label{alg:train_time}
      \end{algorithm}
      \end{minipage}
      \hfill
      \raisebox{0.19\height}{  
      \begin{minipage}{.45\textwidth}
      % \RestyleAlgo{boxruled}
       \begin{algorithm}[H]
      \DontPrintSemicolon
      \KwIn{Target data $\bX_t$, trained model $\widehat h$, barycenter $\widehat{\overline\bSigma}$}
      $\widehat \bSigma_t \leftarrow $ Compute cov. matrix from $\bX_t$\;
      $\widehat{m}_{\filtersize, t}$ $\leftarrow$ Learn $\filtersize$-Monge  map\;
      % $\widetilde \bX_t$ $\leftarrow$ Apply $\filtersize$-Monge map $\widehat{m}_{\filtersize, t}$ \;
      $\widehat\by_t$ $\leftarrow$ $\widehat h\left( \widehat m_{f,t}(\bX_t)\right)$ \;
      \Return  Predictions $\widehat \by_t$
      \caption{\method{} at Test-Time}
      \label{alg:test_time}
      \end{algorithm}
  \end{minipage}}
   % \caption{caption}
    \label{fig:label}
\end{figure*}

\subsubsection{Train-time} 

\method{} performs four steps to learn a predictor from adapted source data and their labels.

\paragraph{Covariance matrix computation (\autoref{alg:train_time}, line 1)}
The first step involves estimating the covariance matrices $\widehat
\bSigma_k$ for each source domain $k \in \intset{n_d}$. 
% of the $n_d$ source domains. 
% (as outlined in line 1 of \autoref{alg:train_time}).
For each source domain, we have a multivariate signal
$\bX = [\bx_1, \dots, \bx_{n_c}]^\trans$. Note that, we can estimate either the covariance matrices or their Fourier domain equivalents, the cross-PSDs.
For both \stmethod{} and \tmethod{}, we employ the Welch estimator~\cite{welch1967use} to estimate the cross-PSDs.
This involves computing the short-time Fourier transform of the signals.
For all $l \in \intset{n_\ell-\filtersize+1}, j \in \intset{\filtersize}, c \in \intset{n_c}$, it is given by
\begin{equation}
    \widehat{x}_{c,l,j} = \sum_{k=1}^\filtersize w_k e^\frac{-2i\pi(j-1)(k-1)}{\filtersize} (\bX)_{c,l+k-1}\;.
\end{equation}
where $\bw = [w_1, \dots, w_\filtersize]^\trans$ is the window function such that $\Vert \bw \Vert_2 = 1$.
For an overlap of $\frac{\filtersize}{2}$ samples between adjacent windows, the Welch estimator used for \stmethod{} is given by
\begin{equation}
    \widehat{\bP} = \diag\left( \widehat{\bP}_1, \dots, \widehat{\bP}_f \right) \in \cH_{n_c \filtersize}^{++}
    \quad \text{where} \quad \left\{
    \begin{aligned}
        & \widehat{\bP}_j = \frac{1}{n_\ell-\filtersize+1} \widehat{\bX}_j \widehat{\bX}_j\hermconj \\ % \in \cH_{n_c}^{++} \\
        & (\widehat{\bX}_j)_{c,l} = \widehat{x}_{c,l,j} \; .
    \end{aligned}
    \right.
    \label{eq:welch_stma}
\end{equation}
In the same manner, the Welch estimator for \tmethod{} provides us with the
diagonal $\widehat{\bp}$ of the cross-PSDs,
\begin{equation}
    \widehat{\bp} = \left[\widehat{\bp}_1^\trans, \dots, \widehat{\bp}_{n_c}^\trans\right]^\trans \in (\bbR^{+*})^{n_c \filtersize}
    \quad \text{where} \quad \left\{
    \begin{aligned}
        &\widehat{\bp}_c = \frac{1}{n_\ell - f +1} \diag\left(\widehat{\bZ}_c \widehat{\bZ}_c\hermconj \right) \\ %\in (\bbR^{+*})^{\filtersize} \\
        & (\widehat{\bZ}_c)_{j,l} = \widehat{x}_{c,l,j} \;.
    \end{aligned}
    \right.
    \label{eq:welch_tma}
\end{equation}
For the \smethod{} method, we employ the classical empirical
covariance matrix estimator, which is suitable since \smethod{} only
requires spatial covariance information, and signals are assumed to be centered,
\begin{equation}
    \widehat{\bXi} = \frac{1}{n_\ell} \bX{\bX}^\trans \in \cS_{n_c}^{++} \;.
    \label{eq:scm_sma}
\end{equation}
% Since our data are supposed to be Gaussian-centered, the barycentric distribution is defined by the covariance $\overline\bSigma$, which can be expressed with the computed covariances.
% The next step is to calculate the barycenter (see line 3 in \autoref{alg:train_time}) with the closed-form if one exists (\ie \tmethod{} \cite{gnassounou2023convolutional}) or use the fixed-point iterations (\ie \stmethod{} and \smethod{}).
% In practice, we initialize the barycenter with the Euclidean mean over the domain's covariances. Then, fixed-point iterations are done until convergence.
% The \autoref{fig:bary_compute} shows an example of barycenter computation for \stmethod{} method. The green line corresponds to the source cross-PSD; the black dotted line is their barycenter.

\begin{figure}[t]
    \begin{subfigure}{.32\textwidth}
        \includegraphics[width=\linewidth]{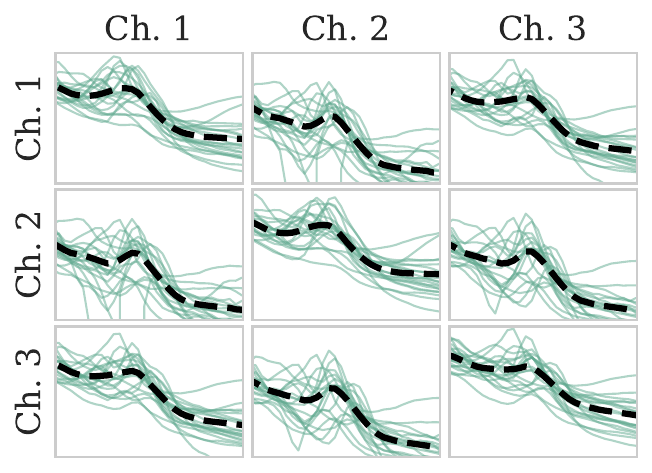}
        \caption{Source cross-PSD}
        \label{fig:bary_compute}
    \end{subfigure}
    \centering
    \begin{subfigure}{.32\textwidth}
        \includegraphics[width=\linewidth]{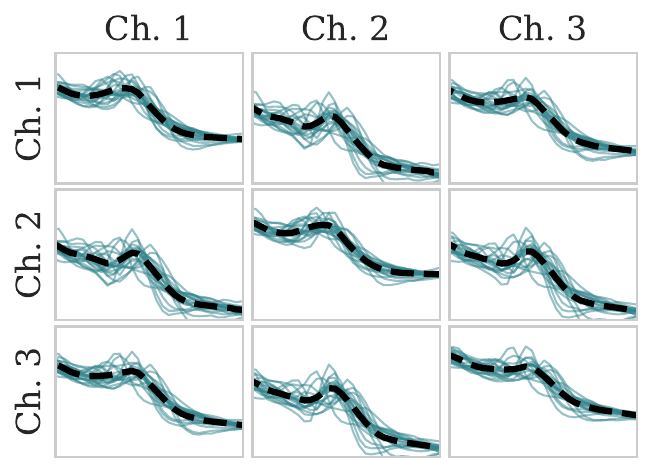}
        \caption{Alignment with $\filtersize=32$}
        \label{fig:bary_32}
    \end{subfigure}
    \begin{subfigure}{.32\textwidth}
        \includegraphics[width=\linewidth]{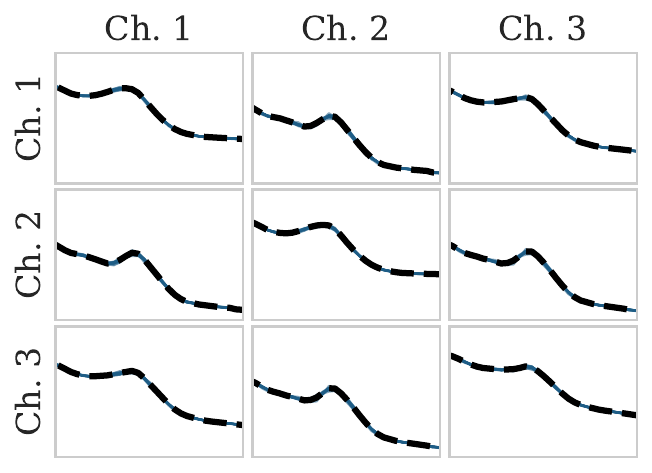}
        \caption{Alignment with $\filtersize=256$}
        \label{fig:bary_256}
    \end{subfigure}
    \caption{Illustration of the \stmethod{} method on sleep staging data. (a) The cross-PSD of the barycenter ($\boldsymbol{--}$) is computed from all the cross-PSD of the source domains. (b) The cross-PSDs of the source are aligned with the Monge mapping with a small filter size. (c) The cross-PSDs of the source are aligned with the Monge mapping with a big filter size. The bigger the filter size is, the more the cross-PSD are well aligned.}
    \label{fig:barycenter}
\end{figure}

\paragraph{Barycenter computation (\autoref{alg:train_time}, line 2)}
Since our data are supposed to be Gaussian-centered, the barycentric distribution is defined by the covariance $\overline\bSigma$, which can be expressed with the computed covariances.
The second step involves computing the barycentric distribution $\overline\bSigma$
% (see line 2 in \autoref{alg:train_time}) 
 from quantities computed in \autoref{eq:welch_stma}, \autoref{eq:welch_tma}, and \autoref{eq:scm_sma}.
For the \stmethod{} and \smethod{} methods and from Lemma~\ref{lemma:spatio_temp_bary} and Lemma~\ref{lemma:spat_bary}, the computation is performed with fixed-point iterations presented in \autoref{eq:bw_fixed_point_iterations}.
In practice, for iterative methods we compute an approximation of the barycenter
by initializing with the Euclidean mean over the
domain's covariance matrices, and performing one fixed-point iteration.
For the \tmethod{} method, the barycenter is obtained using the closed-form solution from Lemma~\ref{lemma:temp_bary}.
An example of barycenter for the \stmethod{} method is illustrated in \autoref{fig:bary_compute}, where the green line corresponds to the source cross-PSDs, and the black dotted line represents their barycenter.

% with the cross-PSDs $\overline\bp_{i,j} \in \bbR^{\filtersize}$ and the PSDs $\overline\bp_i \in \bbR^{\filtersize} $ in spatio-temporal and temporal cases respectively with $i$ and $j$ the index of the channels.

% For \stmethod{} and \smethod{}, there is no closed-form for the barycenter, but only a fixed-point algorithm. 
% In practice, we initialize the barycenter with the Euclidean mean over the domain's covariances. Then, fixed-point iterations are done until convergence.

% The \autoref{fig:bary_compute} shows an example of barycenter computation for \stmethod{} method. The green line corresponds to the source cross-PSD; the black dotted line is their barycenter.

{
    \paragraph{$\filtersize$-Monge maps computation (\autoref{alg:train_time},
    line 3)}
    The third step consists of computing the $\filtersize$-Monge maps
    $m_{f,k}$ for all $k\in\intset{n_d}$. % (see line 3 in
   % \autoref{alg:train_time}).
    The latter maps the $k^\text{th}$ source distribution to the barycenter.
    The formulas of $m_{f,k}$ are provided in
    Propositions~\ref{prop:spatio_temp_map},~\ref{prop:temp_map}
    and~\ref{prop:spat_map} for \stmethod{}, \tmethod and \smethod{}
    respectively and only involves quantities computed in the first two steps,
    \ie covariance matrices and barycenters. An illustration of the effect of
    applying the $\filtersize$-Monge maps is shown in \autoref{fig:bary_32} and
    \autoref{fig:bary_256}     for different filter sizes.
}

{
    \paragraph{Predictor training (\autoref{alg:train_time}, line 4)}
    The final step is to train a supervised predictor $h$, belonging to a function space
    $\Omega$, on the aligned source data.
    %(see line 4 in \autoref{alg:train_time}).
    The trained model is
    \begin{equation}
        \widehat{h} \in \argmin_{h \in \Omega} \;  \sum_{k=1}^{n_d} \mathcal{L} \left(\by_k,h\left(\widehat{m}_{\filtersize, k}(\bX_k)\right)\right)
    \end{equation}
    where $\bX_k$ is the multivariate signal of the $k^\text{th}$ domain and $\mathcal{L}$ is a supervised loss function.
}

% \paragraph{$\filtersize$-Monge maps estimation}
% Once this is done, the $\filtersize$-Monge maps $m_f$ are computed to align the covariance of each domain to the barycenter (see line 4 in \autoref{alg:train_time}). 
% Depending on the type of correlation assumed, filters can be learned and convolved with the signals using Lemma \ref{prop:spatio_temp_map} (\ie spatio-temporal correlation) or using Lemma \ref{prop:temp_map}  (\ie temporal correlation).
% Alternatively, if only spatial correlation is considered, matrix operation can be used, as explained in Lemma \ref{prop:spat_map}. 

% \paragraph{Estimator training}
% The last step is to train a defined estimator on the aligned data (see line 5 in \autoref{alg:train_time}). Since the source domains are labeled, one can fit the estimator $g$ using:
%   \begin{equation}
%     \widehat{h} \in \argmin_h \;  \sum_{k=1}^{n_d} \mathcal{L} \left(\by_k,h(\widehat{m}_{\filtersize, k}(\bX_k)\right)\;.
%     \label{eq:cmm}
% \end{equation}

% These pre-processing steps are independent of the final predictor and can be used with any method, including deep learning, shallow methods, or even coupled with other DA methods.

\subsubsection{Test-time}

% At test time, one has access to a new unlabeled
% target domain $(\bX_t)$. In this phase, there is no access to source domains, and we only need the barycenter computed during train-time and the estimator already trained. 

% The first step is to compute the $\filtersize$-Monge maps to align the target covariance to the barycenter (see line 1 in \autoref{alg:test_time}). 

% % Following the considered assumption on the signal, one can use cross-PSD (\ie \stmethod{}), the PSD (\ie \tmethod{}), or the spatial covariance (\ie \smethod{}).

% After aligning the signal, the last step is to predict the label with the trained predictor (see line 2 in \autoref{alg:test_time}):
% \begin{equation}
%     \widehat \by_t = \widehat h\left(\widehat{m}_{f,t}(\bX_t)\right)
% \end{equation}

% This method for test-time adaptation only requires a small number of unlabeled samples from the target domain to estimate its covariance. The computation of the barycenter and the training of the predictor are done during the training phase. Once the predictor is trained, retraining it during the test phase is unnecessary. However, it still allows for the final predictor to adjust to the spectral
% specificities (\ie \tmethod{}), spatial specificities (\ie \tmethod{}), or both (\ie \stmethod{}) of new domains by utilizing the Monge Alignment.

% For both \stmethod{} and \tmethod{}, the covariance can be estimated with the cross-PSD or the PSD, respectively. In the next section, we estimate these PSDs with the Welch method.

{
    At test-time, we have access to a new unlabeled target domain $\bX_t \in \bbR^{n_c \times n_\ell}$.
    Notably, during this phase, the source domains are no longer accessible, and
    we rely solely on both the source barycenter $\widehat{\overline\bSigma}$ and the trained predictor $\widehat h$, both computed at train time.
    The two first steps are the computations of the target covariance matrix and the $\filtersize$-Monge map $\widehat{m}_{f,t}$ from the target distribution to the barycenter (see lines 1 and 2 in \autoref{alg:test_time}).
    For the three methods, \stmethod{}, \tmethod{}, and \smethod{}, the computations employ the same equations as in the train-time phase.
    % \paragraph{Prediction}
    Next, we predict the label using the trained predictor on the aligned target
    data (see line 3 in \autoref{alg:test_time}) with
    %\begin{equation}
       $ \widehat \by_t = \widehat h\left(\widehat{m}_{f,t}(\bX_t)\right).$
   % \end{equation}
  %  \noindent
    It should be noted that the proposed test-time adaptation method requires only a small number of unlabeled samples from the target domain to estimate its covariance.
    Furthermore, the barycenter computation and predictor training occur during the training phase, eliminating the need for retraining during the test phase.
    Using \method{}, the trained predictor can still adapt to new temporal, spatial, or both shifts.
}

% \paragraph{Adaptability and Efficiency of \method{}}

% \subsection{Filter size and regularization}

% It is also important to note that in practice,
% the method allows for a size of normalization filter $\filtersize$ that is different
% (smaller) than the size $n_t$ of the signals. This consists in practice in
% estimating the PSD using Welch periodogram on signal windows of size $\filtersize\leq n_\ell$ that can be extracted from the raw signal or from already
% extracted fixed-sized source training samples.

% Indeed, if we use $F=T$, then the estimated
% average PSD can be perfectly adapted by the mapping, yet using many
% parameters can lead to overfitting, which can be limited using a smaller filter size $F\leq
% T$,
% It is interesting to note that the special case $F=1$ boils down
% to a scaling of the whole signal similar to what is done with a simple z-score operation. This means
% the filter size $\filtersize$ is a hyperparameter that can be tuned on the data.

% \subsubsection{Covariance estimation and complexities}
% \label{sec:spec}

\subsection{Computational aspects}

{
    We now discuss the computational aspects and numerial complexities  of the proposed method \stmethod{}, \tmethod{} and \smethod{}.
    The methods present interesting trade-offs between the costs of calculation and the shifts considered (spatial, temporal, or both).
    Indeed, the more shifts considered, the higher the calculation cost and conversely.
    This can motivate the practitioner to use one method rather than another, depending on the dimensions of the data and the observed shifts between domains.
    Then, the choice of the filter size $f$ is discussed.
}

\subsubsection{Numerical complexity}
\label{sec:complexity}
% \begin{table*}
%     \centering
%     \footnotesize
%     \begin{tabular}{|c|c|c|c|}
%     \hline
%         Method & Spatial & Temporal & spatio-temporal \\
%     \hline\hline
%         Covariance/PSD & $\mathcal{O}\left(\sum_k^{n_d}n_k n_\ell n_c^2\right)$ & $\mathcal{O}\left(\sum_k^{n_d} n_k n_c n_\ell \log{\filtersize}\right)$  & $\mathcal{O}\left(\sum_k^{n_d} n_k n_c^2 n_\ell \log{\filtersize}\right)$\\
%     \hline
%         Barycenter & $\mathcal{O}\left(n_dn_c^3\right)$ (one iter.) & $\mathcal{O}\left(n_c n_d \filtersize\right)$ & $\mathcal{O}\left(n_d n_c \filtersize^3\right)$ (one iter.) \\
%     \hline
%         Map computation & $\mathcal{O}\left(n_d n_c^3\right)$ & $\mathcal{O}\left(n_c n_d \filtersize\right)$ & $\mathcal{O}\left(n_d n_c \filtersize^3\right)$\\
%     \hline
%         Convolution/Mapping & $\mathcal{O}\left(\sum_k^{n_d} n_k n_c^2 n_\ell \right)$ &  $\mathcal{O}\left(\sum_k^{n_d} n_k n_c n_\ell \log \filtersize\right)$ & $\mathcal{O}\left(\sum_k^{n_d} n_k n_c^2 n_\ell \log \filtersize\right)$ \\
%     \hline
%     \end{tabular}
%     \caption{Computational complexity comparison of spatial, temporal, and spatio-temporal analyses}
%     \label{tab:complexity}
% \end{table*}

\begin{table*}
    \centering
    \footnotesize
    \begin{tabular}{|c|c|c|c|}
        \hline
            Step & \stmethod{} & \tmethod{} & \smethod{} \\
        \hline\hline
            Covariance matrix computation & $\mathcal{O}\left(n_d n_c^2 n_\ell \filtersize \log{\filtersize}\right)$ & $\mathcal{O}\left(n_d n_c n_\ell \filtersize \log{\filtersize}\right)$  & $\mathcal{O}\left(n_d n_c^2 n_\ell\right)$\\
        \hline
            Barycenter computation & $\mathcal{O}\left(n_d n_c^3 \filtersize\right)$ (per iter.) & $\mathcal{O}\left(n_d n_c \filtersize\right)$ & $\mathcal{O}\left(n_dn_c^3\right)$ (per iter.) \\
        \hline
            Mapping computation & $\mathcal{O}\left(n_d (n_c^3 \filtersize + n_c^2 f \log f)\right)$ & $\mathcal{O}\left(n_d n_c \filtersize \log f\right)$ & $\mathcal{O}\left(n_d n_c^3\right)$\\
        \hline
            Mapping application & $\mathcal{O}\left(n_d n_c^2 n_\ell \filtersize\right)$ &  $\mathcal{O}\left(n_d n_c n_\ell  \filtersize\right)$ & $\mathcal{O}\left(n_d n_c^2 n_\ell \right)$ \\
        \hline
    \end{tabular}
    \caption{
        {
            Computational complexity comparison between \stmethod{} (spatio-temporal), \tmethod{} (temporal), and \smethod{} (spatial).
        The complexity is expressed in terms of four key parameters: the
        number of domains $n_d$; the number of sensors $n_c$; the length
        of the signals $n_\ell$ and the filter size $\filtersize$.
            % at train-time.
            % At test-time, only the mapping computation and application are performed, and $n_d$ becomes the number of target domains.
        }}
    \label{tab:complexity}
\end{table*}

{
    The computational complexity of the four main operations at train and test
    times are studied below. 
    % : covariance matrix computation, barycenter computation,
    % $\filtersize$-Monge maps computation, and $\filtersize$-Monge maps
    % application.
    The complexities are presented in \autoref{tab:complexity} and depend on the number of sensors ($n_c$), signal length
    ($n_\ell$), and filter size ($\filtersize$), and vary depending on the method
    (\stmethod{}, \smethod{}, or \tmethod{}).
    %These complexities .
    Note that all methods are linear in the number of domains and signal length.
    \smethod{} does not consider temporal correlations, so its complexity is
    independent of the filter size $\filtersize$ (equivalent to $\filtersize=1$), contrary to \stmethod{} and
    \tmethod{}.
    In the same manner, \tmethod{} does not consider spatial correlations, so
    its complexity is linear in the number of sensors, contrary to \stmethod{} and
    \smethod{}, which have cubic complexities. It should be noted that all
    operations are standard linear algebra and can be performed on modern hardware (CPU and GPU).
   % The selection of $\filtersize$ reduces computation time for both \stmethod{} and \tmethod{}.
    A detailed explanation of the complexities is provided hereafter.
}

{
    \paragraph{Covariance matrix computation}
    \stmethod{} and \tmethod{} use the Welch estimator with Fast Fourier Transform (FFT), which has a complexity of $\mathcal{O}(\filtersize \log \filtersize)$.
    \tmethod{} method has a linear complexity in the number of sensors, while the \stmethod{} and \smethod{} methods are quadratic since they consider correlations between sensors. 
}

{
    \paragraph{Barycenter computation}
    This operation applies directly to the covariance matrices, making it independent of the signal length. 
    \tmethod{} has a closed-form barycenter and linear complexity in the filter size and number of sensors, making it interesting for data with many sensors.
    \stmethod{} also has a complexity per iteration that is linear in the filter size.
    \stmethod{} and \smethod{} have a  cubic complexity in the sensor number because of the computations of Singular Value Decomposition (SVDs) to perform square roots and inverse square roots.
}

{
    \paragraph{$\filtersize$-Monge maps computation}
    This operation also applies directly to the covariance matrices, making it independent of the signal length. 
    In the worst case, the complexity per iteration of \stmethod{} is $f\log f$ (filter size).
    \stmethod{} and \smethod{} have a cubic complexity in the sensor number because of the computations of SVDs.
    \tmethod{} has a linear complexity in the number of sensors, making it again appealing for data with many sensors.
}

{
    \paragraph{$\filtersize$-Monge maps application}
    All the three methods are linear in the signal length.
    \stmethod{} and \smethod{} involve a quadratic number of operations with respect to the  number of sensors contrary to \tmethod{} which has a linear complexity.
}

% From an implementation point of view, one can use the Fast Fourier Transform (FFT) to compute the convolution (for large
% filters) or the direct convolution, which both have very
% efficient implementation on modern hardware (CPU and GPU).

\subsubsection{Filter size selection}

The proposed method introduces $\filtersize$, the filter size.
This key hyper-parameter serves a dual role. 
Primarily, it contributes to reducing computation time; a benefit clearly demonstrated in \autoref{tab:complexity} and  \autoref{sec:complexity}.
Secondly, the filter size dictates the precision with which the PSDs of the domains are mapped to the barycenter.
As visualized in \autoref{fig:psd_mapping} and \autoref{fig:barycenter}, a larger filter size yields a more accurate mapping, while a smaller size only subtly adjusts the PSDs.
This flexibility enables a balance between perfectly aligning the PSDs, which might compromise unique class features, and making sufficient adjustments to reduce noise and domain-specific variability while preserving class-specific characteristics.

\subsection{Concentration bounds of the Monge Alignment}

In this subsection, we propose concentration bounds for the estimations of \stmethod{}, \tmethod{}, and \smethod{} mappings.
These theorems provide a rigorous framework for understanding the reliability and precision of \method{} methods.
Estimating OT plans is challenging due to the curse of dimensionality, with OT estimators generally decaying at a rate of $O(n^{-1/d})$, where $n$ is the number of observations and $d$ is the data dimension \cite{fournier2015rate}. However, \cite{flamary2020concentration} propose a concentration bound for linear Monge mapping estimation with a faster convergence rate of $O(n^{-1/2})$, \ie independent of the data dimension.
This suggests greater accuracy and robustness in high-dimensional settings.
Our proposed bounds account for two key aspects: 
mapping to the Wasserstein barycenter and using PSD computation with the Welch estimator.
These bounds differ from \cite{flamary2020concentration}, which maps to a given empirical covariance matrix.
In particular, we leverage \cite{lamperski2023nonasymptotic} to establish these
bounds, highlighting an induced bias in PSD computation due to the Welch
estimator.
{
To introduce the concentration bound, we first define from
\cite{lamperski2023nonasymptotic} the spatial correlation of delay $\ell$,
denoted $\bR_\ell$, for signal following the Assumption~\ref{assu:spat_temp}, as}
{
\begin{equation}
    \bR_\ell
    =
    \begin{pmatrix}
        (\bSigma_{1, 1})_{1, \ell} & \dots & (\bSigma_{1, n_c})_{1, \ell} \\
        \dots & \dots & \dots \\
        (\bSigma_{n_c, 1})_{1, \ell} & \dots & (\bSigma_{n_c, n_c})_{1, \ell}
    \end{pmatrix} \in \cS_{n_c}^{++}
\end{equation}
}
{For signals following Assumption~\ref{assu:temp} only the diagonal, $(r_c)_l =
(\bSigma_{c, c})_{1, \ell}$ is necessary.}

% The continuous power spectral density is given by:
% \begin{align*}
%     % \bR[\ell] &= \bbE\left[ \bX^\trans[i+\ell]\left(\bX^\trans[i]\right)^\trans \right] \in \bbR^{n_c \times n_c}\;,\\
%     \mathbf{\Phi}(s) &= \sum_{\ell=-\infty}^\infty e^{-j2\pi s \ell }\bR[\ell] \in \bbR^{n_c \times n_c}\;.
% \end{align*}

\subsubsection{\stmethod{} concentration bound}

The next theorem introduces the general concentration bound, \ie corresponding to \stmethod{}.

\begin{theorem}[\stmethod{} concentration bound]
    \label{the:concentration_spatiotemp}
    \sloppy Let $\bX_t \in \bbR^{n_c \times n_\ell}$ be a realization of $\Vectr(\bX) \sim \cN(\mathbf{0}, \bSigma_t)$.
    For $k \in \intset{K}$, let $\bX_k \in \bbR^{n_c \times n_\ell}$ be a realization of $\Vectr(\bX_k) \sim \cN(\mathbf{0}, \bSigma_{k})$.
    Let us assume that $\bSigma_t$ and $\bSigma_k$ follow the Assumption~\ref{assu:spat_temp}, \ie  $\bSigma_t = \bF \bU \bQ_t \bU^\trans \bF\hermconj$ and $\bSigma_k = \bF \bU \bQ_k \bU^\trans \bF\hermconj$, that
    $\Vert \bR_\ell \Vert_2 \leq \gamma \rho^{\vert \ell \vert}$ with $\gamma > 0$ and $\rho \in [0, 1)$
    {and that only fixed point iteration of the barycenter is done}. Then
    there exist numerical constants $c_k > 0$, for all $k$, and $C > 0$, independent from $n_\ell$, $n_c$, {$n_d$ and $\filtersize$}, such that with probability greater than $1-\delta$, we have
    \vspace{0.3cm}
    \begin{multline*}
         \Vert \widehat{\bA} - \bA \Vert \lesssim \left(C + \frac{1}{n_d}\sum_{k=1}^{n_d} c_k \right) \left(
         \eqnmarkbox[orange]{bias}
            {\delta 2\sum_{i=0}^{f-1}\frac{i}{n_\ell} \rho^i + \frac{2\delta \rho^f}{1-\rho}}\right.
             \\ + 
            \left.\eqnmarkbox[cyan]{var}{2  \widetilde{\bQ} 
        \max\left\{ \frac{5}{2\frac{n_\ell}{f}} \log \left(5 f^2 \left(\frac{4 \times 10^{2n_c}}{\delta}\right)^{32}\right), \sqrt{\frac{5}{2\frac{n_\ell}{f}}\log \left(5 f^2 \left(\frac{4 \times 10^{2n_c}}{\delta}\right)^{32}\right)}\right\}}\right) \;,
    \end{multline*}
    {with $\widetilde{\bQ} = \max_{d \in \{1, \dots, n_d, t\}}{\Vert \bQ_d \Vert}_\infty$}.
    \annotate[yshift=0.5em, xshift=1.5em]{above, right}{bias}{Bias}
    \annotate[yshift=-0.5em]{below, right}{var}{Variance}
    % \annotatetwo[yshift=1em]{below, right}{n1}{n2}{Number of window}
\end{theorem}

\begin{figure}[t]
    \begin{minipage}{0.67\textwidth}
        \centering
        \includegraphics[height=5cm]{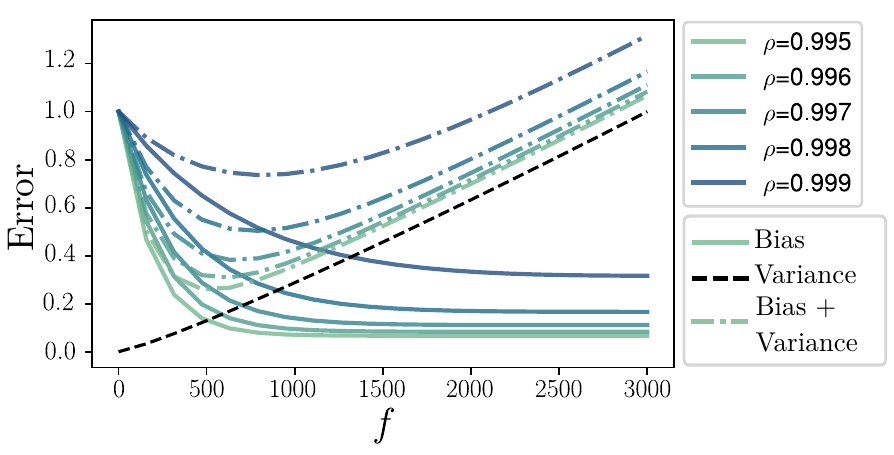}
    \end{minipage}
    \begin{minipage}{0.3\textwidth}
            \caption{
        % Left: assumption of bounded temporal correlation ($\Vert \bR[l]\Vert_2 \leq \gamma \rho^{\vert l \vert}$).
        bias-variance estimation tradeoff versus the filter size $\filtersize$.
        % Both are computed for $n_\ell = 3000$ and different $\rho$ values. 
        % The bigger the $\rho$, the larger the correlation bound but the bigger the bias.
        For a given $\rho$, increasing the filter size reduces the bias, but at the cost of increased variance.
        A single optimal filter size range balances bias and variance, yielding a good tradeoff.
        }
        \label{fig:corrbias}
    \end{minipage}
\end{figure}

\noindent
{We made the choice to limit the theoretical analysis to only one iteration of the barycenter, allowing to provide this result. In practice, the barycenter converges fast, leading to good results after one iteration, and we did not observe performance gains with more iterations in our experiments.}
In contrast to the concentration bound proposed by \cite{flamary2020concentration}, this bound includes a bias term resulting from the estimation of cross-PSDs using the Welch method. 
This bias depends on size of the filter $f$ the temporal correlation of the signal: higher correlation (large $\rho$) leads to greater bias, as shown on the left in \autoref{fig:corrbias}.
However, the variance has a classical bound in $\mathcal{O}(\sqrt{\filtersize/n_\ell})$.
Hence, the filter size can be adjusted to control the bias-variance trade-off.
\autoref{fig:corrbias} shows this trade-off with a dotted line, highlighting the optimal filter size.

\subsubsection{\tmethod{} concentration bound}

{
\tmethod{} exhibits a similar concentration bound, \ie with a bias-variance tradeoff and is exposed here-after.
}
\begin{theorem}[\tmethod{} concentration bound]
    \label{the:concentration_temp}
    \sloppy Let $\bX = [\bx_1, \dots, \bx_{n_c}]^\trans \in \bbR^{n_c \times n_\ell}$ be a realization of $\Vectr(\bX) \sim \cN(\mathbf{0}, \bSigma)$.
    For $k \in \intset{K}$, let $\bX_k \in \bbR^{n_c \times n_\ell}$ be a realization of $\Vectr(\bX_k) \sim \cN(\mathbf{0}, \bSigma_{k})$.
    We assume that $\bSigma$ and $\bSigma_k$ follow Assumption~\ref{assu:temp}, \ie $\bSigma = \bF \diag(\bq_{1}, \dots, \bq_{n_c}) \bF\hermconj$ and $\bSigma_k = \bF \diag(\bq_{1}^{k}, \dots, \bq_{n_c}^{k}) \bF\hermconj$, and for every $c \in \intset{n_c}$ $(r_c)_\ell \leq \gamma \rho^{\vert \ell \vert}$ with $\delta>0$ and $\rho \in [0, 1)$.
    For every $c \in \intset{n_c}$, we denote by
    $\bp_c = g_{\filtersize}(\bq_c)$,
    $\overline\bp_c = g_{\filtersize}(\overline\bq_c)$, and $\bp_{c, k} = g_{\filtersize}(\bq_{c, k})$.
    % $\bh_i = \frac{1}{\sqrt{\filtersize}} \bF_{\filtersize}\hermconj \left({\overline{\bp}_i}^{\odot\frac{1}{2}}\odot {\bp_i}^{\odot-\frac{1}{2}} \right)$.
    Then there exists numerical constants $c_{c,k}' > 0$, for all $k$ and $C_c' > 0$ independent from $n_\ell$, $n_c$, {$n_d$ and $\filtersize$}, such that with probability greater than $1-\delta$, we have
    \vspace{1.5mm}
    \begin{multline*}
         \Vert \widehat{\bA} - \bA \Vert 
         \lesssim \max_c \left(C'_c + \sum_{k=1}^{n_d} c_{c,k}' \right)\left(
          \eqnmarkbox[orange]{bias}{\delta 2\sum_{i=0}^{f-1}\frac{i}{n_\ell} \rho^i + \frac{2\delta \rho^f}{1-\rho}} \right. 
          \\+
          \left.\eqnmarkbox[cyan]{var}{ 2  \widetilde{\bp}_c 
            \max\left\{ \frac{5}{2\frac{n_\ell}{f}} \log \left(5 f^2 \left(\frac{4 \times 10^{2}}{\delta}\right)^{32}\right), \sqrt{\frac{5}{2\frac{n_\ell}{f}} \log \left(5 f^2 \left(\frac{4 \times 10^{2}}{\delta}\right)^{32}\right)}\right\} }\right) \;.
    \end{multline*}
     {with $\widetilde{\bp}_c = \max_{d \in \{1, \dots, n_d, t\}}{\Vert \bp_{c,d}\Vert}_\infty$. }
    \annotate[yshift=0.5em, xshift=2em]{above, right}{bias}{Bias}
    \annotate[yshift=-0.5em]{below, right}{var}{Variance}
\end{theorem}
{
To the best of our knowledge, this is the first concentration bound of the
method proposed by \cite{gnassounou2023convolutional}. This bound includes bias
and variance terms as in \stmethod{} concentration bound. The bound has similar
behaviors but the independence of the channels leads to a smaller variance.
}
\subsubsection{\smethod{} concentration bound}

{
Contrary to \stmethod{} and \tmethod{}, \smethod{} has a concentration bound with a null bias.
This makes it particularly suitable for short and highly correlated in time signals.
Indeed, \smethod{} ensures accurate estimations without the bias introduced by the temporal correlation. 
}
\begin{samepage}
\begin{theorem}[\smethod{} concentration bound]
    \label{the:concentration_spatio}
    \sloppy Let $\bX_t \in \bbR^{n_c \times n_\ell}$ be a realization of $\Vectr(\bX_t) \sim \cN(\mathbf{0}, \bSigma_t)$.
    For $k \in \intset{K}$, let $\bX_k \in \bbR^{n_c \times n_\ell}$ be a realization of $\Vectr(\bX_k) \sim \cN(\mathbf{0}, \bSigma_{k})$.
    We assume that $\bSigma_t$ and $\bSigma_k$ follow Assumption~\ref{assu:spat}, \ie  $\bSigma_t = \bF \bU \bQ_t \bU^\trans \bF\hermconj$ and  $\bSigma_k = \bF \bU \bQ_k \bU^\trans \bF\hermconj$.
    Then there exist numerical constants $c_k > 0$, for all $k$ and $C > 0$ independent from $n_\ell$, $n_c$, {$n_d$ and $\filtersize$}, such that with probability greater than $1-\delta$, we have
    \begin{multline*}
         \Vert \widehat{\bA} - \bA \Vert \lesssim \eqnmarkbox[cyan]{var}{\frac{1}{n_d}\sum_{k=1}^{n_d} c_k\Vert \bSigma_k \Vert \max\left(\sqrt{\frac{\br(\bSigma_k)}{n_\ell} }, \frac{\br(\bSigma_k)}{n_\ell}, \sqrt{\frac{-\ln{\delta}}{n_\ell} }, \frac{-\ln{\delta}}{n_c}\right) }\\   \eqnmarkbox[cyan]{var2}{+ C \Vert \bSigma_t \Vert \max\left(\sqrt{\frac{\br(\bSigma_t)}{n_\ell} }, \frac{\br(\bSigma_t)}{n_\ell}, \sqrt{\frac{-\ln{\delta}}{n_\ell} }, \frac{-\ln{\delta}}{n_\ell}\right)}\;.
    \end{multline*}
    % \annotate[yshift=0.5em]{above, right}{bias}{Bias}
    \annotate[yshift=-0.5em]{below, right}{var2}{Variance}
    % \annotatetwo[yshift=1em]{below, right}{n1}{n2}{Number of window}
\end{theorem}
\end{samepage}

% This new concentration bound shows that the length of the applied window is essential in determining the trade-off between bias and variance in the resulting PSD. When 
% $F=T$, bias is absent in the estimation, but this configuration exhibits high variance.
% 
% \newpage

\section{Experimental results}
\label{sec:exp}

% \ac{
%     A mettre quelque part dans cette seection (ou bien à enlever):
%     The regularization parameter ($\lambda$) is used to shrink the covariance of the identity to control its condition number.
%     An illustrative example of the effect of this parameter is provided in the appendix, offering further insight into its role and impact on the proposed method.
% }

This section evaluates \method{} on different datasets.
% First, we propose a new custom MNIST dataset with multi-directional blur.
% We apply \method{} on these 2D signals and show its efficiency to compensate for the blur.
% Then, \method{} is tested on two biosignals classification tasks. 
% The first experiment extends results on sleep staging proposed in \cite{gnassounou2023convolutional} by using multivariate signals (\ie several channels) and thus exploiting the spatial information. The second experiment focuses on Brain-Computer Interface (BCI) Motor Imagery tasks, where spatial information is crucial to classify.
\method{} is first tested on two biosignals classification tasks. 
The first experiment extends results on sleep staging proposed in \cite{gnassounou2023convolutional} by using multivariate signals (\ie several channels) and thus exploiting the spatial information. The second experiment focuses on Brain-Computer Interface (BCI) Motor Imagery tasks, where spatial information is crucial to classify.
Then, we extend our method to 2D signals (\ie images) on a new custom MNIST
dataset where each domain corresponds to a directional blur.
We apply \method{} on this new dataset and show its efficiency to compensate for the blur.

\subsection{Biosignal tasks}

In this section, we investigate the impact of \method{} on two critical biosignal tasks: Sleep stage classification and Brain-Computer Interface (BCI). Each dataset is introduced alongside the alignment methodology proposed for these tasks.

\subsubsection{Biosignal Datasets}
\label{sec:dataset}
\paragraph{Sleep Staging Datasets} We utilize four publicly available datasets: MASS \cite{MASS}, HomePAP \cite{rosen2012homepap}, CHAT \cite{marcus2013CHAT}, and ABC \cite{jp2018ABC}, accessible via the National Sleep Research Resource \cite{zhang2018NSRR}. Sleep staging is performed using 7-channel EEG signals across all datasets. The EEG channels considered are F3, F4, C3, C4, O1, O2, and A2, referenced to FPz. Each night's data is segmented into 30-second samples with a sampling frequency of \( f_s = 100 \) Hz.

\paragraph{Brain-Computer Interface Datasets} We employ five publicly available datasets: BCI Competition IV \cite{BCI2012}, Weibo2014 \cite{Weibo2014}, PhysionetMI \cite{PhysionetMI}, Cho2017 \cite{cho2017}, and Schirrmeister2017 \cite{schirrmeister2017deep}, accessible through MOABB \cite{Aristimunha2023moabb}.
These datasets consists of two classes motor imagery tasks involving right-hand and left-hand movements.
We utilize a common set of 22 channels across the datasets. The length of time series varies across datasets, and following \cite{xu2020bcicross}, we extract a uniform segment of 3 seconds from the middle of each trial for improved consistency with a sampling frequency of $f_s = 128$ Hz.

\subsubsection{Spatial correlation alignment} 
% These datasets have been studied a lot in the past years \cite{Perslev2021Usleep, chambon2018deep, wimpff2024calibrationfree}. Many studies show the necessity to compensate for the shift between the domain in biosignals data \cite{xu2020bcicross, wimpff2024calibrationfree, Eldele2021atten}.
% It is possible to reduce this shift with data alignment before training and testing. This method allows test-time adaptation where there is no need for the source data to predict on the target domain, named test-time DA.
% For a multivariate signal $\bX_d \in \bbR^{n_c \times n_\ell}$ from the domain $d$, the alignment is given by the following operation:
% \begin{equation}
%     m_{\text{RA}}(X_d) = \bM_d^{-\frac{1}{2}} \bX_d \;,
% \end{equation}
% where $\bM_d \in \bbR^{n_c \times n_c}$ is the Riemannian mean of the covariances of all samples from the domain $d$ where $d$ can mean the source or the target domain.

%  This method, called Riemannian Alignment (RA), is mainly used in BCI tasks \cite{wimpff2024calibrationfree, xu2020bcicross, junqueira2024systematic} and is close to the \smethod{} method since both use only spatial covariances. In the following experiments, we compare RA with our proposed method for sleep staging and BCI application.

{
The selected datasets have been extensively studied in recent years \cite{chambon2018deep, Perslev2021Usleep, wimpff2024calibrationfree}. 
In particular, several studies have highlighted the critical need to address the domain shifts in biosignal data \cite{xu2020bcicross, Eldele2021atten, junqueira2024systematic}.
To compensate for these shifts, they proposed Riemannian Alignment (RA), which is a test-time DA method tailored for spatial shifts.
For a multivariate signal $\bX_d \in \bbR^{n_c \times n_\ell}$ from the domain $d$, this alignment is given by the following operation:
\begin{equation}
    m_{\text{RA}}(X_d) = \bM_d^{-\frac{1}{2}} \bX_d \;,
\end{equation}
where $\bM_d \in \cS_{n_c}^{++}$ is the Riemannian mean of the covariance matrices of all samples from the domain $d$.
In the following experiments, we compare RA with \method{} for sleep staging and BCI motor imagery.
}
\subsubsection{Deep learning architectures and experimental setup}

{
In our experiments, we used two different architectures and setups for sleep staging and BCI. For sleep staging, we referred to the approach described in \cite{chambon2018deep}, and for BCI, we followed the setup proposed by \cite{schirrmeister2017deep} as implemented in MOABB \cite{Aristimunha2023moabb}. In the upcoming section, we provide a detailed description of these two setups.
}
\paragraph{Architecture}
{
Many neural network architectures dedicated to sleep staging have been proposed \cite{Supratak_2017, Perslev2021Usleep, Eldele2021atten, xsleepnet}.
In the following, we choose to focus on the architecture proposed by
\cite{chambon2018deep} that is an end-to-end neural network proposed to deal with
multivariate time series
and is composed of two convolutional layers with non-linear activation functions.

For BCI, different neural network architectures have been developed specifically \cite{schirrmeister2017deep, lawhern2018eegnet}.
We focus on ShallowFBCSPNet \cite{schirrmeister2017deep}, which is an end-to-end neural network proposed to deal with multivariate time series based on convolutional layers and filter bank common spatial patterns (FBCSP).
Both architectures are implemented in \texttt{braindecode} package \cite{schirrmeister2017deep}.
}

\paragraph{Training setup}

{ The training parameters follows the reference implementations of
\cite{chambon2018deep} and \cite{schirrmeister2017deep} detailed below for reproducibility.
For sleep staging experiment, we use the Adam optimizer with a learning rate of $10^{-3}$, $\beta_1 = 0.9$, and $\beta_2 = 0.999$.
The batch size is set to $128$, and the early stopping is
done on a validation set corresponding to {$20\%$ of the subjects in the training set} with a 
patience of $15$ epochs. We optimize the cross-entropy with
class weight for all methods, which amounts to optimizing for balanced accuracy (BACC). Each experiment is done ten times with a different seed.

For BCI, we use the Adam optimizer with a cosine annealing learning rate scheduler
starting at $6.25 \times 10^{-4}$. The batch size is set to $128$, and the
training is stopped after 200 epochs. No validation set is used. We optimize the
cross-entropy for all methods. We report the average accuracy score (ACC)
across ten different seeds.
}

\paragraph{Filter size and Welch's parameters}

\label{sec:sensitivity}
{
\method{} requires tuning the filter size $\filtersize$ parameters.
Yet, experiments show that this parameter is not critical, as good performances are observed over a large range of values. We provide a sensitivity analysis of the performance for different parameters in \autoref{fig:sensi} for sleep staging (left) and BCI (right). It shows that the
value $\filtersize=256$ is a good trade-off for the sleep application while a value of $\filtersize=8$ is better for BCI.
These parameter values are used in what follows.
For the Welch method, we use a Hann window \cite{blackman1958hann} and an overlap of $\frac{\filtersize}{2}$.

\begin{figure}[t]
    \begin{subfigure}{0.49\textwidth}
        \includegraphics[width=\linewidth]{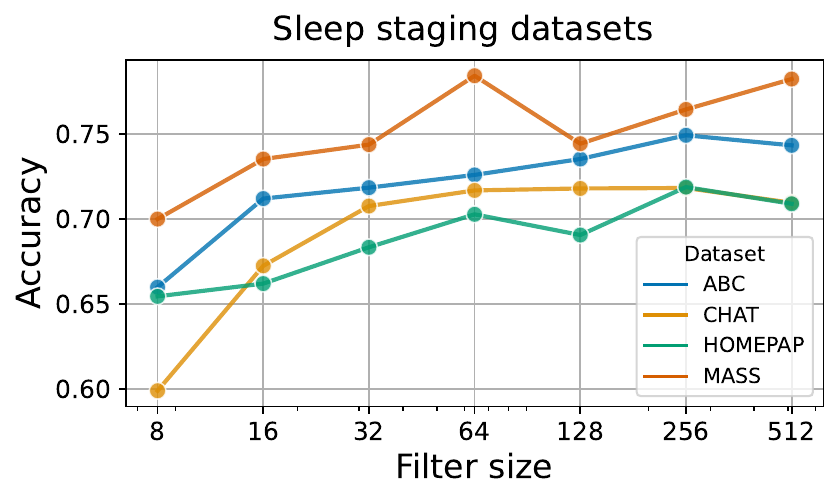}
        % \caption{Sleep staging dataset}
        \label{fig:sensi_sleep}
    \end{subfigure}
    \begin{subfigure}{0.49\textwidth}
        \includegraphics[width=\linewidth]{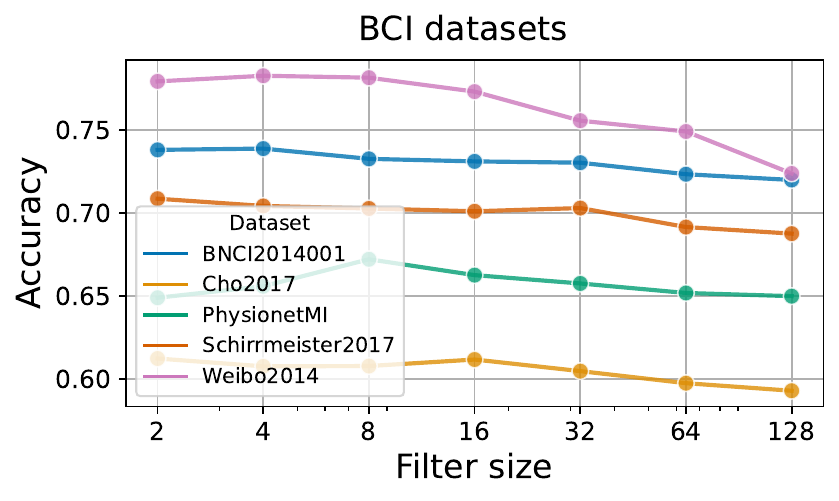}
        % \caption{BCI datasets}
        \label{fig:sensi_bci}
    \end{subfigure}\vspace{-5mm}
    \caption{Evolution of the balanced accuracy score for a leave-one-dataset-out experiment for different values of filter size $\filtersize$.}
    \label{fig:sensi}
\end{figure}
}

\subsection{Application to sleep staging}

{
In our investigation, we build on the findings of
\cite{gnassounou2023convolutional} who showed the benefits of
\tmethod{} for sleep staging when using only two sensors.
Here, we incorporate more spatial information by using 7 EEG channels, as detailed in Section~\ref{sec:dataset}.
This multidimensional dataset offers a more comprehensive perspective, thereby demonstrating the utility of \stmethod{} and \smethod{}. 
In what follows, one subject is considered as one domain.
}

\subsubsection{Comparison between different alignments}

% In this section, we compare our new spatio-temporal normalization with Riemannian alignment, which has never been used in sleep staging since usual experiments are done on 1 or 2 channels (\ie no spatial information) \cite{Supratak_2017, Perslev2021Usleep}. 
% We perform a leave-one-dataset-out  evaluation (4-fold across datasets adaptation). The results are given in the \autoref{fig:perf_sleep}.

% The blue box in the graph indicates that when using standard z-score normalization techniques, there can be difficulties in adapting between different datasets. 
% None of the balanced accuracies reach more than 66\%, and in some cases, like for the CHAT dataset, the accuracy drops below 50\%. 
% The plot illustrates that using alignment improves the accuracy score on an unseen dataset, as shown for both RA and \stmethod{}, where the accuracy is consistently above 70\%. 
% However, the proposed method outperforms RA by 2-3\%, giving greater consistency across subjects. 
% This behavior was intended because temporal information is crucial for sleep classification, while RA only uses spatial alignment.

% This comparison shows the benefit of using spatiotemporal alignment: better accuracy and lower variance. This lower variance indicates that the underperforming subjects without alignment are often the ones that improve the most. In the following section, we will focus on these underperforming subjects.

{
Historically, sleep staging has relied on neural networks processing raw data \cite{Supratak_2017}. \cite{chambon2018deep} improved this by applying z-score normalization to 30-second windows, removing local trends.
More recently, \cite{apicella2023effects} introduced session-wide normalization to eliminate global trends.
Among these, window-based normalization, denoted as “No Align”, has proven most
effective in \cite{gnassounou2023convolutional} and will be used as baseline in
the following.
In our work, we also compare \method{} with RA to evaluate performance comprehensively.
}

\begin{figure}[t]
    \centering
    \includegraphics[width=0.9\linewidth]{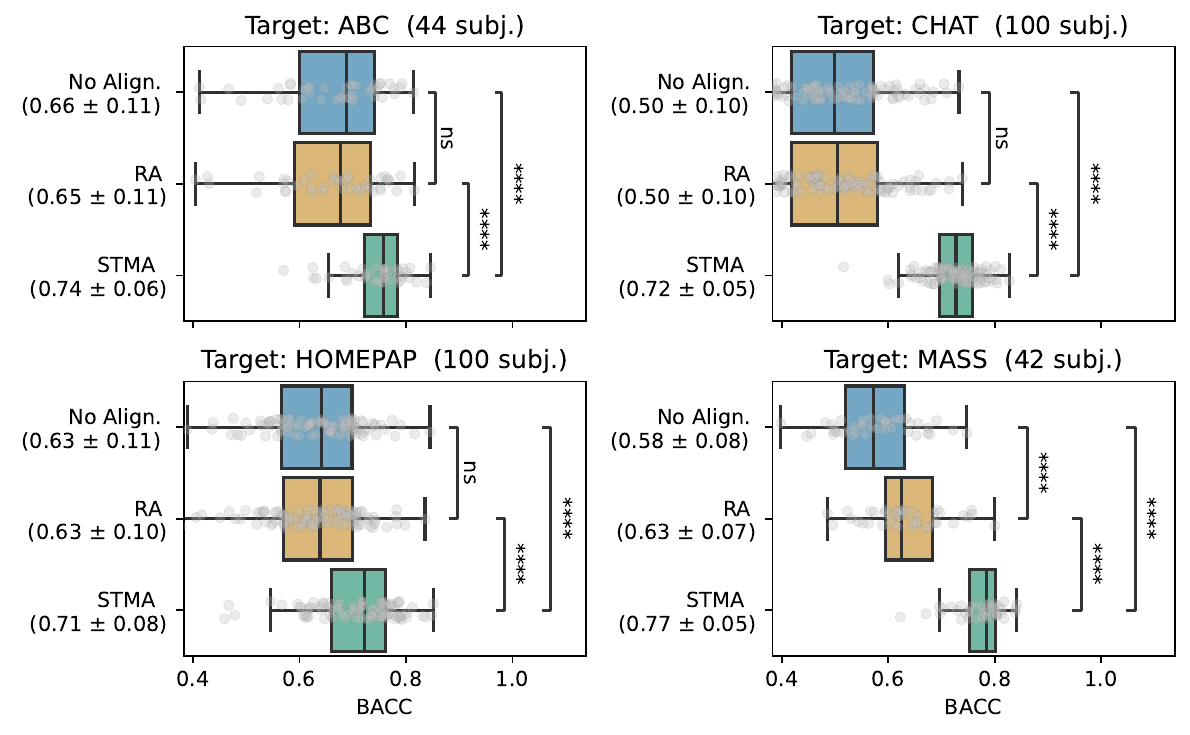}
    \caption{Balanced accuracy for sleep stage classification without alignment (blue), with RA (yellow), and with \stmethod{} (green) for different datasets in the target domain. Each dot represents one subject. \stmethod{} outperforms alternative methods for each target dataset. The number of stars illustrates the level of significance with a Wilcoxon test (ns: non-significative, $**$: $10^{-2}$, $***$: $10^{-3}$, or $****$: $10^{-4}$).
    }
    \label{fig:perf_sleep}
\end{figure}

Using a leave-one-dataset-out evaluation (4-fold across dataset adaptation), results in \autoref{fig:perf_sleep}
show that standard z-score normalization struggles with dataset adaptation, with accuracies below 66\%, sometimes dropping below 50\% (e.g., CHAT dataset).
{Spatio-temporal alignment improves accuracy, with \stmethod{} consistently above 70\% and even reaching an improvement of \~20\% for MASS and CHAT. On the other hand, RA stagnates for three datasets (\ie ABC, CHAT and HOMEPAP) and increase the score for only MASS by 5\%.}
This underscores the importance of temporal information in sleep classification, beyond RA’s spatial focus.
The experiment highlights the benefits of spatiotemporal alignment: higher accuracy and lower variance, suggesting that underperforming subjects improve the most.
The next section will explore these subjects further.

\subsubsection{Study of performance on low-performing domains}

\begin{figure}[t]
    \centering
    \includegraphics[width=0.95\linewidth]{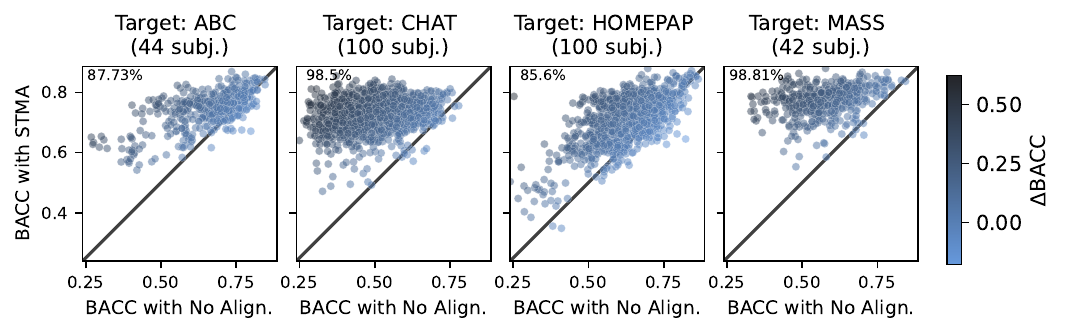}
    \caption{Balanced accuracy for sleep staging with \stmethod{} and without alignment. 
    Each dot represents a subject. The size of the difference between the scores is directly proportional to the darkness of the dot. Darker dots are located on the left (\ie lower-performing subjects). The percentage of the dots above and below the black diagonal line is given on the bottom left.}
    \label{fig:scatter_sleep}
\end{figure}

In this section, we evaluate the performance of individuals to identify the subjects that benefit from the most significant improvements. In the \autoref{fig:scatter_sleep}, we propose to compare the accuracy without alignment ($x$-axis) and with spatio-temporal alignment ($y$-axis). 
A subject (\ie a dot) above the line $y=x$ means that the alignment increases
its score. For each adaptation, more than 85\% of the dots are above the line,
indicating a systematic increase in alignment for each domain. This number is even close
to 100\% for CHAT and MASS datasets. 
The rate of increase can vary for each subject. We color the dots by the $\Delta$BACC, representing the difference between the balanced accuracy with and without alignment. The darker the dots are, the better the $\Delta$BACC is. 
On the scatter plot, the more low-performing subjects (\ie dots on the left) increase more than the other subjects.

To highlight our findings, we propose the \autoref{tab:subject} that reports the
mean and variance of the $\Delta$BACC. 
Furthermore, we also report the $\Delta$BACC@20 metric, proposed in \cite{gnassounou2023convolutional}, which is the variation of balanced accuracy for the 20\% lowest-performing subjects (\ie the subjects with the lowest scores without alignment). 
For each adaptation, the gain is almost double for the low-performing subjects compared to all subjects.
Hence, \stmethod{} is valuable for medical applications where a low failure rate is more important than a high average accuracy.

\begin{table}[t]
    \begin{minipage}{0.49\textwidth}
        \center
        \small
        \begin{tabular}{|l|c|c|}
        \hline
         Target   & $\Delta$BACC@20 & $\Delta$BACC \\
        \hline
        \hline
        ABC & 0.21 $\pm$ 0.06 & 0.08 $\pm$ 0.08 \\
        CHAT & 0.33 $\pm$ 0.07 & 0.21 $\pm$ 0.10 \\
        HOMEPAP & 0.17 $\pm$ 0.11 & 0.09 $\pm$ 0.08 \\
        MASS & 0.27 $\pm$ 0.07 & 0.20 $\pm$ 0.07 \\
        \hline
    \end{tabular}

    \end{minipage}
    \begin{minipage}{0.49\textwidth}
        \caption{Difference of BACC with \stmethod{} and without alignment in sleep staging for all the subjects ($\Delta$BACC column) and the 20\% lowest-performing subjects ($\Delta$BACC@20 column).}
        \label{tab:subject}
    \end{minipage}
\end{table}

\subsubsection{Impact of spatial and temporal information}

\begin{figure}[t]
    \centering
    \includegraphics[width=0.9\linewidth]{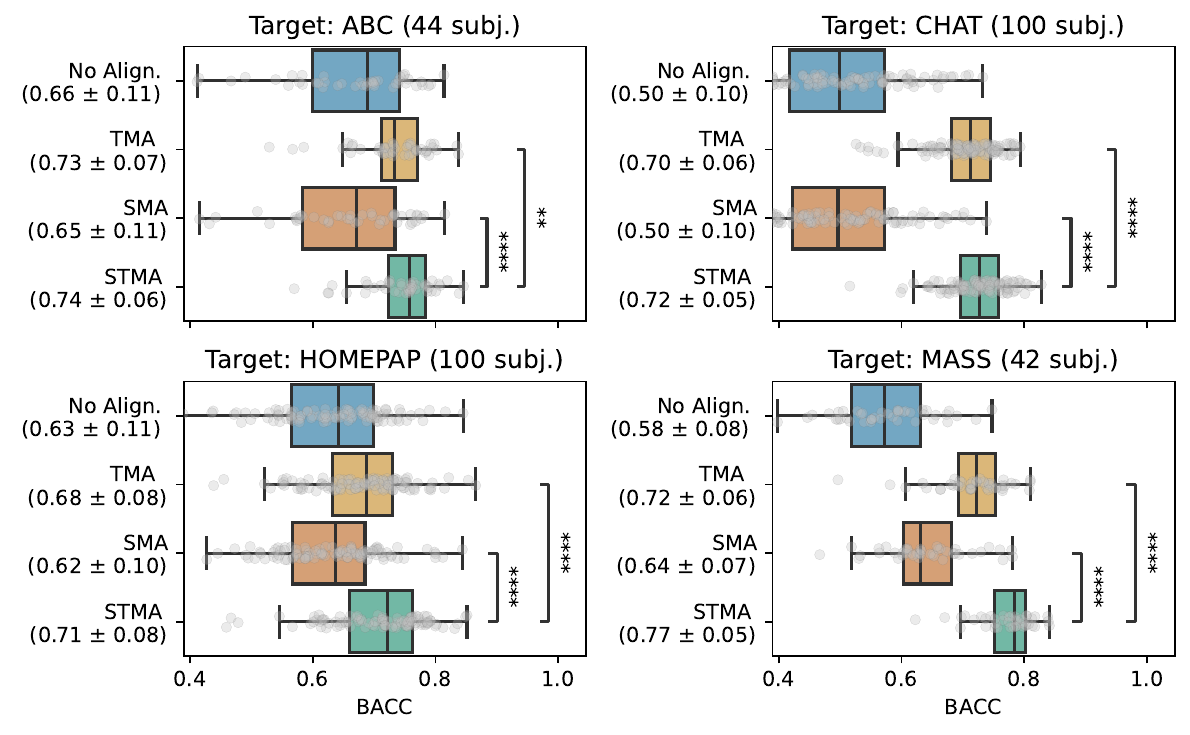}
    \caption{BACC for sleep stage classification with the different \method{}s. The number of stars illustrates the level of statistical significance (ns: non-significant, $**$: $10^{-2}$, or $****$: $10^{-4}$).}
    \label{fig:ablation}
\end{figure}

The previous experiment shows the benefit of alignment for sleep staging, especially when spatio-temporal information is used. 
In this section we propose to apply a \smethod{} (\ie no temporal correlation) or \tmethod{} (\ie no spatial correlation) on the sleep data. 
The \autoref{fig:ablation} shows the results of these methods for a leave-one-dataset-out experiment compared to without alignment and with spatio-temporal alignment. 

As intended, the pure spatial alignment struggles to improve the accuracy for each adaptation except for MASS, while pure temporal alignment improves the BACC by 10\% on average. 
It can be explained by the fact that the frequency-specific activity of the brain is critical for sleep classification. In contrast, spatial activity only brings marginal gains mainly due to channel redundancy except for MASS where RA and \smethod{} succeed to increase slightly the score.
However, the spatio-temporal alignment is statistically better than both single alignments.
\stmethod{} combines both alignments, providing the best of both worlds.

% \begin{table}[]
%     \centering
%     \begin{tabular}{|c|c|c|c|c|}
%     \hline
%     Dataset target &              ABC &             CHAT &          HOMEPAP &             MASS \\
%     \hline
%     \hline
%     No Align.        &  0.66 $\pm$ 0.12 &  0.47 $\pm$ 0.11 &  0.58 $\pm$ 0.13 &  0.64 $\pm$ 0.08 \\
%     \tmethod{}       &  0.72 $\pm$ 0.08 &  0.70 $\pm$ 0.06 &  0.68 $\pm$ 0.09 &  0.74 $\pm$ 0.07 \\
%     \stmethod{} & \textbf{ 0.75 $\pm$ 0.06} &  \textbf{0.73 $\pm$ 0.05} &  \textbf{0.70 $\pm$ 0.08} &  \textbf{0.78 $\pm$ 0.04} \\
%     \hline
%     \end{tabular}
%     \caption{BACC for different adaptation for each specific Monge mapping}
%     \label{tab:ablation}
% \end{table}

\subsection{Application to Brain Computer Interface (BCI)}

In sleep staging experiments, it has been demonstrated that spatio-temporal alignment is beneficial. 
Temporal information is essential for such a classification task.
In this section, we aim to evaluate the effectiveness of this alignment technique on another biosignal task.
While BCI relies less on temporal information for motor imagery classification, spatial information is crucial for distinguishing between classes.

% \subsubsection{Deep learning architectures and experimental setup}
% \paragraph{Architecture}
% Different neural network architectures have been developed specifically for BCI \cite{schirrmeister2017deep, lawhern2018eegnet}.
% {We focus on ShallowFBCSPNet \cite{schirrmeister2017deep}, which is an end-to-end neural network proposed to deal with multivariate time series based on convolutional layers and filter bank common spatial patterns (FBCSP).}
% The implementation used is from the braindecode library \cite{braindecode}.

% \paragraph{Training setup}
% We use the Adam optimizer with a cosine annealing learning rate scheduler
% starting at $6.25 \times 10^{-4}$. The batch size is set to $128$, and the
% training is stopped after 200 epochs. No validation set is used following the
% experimental setup by \cite{schirrmeister2017deep}. We optimize the
% cross-entropy for all methods. We report the average accuracy score (ACC)
% across ten different seeds.

% \paragraph{Filter size and Welch's parameters}
% The same sensitivity analysis performed on sleep staging data is applied to BCI
% data, with the results presented in \autoref{fig:sensi} (right).
% Based on this analysis, we select a filter size of 8.
% Note that for BCI, where the number of samples is much smaller than one night of sleep staging (3\,s $\ll$ 8\,h), the filter size is adjusted to a smaller value. Like for sleep staging setup, we use a Hann window \cite{blackman1958hann} and an overlap of $\frac{n_\ell}{2}$.

\subsubsection{Comparison between different alignments}

If Riemannian alignment was a new approach in sleep staging, it is often used in BCI applications \cite{xu2020bcicross, wimpff2024calibrationfree, Eldele2021atten}.
In previous studies, adaptation was done within one dataset or between two datasets.
We propose a new BCI leave-one-dataset-out experiment setup that compares results using 5 folds across datasets adaptation. 

\begin{figure}[t]
    \centering
    \includegraphics[width=0.8\linewidth]{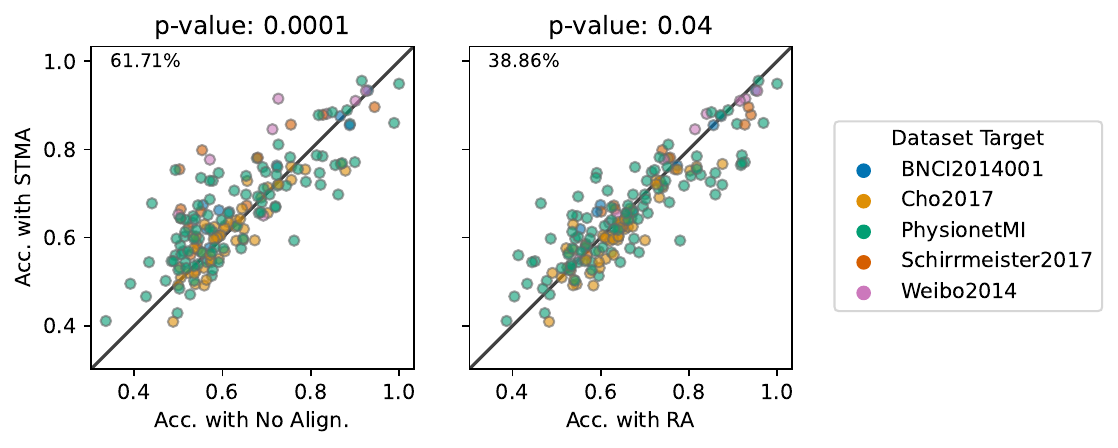}
    \caption{Comparison of the accuracy in BCI with \stmethod{} against No Align. and RA for each subject in target (colorized by dataset). The p-values are given at the top of the plot. The percentage of the dots above and below the black diagonal line is given on the bottom left.}
    \label{fig:scatter_bci}
\end{figure}

\autoref{tab:align_bci} compares scores between no alignment, RA, and \stmethod{} in leave-one-dataset-out settings.
{Alignments improve scores for each dataset. However, both methods exhibits similar performances. For three datasets, the two methods perform the same, when for the BNCI2014001 dataset \stmethod{} outperforms RA by 2\% and RA outperforms by 2\% \stmethod{} for Cho2017 dataset. }
{The BCI datasets have a low number of subjects, making a Wilcoxon test unreliable within each dataset.
To address this issue and increase statistical power, we perform a Wilcoxon test by pooling together all subjects from the five datasets.}
The \autoref{fig:scatter_bci} shows the comparison of the accuracy with \stmethod{} and without alignment (on the left) and with RA (on the right). 
The Wilcoxon test reveals that the first comparison exhibits a significant difference (p-value = $10^{-4}$) when 62\% of the subjects were increased.
The second comparison between \stmethod{} and RA showed a moderate difference in favor of RA, as indicated by the p-value.

Motor Imagery BCI classification relies primarily on spatial information compared to sleep staging, making the RA well-suited for the task; however, \stmethod{} still contributes to increasing the score, and it works even better in one scenario.

\begin{table}[t]
    \centering
    \small
    % \begin{tabular}{|l|c|c|c|c|c|}
    %     \hline
    %     Dataset Target &      BNCI2014001 &          Cho2017 &      PhysionetMI & Schirrmeister2017 &        Weibo2014 \\
    %     \hline
    %     \hline
    %     No Align.              &  0.70 $\pm$ 0.16 &  0.61 $\pm$ 0.09 &  0.63 $\pm$ 0.15 &   0.62 $\pm$ 0.13 &  0.69 $\pm$ 0.15 \\
    %     RA &  0.71 $\pm$ 0.14 & \textbf{ 0.63 $\pm$ 0.09} &  0.66 $\pm$ 0.15 &   \textbf{0.71 $\pm$ 0.14} &  0.76 $\pm$ 0.14 \\
    %     \stmethod{}    &  \textbf{0.75 $\pm$ 0.12} &  0.61 $\pm$ 0.08 &  \textbf{0.66 $\pm$ 0.13} &   0.70 $\pm$ 0.12 &  \textbf{0.77 $\pm$ 0.14} \\
    %     \hline
    % \end{tabular}
    \begin{tabular}{|l|c|c|c|c|c|}
        \hline
    Dataset Target & BNCI2014001 & Cho2017 & PhysionetMI & Schirrmeister2017 & Weibo2014 \\
        \hline
        \hline
    No Align.  & 0.70 $\pm$ 0.15 & 0.61 $\pm$ 0.09 & 0.63 $\pm$ 0.14 & 0.62 $\pm$ 0.14 & 0.70 $\pm$ 0.15 \\
    RA & 0.72 $\pm$ 0.14 & \textbf{0.63 $\pm$ 0.09} & 0.66 $\pm$ 0.15 & 0.71 $\pm$ 0.14 & \textbf{0.77 $\pm$ 0.14} \\
    \stmethod{} & \textbf{0.74 $\pm$ 0.12} & 0.61 $\pm$ 0.08 & \textbf{0.66 $\pm$ 0.12} & \textbf{0.71 $\pm$ 0.12} &\textbf{ 0.77 $\pm$ 0.14} \\
        \hline
    \end{tabular}

    \caption{Accuracy score for BCI Motor Imagery for 5 different datasets as target. Both RA and \stmethod{} methods improve the score. \stmethod{} is the best in three out of five scenarios.}
    \label{tab:align_bci}
\end{table}
% \begin{figure}[t]
%     \centering
%     \includegraphics[width=\linewidth]{images/LODO_bci.pdf}
%     \caption{Balanced accuracy for BCI motor imagery without normalization (blue), with Riemanian alignment (yellow), and with \stmethod{} (green) for different datasets in the target domain. Each dot represents one subject. \stmethod{} outperforms other methods for each target.}
%     \label{fig:perf_bci}
% \end{figure}

\subsubsection{Impact of spatial and temporal informations}
The previous section suggests that the spatial information is the most valuable part of spatio-temporal filtering.
We propose comparing pure spatial and temporal alignment with \stmethod{} in this section to support this statement. 
The \autoref{tab:ablation_bci} shows the results for 5-fold leave-one-dataset-out for all the \method{}s.
After aligning the data, spatio-temporal is the best option except for Cho2017, where no alignment works. 
However, when the temporal alignment improves slightly, spatial alignment achieves almost the same accuracy as \stmethod{} on average. 

Due to the specificities of the data, the BCI experiment leads to different
conclusions than sleep staging about the impact of spatial and temporal filtering. 
Choosing the appropriate filtering method (spatial or temporal) based on the
data allows for a quick and efficient alignment.

\begin{table}[t]
    \centering
    \small
    % \begin{tabular}{|l|c|c|c|c|c|}
    %     \hline
    %     Dataset Target &      BNCI2014001 &          Cho2017 &      PhysionetMI & Schirrmeister2017 &        Weibo2014 \\
    %     \hline
    %     \hline
    %     No Align.        &  0.70 $\pm$ 0.16 &  \textbf{0.61 $\pm$ 0.09} &  0.63 $\pm$ 0.15 &   0.62 $\pm$ 0.13 &  0.69 $\pm$ 0.15 \\
    %     \tmethod{}       &  0.71 $\pm$ 0.14 &  0.58 $\pm$ 0.07 &  0.63 $\pm$ 0.12 &   0.68 $\pm$ 0.13 &  0.73 $\pm$ 0.13 \\
    %     \smethod{}     &  0.74 $\pm$ 0.13 &  0.61 $\pm$ 0.08 &  0.65 $\pm$ 0.13 &   0.70 $\pm$ 0.13 &  0.77 $\pm$ 0.15 \\
    %     \stmethod{} &  \textbf{0.75 $\pm$ 0.12} &  0.61 $\pm$ 0.08 &  \textbf{0.66 $\pm$ 0.13} &   \textbf{0.70 $\pm$ 0.12} &  \textbf{0.77 $\pm$ 0.14} \\
    %     \hline
    % \end{tabular}
    \begin{tabular}{|l|c|c|c|c|c|}
        \hline
     Dataset Target & BNCI2014001 & Cho2017 & PhysionetMI & Schirrmeister2017 & Weibo2014 \\
        \hline
        \hline
     No Align. & 0.70 $\pm$ 0.15 & 0.61 $\pm$ 0.09 & 0.63 $\pm$ 0.14 & 0.62 $\pm$ 0.14 & 0.70 $\pm$ 0.15 \\
    \tmethod{}  & 0.71 $\pm$ 0.12 & 0.58 $\pm$ 0.07 & 0.63 $\pm$ 0.12 & 0.69 $\pm$ 0.12 & 0.73 $\pm$ 0.12 \\
    \smethod{} &\textbf{ 0.74 $\pm$ 0.12 }& \textbf{0.61 $\pm$ 0.08} & 0.65 $\pm$ 0.13 & \textbf{0.71 $\pm$ 0.12} & \textbf{0.77 $\pm$ 0.13} \\
    \stmethod{} & \textbf{0.74 $\pm$ 0.12 }& \textbf{0.61 $\pm$ 0.08} &\textbf{ 0.66 $\pm$ 0.12} & \textbf{0.71 $\pm$ 0.12} & 0.77 $\pm$ 0.14 \\
    \hline
    \end{tabular}
    \caption{Study of the impact of spatial and temporal informations on accuracy score for BCI Motor Imagery for 5 different datasets as target. \stmethod{} is almost always the best one, but \smethod{} also provides a good alignment.}
    \label{tab:ablation_bci}
\end{table}
% \begin{figure}[t]
%     \centering
%     \includegraphics[width=\linewidth]{images/LODO_ablation_bci.pdf}
%     \caption{Caption}
%     \label{fig:enter-label}
% \end{figure}

\begin{figure}[!ht]
    \centering
    \includegraphics[width=0.98\linewidth]{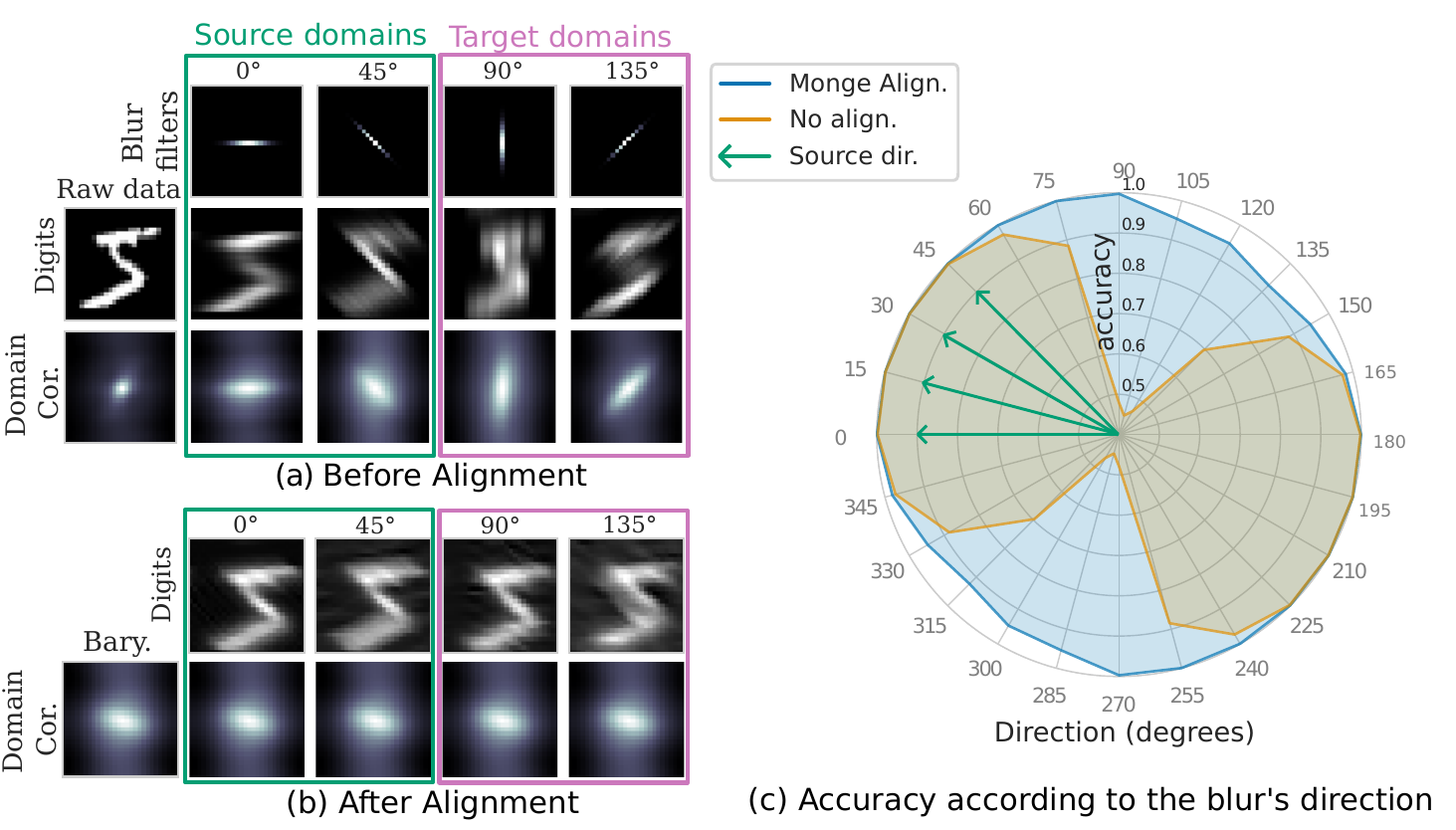}
    \caption{Blur MNIST visualization. 
    % In this example, several blur directions are considered. 
    The figure shows the sample and correlation for four blurred domains in different directions (a) before alignment and (b) after \method{}. 
    % Source domains are used to compute the barycenter and train the predictor, and target domains are only used for prediction. 
    (c) Accuracy of classification when trained on data containing only four blur directions (0°, 15°, 30°, and 45°) and testing on all possible directions. The accuracy drops when the testing data are blurred differently from the train. With \method{}, the accuracy improves significantly and is uniformly above 0.9.}
    \label{fig:mnist}
\end{figure}

\subsection{Illustration of \method{} on 2D signals}
% In this section, we propose an experiment on a toy dataset to show the effect of \method{}. In the following, we introduce the multi-directional blur MNIST and then show the impact of alignment on data and classification performance.
{The previous experiments demonstrate the efficiency of \method{} on
multivariate time series. Here, we want to illustrate the potential impact of the proposed methods beyond time series by considering images (\ie 2D signals).
% In this section, we extend the analysis of \method{}
% beyond 1D signals to 2D signals (\ie images).
On those signals, convolution and FFT are extended to
2D, allowing us to adapt \method{} with 2D filters. To illustrate the effect of
2D \method{}, we introduce a new custom multi-directional blur MNIST dataset.}

\paragraph{Toy dataset}
One considers a the toy dataset created from the digits classification dataset MNIST \cite{lecun1998mnist}. This dataset consists of several domains where the digits are blurred in one direction. 
In the \autoref{fig:mnist}, in the top left, four different domains are plotted with respective blur directions (\ie 0°, 45°, 90°, and 135°). 
The effect of the blur is clearly visible on both the raw digits and the correlation of the domains.
For analysis, twelve domains are created with a blurred direction ranging from 0° to 165°. Note that the blur is symmetric, which means that the blur directions from 180° to 345° are also covered.

\paragraph{Training setup}
The goal is to train on a subset of domains and predict the left-out domains. The source domains are the ones with directions going from 0° to 45° (\ie the green arrows on the right of \autoref{fig:mnist}). The target domains are the remaining ones (60° to 165°). 
To train the classifier, we use the architecture in the MNIST example of PyTorch\footnote{https://github.com/pytorch/examples/tree/main/mnist}. We use a Convolutional Neural Network (CNN) as a classifier. We use a batch size of 1000 and a learning rate of 1 with the Adadelta optimizer from PyTorch \cite{pytorch}. We stop the training when the validation loss does not decrease after ten epochs.

\paragraph{2D \method{} on blurred MNIST}
In this problem, the images are considered like 2D signals. One can compute an
estimation of the
PSD of an image with the squared FFT. The PSD of a domain is then the average of
the PSD of all images of the given domain. Taking the inverse Fourier of the PSD
of one domain provides the domain correlation.
To align the domain, first, the PSD barycenter is computed on the PSD of the source domains (\ie 0° to 45°). The resulting correlation of the barycenter is given in the left of \autoref{fig:mnist} (b). Since the source domains are concentrated in a specific range of directions, the correlation of the barycenter is focused on these directions.
After computing the barycenter, one can apply the Monge mapping with
\autoref{prop:temp_map} since we have univariate 2D signals here. The digits and the
correlation after alignment are given in \autoref{fig:mnist} (b). 
Now, the correlation of all the domains, source, or target is the same, and the digits look more alike.

\paragraph{Results with and without alignment}
One classifier is trained on domains that are not aligned, and another classifier is trained on aligned domains. The results are plotted on a spider plot in \autoref{fig:mnist} (c). The angle corresponds to the blur directions of the domain. The radius corresponds to the accuracy of the predictions.
For no alignment (orange line), the accuracy is close to 1 for blur direction from 0° to 45°, which is logical since it is the data used for training. 
However, as soon as the domains get further from the train angles, the accuracy decreases to reach the lowest score of 0.55 for the angle 105°.  
After using \method{}, the accuracy still decreased after getting away from 0° and 45° but stays above 0.9 of the accuracy score. 
Aligning the models helps classify unseen domains, even if the blur directions differ.
This simulated example shows that the method \method{} is generic and can be extended to
multi-dimensional signals.

\section{Conclusions}
{In this paper, we proposed Spatio-Temporal Monge Alignment (\stmethod{}), an Optimal Transport (OT) based method for multi-source and test-time Domain Adaptation (DA) of multivariate signals.
Our approach addresses the variability present in signals due to different recording conditions or hardware devices, which often leads to performance drops in machine learning applications.
\stmethod{} aligns signals' cross-power spectrum density (cross-PSD) to the Wasserstein barycenter of source domains, enabling predictions for new domains without retraining.
We also introduced two special cases of the method: Temporal Monge Alignment (\tmethod{}) and Spatial Monge Alignment (\smethod{}), each tailored for specific shift assumptions.
Our theoretical analysis provided non-asymptotic concentration bounds for the mapping estimation, demonstrating a bias-plus-variance error structure with a favorable variance decay rate of $\mathcal{O}(n_\ell^{-1/2})$.
Our numerical experiments on images and multivariate biosignal data, specifically in sleep staging and Brain-Computer Interface (BCI) tasks, showed that \stmethod{} leads to significant and consistent performance improvements over state-of-the-art methods. 
Importantly, \stmethod{} serves as a pre-processing step and is compatible with both shallow and deep learning methods, enhancing their performance without requiring refitting or access to source data at test-time.
In summary, \stmethod{} provides an efficient, solution to domain adaptation
challenges in multivariate signal processing. Since it remains simple and
requires only covariance estimation, it can be adapted to privacy preserving
applications using differential private covariances \cite{amin2019differentially} .
}

\section*{Acknowledgements}
\noindent
This work was supported by the grants ANR-22-PESN-0012 to AC under the France 2030 program, ANR-20-CHIA-0016 and ANR-20-IADJ-0002 to AG while at Inria, and ANR-23-ERCC-0006 to RF, all from Agence nationale de la recherche (ANR).
This project has also received funding from the European Union’s Horizon  Europe  research  and  innovation  programme  under  grant  agreement  101120237 (ELIAS).   

All the datasets used for this work were accessed and processed on the Inria compute infrastructures.
\\
\sloppy
Numerical computation was enabled by the 
scientific Python ecosystem: 
Matplotlib~\cite{matplotlib}, 
Scikit-learn~\cite{scikit-learn}, 
Numpy~\cite{numpy}, Scipy~\cite{scipy}, PyTorch~\cite{pytorch} and PyRiemann~\cite{pyriemann}, MNE~\cite{GramfortEtAl2013a}.

% \newpage
% \bibliographystyle{utphys}
\bibliographystyle{utphys}
\bibliography{biblio}
\newpage

\section{Supplementary Materials}

\subsection{Proof of Proposition~\ref{prop:filtering}: PSD and filter sub-sampling}
{
Given $t \in \intset{\filtersize}$, and since $(\bA)_{1l} = 0$ for $l \not\in \mathfrak{I} \triangleq \intset{\ceil{\filtersize/2}} \cup \llbracket n_\ell - \floor{\filtersize/2} + 1, n_\ell \rrbracket$, we have
\begin{equation*}
    \begin{aligned}
        \left(\frac{1}{\sqrt{\filtersize}} \bF_{\filtersize}\hermconj \bp\right)_t &= \frac{1}{\filtersize}\sum_{l=1}^\filtersize \exp\left(2i \pi \frac{(l-1) (t-1)}{\filtersize}\right) (\bp)_l \\
        &= \frac{1}{\filtersize} \sum_{l=1}^\filtersize \exp\left(2i \pi \frac{(l-1) (t-1)}{n_\ell} \frac{n_\ell}{\filtersize}\right) (\bq)_{\frac{(l-1)n_\ell}{\filtersize} + 1} \\
        &= \frac{1}{\filtersize} \sum_{\substack{m \in \{ 1, \frac{n_\ell}{\filtersize} + 1, \dots,\\ \frac{(\filtersize-1)n_\ell}{\filtersize} + 1\} }} \exp\left(2i \pi \frac{(m-1) (t-1)}{n_\ell} \right) (\bq)_m \\
        &= \frac{1}{\filtersize} \sum_{\substack{m \in \{ 1, \frac{n_\ell}{\filtersize} + 1, \dots,\\ \frac{(\filtersize-1)n_\ell}{\filtersize} + 1\} }} \sum_{l=1}^{n_\ell} \exp\left(-2i \pi \frac{(m-1) (l-t)}{n_\ell} \right) (\bA)_{1l} \\
        &= \frac{1}{\filtersize} \sum_{l=1}^{n_\ell} \sum_{k = 0}^{\filtersize-1} \exp\left(-2i \pi \frac{k (l-t)}{\filtersize} \right) (\bA)_{1l} \\
        &= \frac{1}{\filtersize} \sum_{l\in \mathfrak{I}} \sum_{k = 0}^{\filtersize-1} \exp\left(-2i \pi \frac{k (l-t)}{\filtersize} \right) (\bA)_{1l} \enspace .
    \end{aligned}
\end{equation*}
Then, we split the sum over $l$ with terms such that $l-t$ is a multiple of $f$, denoted $l-t \propto f$, and terms where $l-t$ is not a multiple of $f$, denoted $l-t \not\propto f$,
\begin{equation*}
    \begin{aligned}
        \left(\frac{1}{\sqrt{\filtersize}} \bF_{\filtersize}\hermconj \bp\right)_t &= \frac{1}{\filtersize} \sum_{\substack{l\in \mathfrak{I}  \\ l - t \propto \filtersize}} \sum_{k = 0}^{\filtersize-1} \underbrace{\exp\left(-2i \pi \frac{k (l-t)}{\filtersize} \right)}_{=1} (\bA)_{1l} + \frac{1}{\filtersize} \sum_{\substack{l\in \mathfrak{I}  \\ l - t \not\propto \filtersize}} \sum_{k = 0}^{\filtersize-1} \exp\left(-2i \pi \frac{k (l-t)}{\filtersize} \right) (\bA)_{1l} \\
        &= \sum_{\substack{l\in \mathfrak{I}  \\ l - t \propto \filtersize}} (\bA)_{1l} + \frac{1}{\filtersize} \sum_{\substack{l\in \mathfrak{I}  \\ l - t \not\propto \filtersize}} \underbrace{\frac{1 - \exp\left(-2i \pi (l-t)\right)}{1 - \exp\left(-2i \pi \frac{(l-t)}{\filtersize} \right)}}_{=0} (\bA)_{1l} = \sum_{\substack{l\in \mathfrak{I}  \\ l - t \propto \filtersize}} (\bA)_{1l} \enspace .
    \end{aligned}
\end{equation*}
\paragraph{Case $t\in \intset{\ceil{\filtersize/2}}$:}
Given $t\in \intset{\ceil{\filtersize/2}}$ and $l\in \mathfrak{I}$.
Since $n_\ell \propto \filtersize$, there exists $k \in \bbZ$ such that $l-t \in \llbracket 1 - \ceil{\filtersize/2}, \ceil{\filtersize/2} - 1 \rrbracket \cup \llbracket (k-1)\filtersize + 1, k\filtersize - 1 \rrbracket$.
Since $l-t \propto f$, we get that $l-t = 0$ and thus,
\begin{equation*}
    \left(\frac{1}{\sqrt{\filtersize}} \bF_{\filtersize}\hermconj \bp\right)_t = (\bA)_{1t} \enspace.
\end{equation*}

\paragraph{Case $t\in \llbracket \ceil{\filtersize/2} + 1, \filtersize \rrbracket$:}
Given $t\in \llbracket \ceil{\filtersize/2} + 1, \filtersize \rrbracket$ and $l\in \mathfrak{I}$, we get that $l-t \in \llbracket -\filtersize + 1, - 1 \rrbracket \cup \llbracket n_\ell - \floor{\filtersize/2} - \filtersize + 1, n_\ell - \floor{\filtersize/2} - 1 \rrbracket$.
Since $n_\ell \propto \filtersize$ and $l-t \propto \filtersize$, it implies that $l-t = n_\ell - \filtersize$. 
Hence, we get that
\begin{equation*}
    \left(\frac{1}{\sqrt{\filtersize}} \bF_{\filtersize}\hermconj \bp\right)_t = (\bA)_{1(n_\ell - \filtersize + t)} \enspace.
\end{equation*}
}
% Given $t \in \intset{\floor{\filtersize/2}}$, we have
% \begin{equation*}
%     \begin{aligned}
%         \left(\frac{1}{\sqrt{\filtersize}} \bF_{\filtersize}\hermconj \bp\right)_{\filtersize - t + 1} &= \frac{1}{\filtersize}\sum_{l=1}^\filtersize \exp\left(2i \pi \frac{(l-1) (f-t)}{\filtersize}\right) (\bp)_l \\
%         &=\frac{1}{\filtersize}\sum_{l=1}^\filtersize \exp\left(2i \pi (l-1)\right) \exp\left(-2i \pi \frac{(l-1) t}{\filtersize}\right) (\bp)_l \\
%         &=\frac{1}{\filtersize}\sum_{l=1}^\filtersize \exp\left(-2i \pi \frac{(l-1) t}{\filtersize}\right) (\bp)_l \\
%         &=\text{conj}\left(\frac{1}{\sqrt{\filtersize}} \bF_{\filtersize}\hermconj \bp\right)_{t + 1} = \left(\frac{1}{\sqrt{\filtersize}} \bF_{\filtersize}\hermconj \bp\right)_{t + 1} 
%     \end{aligned}
% \end{equation*}
% which completes the proof.

\subsection{Proof of Proposition~\ref{prop:spatio_temp_map}: spatio-temporal mapping}
\label{proof:mapping}
We recall that the Monge mapping is
\begin{equation*}
    \bA = \bSigma_s^{-\frac{1}{2}}\left( \bSigma_s^{\frac{1}{2}} \bSigma_t\bSigma_s^{\frac{1}{2}} \right)^{\frac{1}{2}}  \bSigma_s^{-\frac{1}{2}}\;.
\end{equation*}
\sloppy
We inject the decomposition from the \autoref{eq:blockdiag_T}, \ie 
$\bSigma_d = \bF \bU \bQ_d \bU^\trans \bF\hermconj$ for $d\in\{s,t\}$ in the Monge mapping.
Since $\bF \bU$ is an unitary matrix, for every $\bQ \in {\cH_{n_c n_\ell}^{++}}$, we have $(\bF \bU \bQ \bU^\trans \bF\hermconj)^{-\frac{1}{2}} = \bF \bU {\bQ}^{-\frac{1}{2}} \bU^\trans \bF\hermconj$.
Hence, we get
\begin{align*}
    \bA &= \bF \bU \bQ_s^{-\frac{1}{2}} \bU^\trans \bF\hermconj \left( \bF \bU \bQ_s^{\frac{1}{2}} \bU^\trans \bF\hermconj\bF \bU \bQ_t \bU^\trans \bF\hermconj\bF \bU \bQ_s^{\frac{1}{2}} \bU^\trans \bF\hermconj \right)^{\frac{1}{2}} \bF \bU \bQ_s^{-\frac{1}{2}} \bU^\trans \bF\hermconj \\
    &= \bF \bU \bQ_s^{-\frac{1}{2}} \bU^\trans \bF\hermconj \bF \bU  \left(\bQ_s^{\frac{1}{2}} \bQ_t \bQ_s^{\frac{1}{2}}\right)^{\frac{1}{2}} \bU^\trans \bF\hermconj \bF \bU \bQ_s^{-\frac{1}{2}} \bU^\trans \bF\hermconj = \bF \bU \bQ_s^{-\frac{1}{2}} \left(\bQ_s^{\frac{1}{2}} \bQ_t \bQ_s^{\frac{1}{2}}\right)^{\frac{1}{2}} \bQ_s^{-\frac{1}{2}} \bU^\trans \bF\hermconj \;.
\end{align*}
Furthermore, $\bQ_s^{-\frac{1}{2}} \left(\bQ_s^{\frac{1}{2}} \bQ_t \bQ_s^{\frac{1}{2}}\right)^{\frac{1}{2}} \bQ_s^{-\frac{1}{2}}$ is a block-diagonal matrix since $\bQ_s$ and $\bQ_t$ are block-diagonal matrices.
Thus, there exists $\bq_{i,j} \in {\bbC^{n_\ell}}$ for $i,j \in \intset{n_c}$ such that
\begin{align*}
    \bA =
    \bF \begin{pmatrix}
        \diag(\bq_{1, 1}) & \dots & \diag(\bq_{1, n_c}) \\
        \dots & \dots & \dots \\
        \diag(\bq_{n_c, 1}) & \dots & \diag(\bq_{n_c, n_c})
    \end{pmatrix} \bF\hermconj \; .
    % \label{eq:Monge_map_diag}
\end{align*}
It follows that the block matrices of $\bA$ are $\bA_{i,j} = \bF_{n_\ell} \diag(\bq_{i, j}) \bF_{n_\ell}\hermconj$ for $i,j \in \intset{n_c}$.
Then, the $\filtersize$-Monge mapping $\widetilde{\bA}$ introduced in the Definition~\ref{def:approx_mapping} has block matrices $\widetilde{\bA}_{i,j} = \cP_{\filtersize}(\bA_{i,j})$ for $i,j \in \intset{n_c}$.
Hence, given $\bX = [\bx_1, \dots, \bx_{n_c}]^\trans \in \bbR^{n_c \times n_\ell}$, the $\filtersize$-Monge mapping is
\begin{align*}
    m_{\filtersize}\left(\bX\right) &=  \Vectr^{-1}\left(\widetilde{\bA} \Vectr(\bX)\right) \\
    &=\Vectr^{-1}\left(
    \begin{pmatrix}
        \cP_{\filtersize}(\bA_{1,1}) & \dots & \cP_{\filtersize}(\bA_{1,C}) \\
        \dots & \dots & \dots \\
        \cP_{\filtersize}(\bA_{C,1}) & \dots & \cP_{\filtersize}(\bA_{C,C})
    \end{pmatrix}
    \begin{pmatrix}
        \bx_1 \\
        \vdots \\
        \bx_{n_c}
    \end{pmatrix} \right)=
    \Vectr^{-1}\begin{pmatrix}
        \sum_{j=1}^{n_c} \cP_{\filtersize}(\bA_{1,j}) \bx_j \\
        \vdots \\
        \sum_{j=1}^{n_c} \cP_{\filtersize}(\bA_{C,j}) \bx_j
    \end{pmatrix}\;.
\end{align*}
Hence, from Proposition~\ref{prop:filtering}, we get
\begin{equation*}
    m_{\filtersize}\left(\bX\right) = \left[\sum_{j=1}^{n_c} \bh_{1,j} * \bx_j, \dots, \sum_{j=1}^{n_c} \bh_{C,j} * \bx_j \right]^\trans
\end{equation*}
where
%\begin{equation*}
$\bh_{i,j} = \frac{1}{\sqrt{\filtersize}} \bF_{\filtersize}\hermconj  \bp_{i,j}$
    % \bh_{i,j} = \frac{1}{\sqrt{\filtersize}} \bS_F\bF_{\filtersize}\hermconj  \bp_{i,j}
%\end{equation*}
and $\bp_{i,j} = g_{\filtersize}(\diag(\bF_{n_\ell}\hermconj \cP_{\filtersize}(\bA_{i,j}) \bF_{n_\ell})) = g_{\filtersize}(\bq_{i,j})$.
Denoting
\begin{equation*}
    \bQ \triangleq \diag\left(\bB_1, \dots, \bB_{n_\ell}\right) \triangleq \bQ_s^{-\frac{1}{2}} \left(\bQ_s^{\frac{1}{2}} \bQ_t \bQ_s^{\frac{1}{2}}\right)^{\frac{1}{2}} \bQ_s^{-\frac{1}{2}} \in \bbR^{n_c n_\ell \times n_c n_\ell}, \quad
    \bB_l = 
    \begin{pmatrix}
        (\bq_{1, 1})_l & \dots & (\bq_{1, n_c})_l \\
        \dots & \dots & \dots \\
        (\bq_{n_c, 1})_l & \dots & (\bq_{n_c, n_c})_l
    \end{pmatrix},
\end{equation*}
the computation of $\bh_{i,j}$ only requires the computation of
\begin{equation*}
    g_{\filtersize}(\bQ) \triangleq \diag\left(\bB_1, \bB_{\frac{n_\ell}{\filtersize} + 1}, \dots, \bB_{\frac{(\filtersize-1)n_\ell}{\filtersize} + 1}\right) \in \bbR^{n_c \filtersize \times n_c \filtersize}
\end{equation*}
where $g_{\filtersize}$ has been extended to block-diagonal matrices. 
Thus, by denoting $\bP_d = g_{\filtersize}(\bQ_d)$ for $d\in \{s,t\}$, we get
\begin{equation*}
    \begin{pmatrix}
        \diag(\bp_{1, 1}) & \dots & \diag(\bp_{1, n_c}) \\
        \dots & \dots & \dots \\
        \diag(\bp_{n_c, 1})  & \dots & \diag(\bp_{n_c, n_c}) 
    \end{pmatrix}
    = \bV \bP_s^{-\frac{1}{2}} \left(\bP_s^{\frac{1}{2}} \bP_t \bP_s^{\frac{1}{2}}\right)^{\frac{1}{2}} \bP_s^{-\frac{1}{2}} \bV^\trans \in {\cH_{n_c \filtersize}^{++}},
\end{equation*}
with $\bV\in\bbR^{n_c \filtersize \times n_c \filtersize}$ the permutation matrix defined in \autoref{eq:blockdiag_{\filtersize}}.
    
\subsection{Proof of Lemma~\ref{lemma:spatio_temp_bary}: Spatio-Temporal barycenter}
We recall the formulation of the fixed-point
\begin{equation}
        \overline{\bSigma} = \frac{1}{n_d}\sum_{k=1}^{n_d} \left( \overline{\bSigma}^{\frac{1}{2}} \bSigma_k\overline{\bSigma}^{\frac{1}{2}} \right)^{\frac{1}{2}}
\end{equation}
We inject the decomposition from the \autoref{eq:blockdiag_T}, \ie 
$\bSigma_k = \bF \bU \bQ_k \bU^\trans \bF\hermconj$ for $k\in\intset{n_d}$ in the fixed-point equation
\begin{align*}
        \overline{\bSigma} = \bF \bU \overline{\bQ} \bU^\trans \bF\hermconj &= \frac{1}{n_d}\sum_{k=1}^{n_d} \left( \left( \bF \bU \overline{\bQ} \bU^\trans \bF\hermconj\right)^{\frac{1}{2}} \bF \bU \bQ_k \bU^\trans \bF\hermconj\left( \bF \bU \overline{\bQ} \bU^\trans \bF\hermconj\right)^{\frac{1}{2}} \right)^{\frac{1}{2}}\\
        &= \bF \bU \left(\frac{1}{n_d}\sum_{k=1}^{n_d} \left( \overline{\bQ}^{\frac{1}{2}} \bQ_k \overline{\bQ}^{\frac{1}{2}} \right)^{\frac{1}{2}}\right)\bU^\trans \bF\hermconj
\end{align*}
\subsection{Proof of Proposition~\ref{prop:temp_map}: Temporal mapping}
We recall that the Monge mapping is
\begin{equation*}
    \bA = \bSigma_s^{-\frac{1}{2}}\left( \bSigma_s^{\frac{1}{2}} \bSigma_t\bSigma_s^{\frac{1}{2}} \right)^{\frac{1}{2}}  \bSigma_s^{-\frac{1}{2}}\;.
\end{equation*}
Source and target signals follow the Assumption~\ref{assu:temp}, \ie $\bSigma_d = \bF \diag(\bq_{1,d}, \dots, \bq_{C,d}) \bF\hermconj$.
Thus, we get that $\bA$ is a block diagonal matrix, \ie
\begin{equation*}
    \bA = \diag\left(\bF_{n_\ell} \diag\left(\bq_{1,t}^{\odot\frac{1}{2}}\odot\bq_{1,s}^{\odot-\frac{1}{2}}\right) \bF_{n_\ell}, \dots, \bF_{n_\ell} \diag\left(\bq_{C,t}^{\odot\frac{1}{2}}\odot\bq_{C,s}^{\odot-\frac{1}{2}}\right) \bF_{n_\ell}\right) \;.
\end{equation*}
Hence, given $\bX = [\bx_1, \dots, \bx_{n_c}]^\trans \in \bbR^{n_c \times n_\ell}$, the $\filtersize$-Monge mapping is
\begin{align*}
    m_{\filtersize}\left(\bX\right) &=  \Vectr^{-1}\left(\widetilde{\bA} \Vectr(\bX)\right) \\
    &=
    \Vectr^{-1}\begin{pmatrix}
        \cP_{\filtersize}\left( \bF_{n_\ell}\diag\left(\bq_{1,t}^{\odot\frac{1}{2}}\odot\bq_{1,s}^{\odot-\frac{1}{2}}\right)\bF_{n_\ell} \right) \bx_1 \\
        \vdots \\
        \cP_{\filtersize}\left( \bF_{n_\ell}\diag\left(\bq_{C,t}^{\odot\frac{1}{2}}\odot\bq_{C,s}^{\odot-\frac{1}{2}}\right)\bF_{n_\ell} \right) \bx_{n_c} 
    \end{pmatrix} \\
    &= \left[
        \cP_{\filtersize}\left( \bF_{n_\ell}\diag\left(\bq_{1,t}^{\odot\frac{1}{2}}\odot\bq_{1,s}^{\odot-\frac{1}{2}}\right)\bF_{n_\ell} \right) \bx_1, 
        \dots,
        \cP_{\filtersize}\left( \bF_{n_\ell}\diag\left(\bq_{C,t}^{\odot\frac{1}{2}}\odot\bq_{C,s}^{\odot-\frac{1}{2}}\right)\bF_{n_\ell} \right) \bx_{n_c} 
    \right] \;.
\end{align*}
Hence, from Proposition~\ref{prop:filtering}, we get
\begin{equation*}
    m_{\filtersize}\left(\bX\right) = \left[\bh_1 * \bx_1, \dots, \bh_C * \bx_{n_c} \right]^\trans
\end{equation*}
where $\bh_i = \frac{1}{\sqrt{\filtersize}} \bF_{\filtersize}\hermconj \left(\bp_{i,t}^{\odot\frac{1}{2}}\odot\bp_{i,s}^{\odot-\frac{1}{2}}\right) \in \bbR^{\filtersize}$ and $\bp_{i,d} = g_{\filtersize}(\bq_{i,d})$.
% where $\bh_i = \frac{1}{\sqrt{\filtersize}} \bS_F\bF_{\filtersize}\hermconj \left(\bp_{i,t}^{\odot\frac{1}{2}}\odot\bp_{i,s}^{\odot-\frac{1}{2}}\right) \in \bbR^{\filtersize}$ and $\bp_{i,d} = g_{\filtersize}(\bq_{i,d})$.

\subsection{Proof of Lemma~\ref{lemma:temp_bary}: Temporal barycenter}
We recall the formulation of the fixed-point
\begin{equation}
        \overline{\bSigma} = \frac{1}{n_d}\sum_{k=1}^{n_d} \left( \overline{\bSigma}^{\frac{1}{2}} \bSigma_k\overline{\bSigma}^{\frac{1}{2}} \right)^{\frac{1}{2}}
\end{equation}

Source and target signals follow the Assumption~\ref{assu:temp}, \ie $\bSigma_k = \bF \diag(\bq_{1,k}, \dots, \bq_{n_c,k}) \bF\hermconj$.
Thus, we get that $\overline{\bSigma}$ is a block diagonal matrix, \ie
\begin{align*}
    \overline{\bSigma} &= \bF \diag(\overline{\bq}_{1,d}, \dots, \overline{\bq}_{C,d}) \\
    &= \bF\diag\left( \frac{1}{n_d}\sum_{k=1}^{n_d} \left( \overline{\bq}^{\odot\frac{1}{2}}\odot \bq_{1,k}\odot \overline{\bq}^{\odot\frac{1}{2}} \right)^{\odot\frac{1}{2}} , \dots, \frac{1}{n_d}\sum_{k=1}^{n_d} \left( \overline{\bq}^{\odot\frac{1}{2}}\odot \bq_{n_c,k} \odot \overline{\bq}^{\odot\frac{1}{2}} \right)^{\odot\frac{1}{2}} \right) \\
    &= \bF\diag\left( \frac{1}{n_d}\sum_{k=1}^{n_d} \overline{\bq}^{\odot\frac{1}{2}}\odot \bq_{1,k}^{\odot\frac{1}{2}}, \dots, \frac{1}{n_d}\sum_{k=1}^{n_d} \overline{\bq}^{\odot\frac{1}{2}}\odot \bq_{n_c,k}^{\odot\frac{1}{2}}\right)\\
     &= \bF\diag\left( \left(\frac{1}{n_d}\sum_{k=1}^{n_d}  \bq_{1,k}^{\odot\frac{1}{2}}\right)^{\odot2}, \dots, \left(\frac{1}{n_d}\sum_{k=1}^{n_d}  \bq_{n_c,k}^{\odot\frac{1}{2}}\right)^{\odot2}\right)
    \;.
\end{align*}

\subsection{Proof of Proposition~\ref{prop:spat_map}: Spatial mapping}
From Assumption~\ref{assu:spat}, we have
\begin{equation*}
    \bSigma_d = \bXi_d \otimes \bI_{n_\ell}.
\end{equation*}
Since,
\begin{equation*}
    \bSigma_d = \bF \bU \bQ_d \bU^\trans \bF\hermconj
\end{equation*}
we get that
\begin{equation*}
    \bQ_d = \bU^\trans \bF\hermconj (\bXi_d \otimes \bI_{n_\ell}) \bF \bU \;.
\end{equation*}
From Proposition~\ref{prop:spatio_temp_map}, this implies that
\begin{align*}
    \bA &= \bF \bU \bQ_s^{-\frac{1}{2}} \left(\bQ_s^{\frac{1}{2}} \bQ_t \bQ_s^{\frac{1}{2}}\right)^{\frac{1}{2}} \bQ_s^{-\frac{1}{2}} \bU^\trans \bF\hermconj \\
    &= (\bXi_s^{-\frac{1}{2}} \otimes \bI_{n_\ell}) \left((\bXi_s^{\frac{1}{2}} \otimes \bI_{n_\ell}) (\bXi_t \otimes \bI_{n_\ell}) (\bXi_s^{\frac{1}{2}} \otimes \bI_{n_\ell})\right)^{\frac{1}{2}} (\bXi_s^{-\frac{1}{2}} \otimes \bI_{n_\ell}) \\
    &= (\bXi_s^{-\frac{1}{2}} \otimes \bI_{n_\ell}) \left((\bXi_s^{\frac{1}{2}} \bXi_t \bXi_s^{\frac{1}{2}})^{\frac{1}{2}} \otimes \bI_{n_\ell}\right)  (\bXi_s^{-\frac{1}{2}} \otimes \bI_{n_\ell}) \\
    &= \left(\bXi_s^{-\frac{1}{2}}(\bXi_s^{\frac{1}{2}} \bXi_t \bXi_s^{\frac{1}{2}})^{\frac{1}{2}}\bXi_s^{-\frac{1}{2}} \otimes \bI_{n_\ell}\right) \; .
\end{align*}
Thus, given $\bX = [\bx_1, \dots, \bx_{n_c}]^\trans$, recalling the formula $\Vectr(\bXi \bX) = (\bXi \otimes \bI_{n_\ell}) \Vectr(\bX)$, the $\filtersize$-Monge mapping is written
\begin{align*}
    m_{\filtersize}\left(\bX\right) &=  \Vectr^{-1}\left(\bA \Vectr(\bX)\right) \\
    &= \Vectr^{-1}\left(\left(\bXi_s^{-\frac{1}{2}}(\bXi_s^{\frac{1}{2}} \bXi_t \bXi_s^{\frac{1}{2}})^{\frac{1}{2}}\bXi_s^{-\frac{1}{2}} \otimes \bI_{n_\ell}\right) \Vectr(\bX)\right) = \bXi_s^{-\frac{1}{2}}(\bXi_s^{\frac{1}{2}} \bXi_t \bXi_s^{\frac{1}{2}})^{\frac{1}{2}}\bXi_s^{-\frac{1}{2}} \bX \;.
\end{align*}

\subsection{Proof of Lemma~\ref{lemma:spat_bary}: Spatial barycenter}
From Assumption~\ref{assu:spat}, we have for all $k$
\begin{equation*}
    \bSigma_k = \bXi_k \otimes \bI_{n_\ell}.
\end{equation*}
Since,
\begin{equation*}
    \bSigma_k = \bF \bU \bQ_k \bU^\trans \bF\hermconj
\end{equation*}
we get that
\begin{equation*}
    \bQ_k = \bU^\trans \bF\hermconj (\bXi_k \otimes \bI_{n_\ell}) \bF \bU \;.
\end{equation*}
From Lemma~\ref{lemma:spatio_temp_bary}, this implies that
\begin{align*}
        \overline{\bSigma} = \overline{\bXi}\otimes \bI_{n_\ell} &= \bF \bU \overline{\bQ} \bU^\trans \bF\hermconj \\
        &= \bF \bU \left(\frac{1}{n_d}\sum_{k=1}^{n_d} \left( (\overline{\bXi}^{\frac{1}{2}}\otimes \bI_{n_\ell})(\bXi_k \otimes \bI_{n_\ell}) (\overline{\bXi}^{\frac{1}{2}}\otimes \bI_{n_\ell}) \right)^{\frac{1}{2}}\right)\bU^\trans \bF\hermconj\\
        &= \bF \bU \left(\frac{1}{n_d}\sum_{k=1}^{n_d} \left(\left(\overline{\bXi}^{\frac{1}{2}}\bXi_k \overline{\bXi}^{\frac{1}{2}}\right)^{\frac{1}{2}}\otimes \bI_{n_\ell}\right)\right)\bU^\trans \bF\hermconj
\end{align*}

\subsection{Proof of \autoref{the:concentration_spatiotemp}: \stmethod{} concentration bound}
From~\cite{flamary2020concentration}, we have that
\begin{equation*}
    \Vert \widehat{\bA} - \bA \Vert \lesssim \frac{\kappa(\bSigma)}{\lambda_\text{min}^{\nicefrac{1}{2}}(\bSigma^{\nicefrac{1}{2}} \overline{\bSigma} \bSigma^{\nicefrac{1}{2}})} \Vert \widehat{\overline{\bSigma}} - \overline{\bSigma} \Vert + \frac{\kappa(\overline{\bSigma}) \Vert \overline{\bSigma} \Vert \Vert \bSigma^{-1} \Vert}{\lambda_\text{min}^{\nicefrac{1}{2}}(\overline{\bSigma}^{\nicefrac{1}{2}} \bSigma \overline{\bSigma}^{\nicefrac{1}{2}})} \Vert \widehat{\bSigma}_t - \bSigma_t \Vert \;.
\end{equation*}
We first need to upper bound the term $\Vert \widehat{\overline{\bSigma}} - \overline{\bSigma} \Vert$.

\subsubsection{Upper-bound Wasserstein barycenter for one iteration}
From \cite{peyre_computational_2020} we know that the fixed point equation is not contracting. We can not show the convergence for infinite iterations. In our case, we propose to upper-bound the first iteration of the barycenter when the initialization is done using the Euclidean mean.
We want to bound $\Vert \widehat{\overline{\bSigma}}_1 - \overline{\bSigma}_1 \Vert$ with 
% $$
% \overline{\bSigma}_1 = \frac{1}{n_d} \sum_{k=1}^{n_d} \left(\overline{\bSigma}_0^{\nicefrac{1}{2}}\bSigma_k \overline{\bSigma}_0^{\nicefrac{1}{2}}\right)^{\nicefrac{1}{2}}
% $$ 
% with
\begin{equation}
    \overline{\bSigma}_1 = \frac{1}{n_d} \sum_{k=1}^{n_d} \left(\overline{\bSigma}_0^{\nicefrac{1}{2}}\bSigma_k \overline{\bSigma}_0^{\nicefrac{1}{2}}\right)^{\nicefrac{1}{2}},\quad \overline{\bSigma}_0 = \frac{1}{n_d}\sum_{k=1}^{n_d} \bSigma_k,\quad \text{and} \quad \widehat{\overline{\bSigma}}_0 = \frac{1}{n_d}\sum_{k=1}^{n_d} \widehat{\bSigma_k}.
    \label{eq:sig0}
\end{equation}
% and
% \begin{equation}
% \widehat{\overline{\bSigma}}_0 = \frac{1}{n_d}\sum_{k=1}^{n_d} \widehat{\bSigma_k}\;.
%     \label{eq:sig0hat}
% \end{equation}
We then have
\begin{equation*}
    \Vert \widehat{\overline{\bSigma}}_1 - \overline{\bSigma}_1 \Vert \leq \frac{1}{n_d} \sum_{k=1}^{n_d} \left\Vert \left(\widehat{\overline{\bSigma}}_0^{\nicefrac{1}{2}}\widehat{\bSigma}_k \widehat{\overline{\bSigma}}_0^{\nicefrac{1}{2}}\right)^{\nicefrac{1}{2}} - \left(\overline{\bSigma}_0^{\nicefrac{1}{2}}\bSigma_k \overline{\bSigma}_0^{\nicefrac{1}{2}}\right)^{\nicefrac{1}{2}} \right\Vert.
\end{equation*}
We apply Lemma 2.1 in \cite{SCHMITT1992215} to get
\begin{equation*}
    \left\Vert \left(\widehat{\overline{\bSigma}}_0^{\nicefrac{1}{2}}\widehat{\bSigma}_k \widehat{\overline{\bSigma}}_0^{\nicefrac{1}{2}}\right)^{\nicefrac{1}{2}} - \left(\overline{\bSigma}_0^{\nicefrac{1}{2}}\bSigma_k \overline{\bSigma}_0^{\nicefrac{1}{2}}\right)^{\nicefrac{1}{2}} \right\Vert \leq \frac{1}{\xi_k}\left\Vert \widehat{\overline{\bSigma}}_0^{\nicefrac{1}{2}}\widehat{\bSigma}_k \widehat{\overline{\bSigma}}_0^{\nicefrac{1}{2}} - \overline{\bSigma}_0^{\nicefrac{1}{2}}\bSigma_k \overline{\bSigma}_0^{\nicefrac{1}{2}} \right\Vert\;,
\end{equation*}
where $\xi_k = \lambda_\text{min}^{\nicefrac{1}{2}}(\overline{\bSigma}_0^{\nicefrac{1}{2}} \bSigma_k \overline{\bSigma}_0^{\nicefrac{1}{2}})$. Combining the last two displays we have
\begin{equation*}
    \Vert \widehat{\overline{\bSigma}}_1 - \overline{\bSigma}_1 \Vert \leq \frac{1}{n_d} \sum_{k=1}^{n_d} \frac{1}{\xi_k}\left\Vert \widehat{\overline{\bSigma}}_0^{\nicefrac{1}{2}}\widehat{\bSigma}_k \widehat{\overline{\bSigma}}_0^{\nicefrac{1}{2}} - \overline{\bSigma}_0^{\nicefrac{1}{2}}\bSigma_k \overline{\bSigma}_0^{\nicefrac{1}{2}} \right\Vert.
\end{equation*}
\begin{align*}
    &\Vert \widehat{\overline{\bSigma}}_1 - \overline{\bSigma}_1 \Vert \leq \frac{1}{n_d} \sum_{k=1}^{n_d} \frac{1}{\xi_k} \left[\Vert \widehat{\overline{\bSigma}}_0^{\nicefrac{1}{2}}\widehat{\bSigma}_k \widehat{\overline{\bSigma}}_0^{\nicefrac{1}{2}} - \overline{\bSigma}_0^{\nicefrac{1}{2}}\widehat{\bSigma}_k \widehat{\overline{\bSigma}}_0^{\nicefrac{1}{2}} \Vert + \Vert \overline{\bSigma}_0^{\nicefrac{1}{2}}\widehat{\bSigma}_k \widehat{\overline{\bSigma}}_0^{\nicefrac{1}{2}} - \overline{\bSigma}_0^{\nicefrac{1}{2}}\bSigma_k \overline{\bSigma}_0^{\nicefrac{1}{2}} \Vert \right] \\
    &\leq \frac{1}{n_d} \sum_{k=1}^{n_d} \frac{1}{\xi_k} \left[ \Vert \widehat{\bSigma}_k \Vert \Vert \widehat{\overline{\bSigma}}_0^{\nicefrac{1}{2}} \Vert \Vert \widehat{\overline{\bSigma}}_0^{\nicefrac{1}{2}} - \overline{\bSigma}_0^{\nicefrac{1}{2}} \Vert + \Vert \overline{\bSigma}_0^{\nicefrac{1}{2}} \Vert \Vert \widehat{\bSigma}_k \widehat{\overline{\bSigma}}_0^{\nicefrac{1}{2}} - \bSigma_k \overline{\bSigma}_0^{\nicefrac{1}{2}} \Vert \right] \\
    &\leq \frac{1}{n_d} \sum_{k=1}^{n_d} \frac{1}{\xi_k} \left[ \Vert \widehat{\bSigma}_k \Vert \Vert \widehat{\overline{\bSigma}}_0^{\nicefrac{1}{2}} \Vert \Vert \widehat{\overline{\bSigma}}_0^{\nicefrac{1}{2}} - \overline{\bSigma}_0^{\nicefrac{1}{2}} \Vert + \Vert \overline{\bSigma}_0^{\nicefrac{1}{2}} \Vert \left[ \Vert \widehat{\bSigma}_k \widehat{\overline{\bSigma}}_0^{\nicefrac{1}{2}} - \widehat{\bSigma}_k \overline{\bSigma}_0^{\nicefrac{1}{2}}\Vert + \Vert\widehat{\bSigma}_k \overline{\bSigma}_0^{\nicefrac{1}{2}} - \bSigma_k \overline{\bSigma}_0^{\nicefrac{1}{2}} \Vert \right] \right] \\
    &\leq \frac{1}{n_d} \sum_{k=1}^{n_d} \frac{1}{\xi_k} \left[ \Vert \widehat{\bSigma}_k \Vert \Vert \widehat{\overline{\bSigma}}_0^{\nicefrac{1}{2}} \Vert \Vert \widehat{\overline{\bSigma}}_0^{\nicefrac{1}{2}} - \overline{\bSigma}_0^{\nicefrac{1}{2}} \Vert + \Vert \overline{\bSigma}_0^{\nicefrac{1}{2}} \Vert \Vert \widehat{\bSigma}_k \Vert \Vert\widehat{\overline{\bSigma}}_0^{\nicefrac{1}{2}} - \overline{\bSigma}_0^{\nicefrac{1}{2}}\Vert + \Vert \overline{\bSigma}_0 \Vert \Vert\widehat{\bSigma}_k  - \bSigma_k \Vert \right] \\
    &\leq \frac{1}{n_d} \sum_{k=1}^{n_d} \frac{\Vert \widehat{\bSigma}_k \Vert}{\xi_k} \left( \Vert \widehat{\overline{\bSigma}}_0^{\nicefrac{1}{2}} \Vert + \Vert \overline{\bSigma}_0^{\nicefrac{1}{2}} \Vert \right) \Vert \widehat{\overline{\bSigma}}_0^{\nicefrac{1}{2}} - \overline{\bSigma}_0^{\nicefrac{1}{2}} \Vert + \Vert \overline{\bSigma}_0 \Vert \frac{1}{n_d} \sum_{k=1}^{n_d} \frac{\Vert\widehat{\bSigma}_k  - \bSigma_k \Vert}{\xi_k} \;.
\end{align*}
We apply again the Lemma 2.1 in \cite{SCHMITT1992215} and get
\begin{equation}
\Vert \widehat{\overline{\bSigma}}_0^{\nicefrac{1}{2}} - \overline{\bSigma}_0^{\nicefrac{1}{2}} \Vert \leq \frac{1}{\lambda_\text{min}^{\nicefrac{1}{2}}(\overline{\bSigma}_0)} \Vert \widehat{\overline{\bSigma}}_0 - \overline{\bSigma}_0 \Vert\;.
\end{equation}
That leads to
\begin{equation*}
    \Vert \widehat{\overline{\bSigma}}_1 - \overline{\bSigma}_1 \Vert \leq \frac{1}{n_d} \sum_{k=1}^{n_d} \frac{\Vert \widehat{\bSigma}_k \Vert}{\xi_k} \left( \Vert \widehat{\overline{\bSigma}}_0^{\nicefrac{1}{2}} \Vert + \Vert \overline{\bSigma}_0^{\nicefrac{1}{2}} \Vert \right) \frac{1}{\lambda_\text{min}^{\nicefrac{1}{2}}(\overline{\bSigma}_0)}\Vert \widehat{\overline{\bSigma}}_0 - \overline{\bSigma}_0 \Vert + \Vert \overline{\bSigma}_0 \Vert \frac{1}{n_d} \sum_{k=1}^{n_d} \frac{\Vert\widehat{\bSigma}_k  - \bSigma_k \Vert}{\xi_k}\;.
\end{equation*}
At this point, we want to express the bound only with $\bSigma_k$. For that we need to deal with $\xi_k$, $\overline{\bSigma}_0$ and $\widehat{\overline{\bSigma}}_0$. First, we have
\begin{equation*}
    \frac{1}{\xi_k} = \Vert \overline{\bSigma}_0^{-\nicefrac{1}{2}} \bSigma_k^{-1} \overline{\bSigma}_0^{-\nicefrac{1}{2}} \Vert^{\nicefrac{1}{2}} \leq \Vert \bSigma_k^{-\nicefrac{1}{2}} \Vert \Vert \overline{\bSigma}_0^{-\nicefrac{1}{2}} \Vert\;, \text{and} \; \frac{1}{\lambda_\text{min}^{\nicefrac{1}{2}}(\overline{\bSigma}_0)} = \Vert \overline{\bSigma}_0^{-\nicefrac{1}{2}} \Vert
\end{equation*}
leading to
\begin{align*}
    \Vert \widehat{\overline{\bSigma}}_1 - \overline{\bSigma}_1\Vert \leq &\left[\frac{1}{n_d} \sum_{k=1}^{n_d} \Vert \widehat{\bSigma}_k \Vert \Vert \bSigma_k^{-\nicefrac{1}{2}} \Vert \right] \Vert \overline{\bSigma}_0^{-1}\Vert \left( \Vert \widehat{\overline{\bSigma}}_0^{\nicefrac{1}{2}} \Vert + \Vert \overline{\bSigma}_0^{\nicefrac{1}{2}} \Vert \right)\Vert \widehat{\overline{\bSigma}}_0 - \overline{\bSigma}_0 \Vert \\
    &+ \Vert \overline{\bSigma}_0 \Vert \Vert \overline{\bSigma}_0^{-\nicefrac{1}{2}} \Vert \frac{1}{n_d} \sum_{k=1}^{n_d} \Vert \bSigma_k^{-\nicefrac{1}{2}} \Vert \Vert\widehat{\bSigma}_k  - \bSigma_k \Vert\;.
\end{align*}
Next, in view of \eqref{eq:sig0} we have
\begin{equation*}
    \Vert \overline{\bSigma}_0 \Vert \leq \sum_{k=1}^{n_d} \Vert \bSigma_k \Vert,  \quad \Vert (\overline{\bSigma}_0)^{-1} \Vert \leq \sum_{k=1}^{n_d} \Vert \bSigma_k^{-1} \Vert, \quad  \Vert \overline{\bSigma}_0^{\nicefrac{1}{2}} \Vert \leq \sum_{k=1}^{n_d} \Vert \bSigma_k^{\nicefrac{1}{2}} \Vert \quad\text{and}\quad  \Vert \overline{\bSigma}_0^{-\nicefrac{1}{2}} \Vert \leq \sum_{k=1}^{n_d} \Vert \bSigma_k^{-\nicefrac{1}{2}} \Vert.
\end{equation*}
% \begin{equation*}
%     \Vert \overline{\bSigma}_0^{-1} \Vert \leq \sum_{k=1}^{n_d} \Vert \bSigma_k^{-1} \Vert \;,
% \end{equation*}
% \begin{equation*}
%     \Vert \overline{\bSigma}_0^{\nicefrac{1}{2}} \Vert \leq \sum_{k=1}^{n_d} \Vert \bSigma_k^{\nicefrac{1}{2}} \Vert \;,
% \end{equation*}
% and 
% \begin{equation*}
%     \Vert \overline{\bSigma}_0^{-\nicefrac{1}{2}} \Vert \leq \sum_{k=1}^{n_d} \Vert \bSigma_k^{-\nicefrac{1}{2}} \Vert \;.
% \end{equation*}
Combining the last two displays, we get
\begin{align*}
    \Vert \widehat{\overline{\bSigma}}_1 - \overline{\bSigma}_1 \Vert \leq &\left[\frac{1}{n_d} \sum_{k=1}^{n_d} \Vert \widehat{\bSigma}_k \Vert \Vert \bSigma_k^{-\nicefrac{1}{2}} \Vert \right]\left[ \Vert \left(\frac{1}{n_d}\sum_{k=1}^{n_d}\bSigma_k \right)^{-1}\Vert\right] \\ 
    & \left[\frac{1}{n_d} \sum_{k=1}^{n_d} \left( \Vert \widehat{\bSigma}_k^{\nicefrac{1}{2}} \Vert + \Vert \bSigma_k^{\nicefrac{1}{2}} \Vert \right) \right]\left[\frac{1}{n_d} \sum_{k=1}^{n_d}\Vert \widehat{\bSigma}_k - \bSigma_k \Vert\right] \\ 
    &+ \left[\frac{1}{n_d} \sum_{k=1}^{n_d}\Vert \bSigma_k \Vert\right] \left[\Vert \left( \frac{1}{n_d}\sum_{k=1}^{n_d}\bSigma_k\right)^{\nicefrac{-1}{2}} \Vert \right]\frac{1}{n_d} \sum_{k=1}^{n_d} \Vert \bSigma_k^{-\nicefrac{1}{2}} \Vert \Vert\widehat{\bSigma}_k  - \bSigma_k \Vert
\end{align*}
\begin{align*}
    \Vert \widehat{\overline{\bSigma}}_1 - \overline{\bSigma}_1 \Vert \leq &\left[\frac{1}{n_d} \sum_{k=1}^{n_d} \Vert \widehat{\bSigma}_k \Vert \Vert \bSigma_k^{-\nicefrac{1}{2}} \Vert \right]\left[ \frac{1}{n_d} \sum_{k=1}^{n_d}\Vert\bSigma_k^{-1}\Vert\right] \\ 
    &\left[\frac{1}{n_d} \sum_{k=1}^{n_d} \left( \Vert \widehat{\bSigma}_k^{\nicefrac{1}{2}} \Vert + \Vert \bSigma_k^{\nicefrac{1}{2}} \Vert \right) \right]\left[\frac{1}{n_d} \sum_{k=1}^{n_d}\Vert \widehat{\bSigma}_k - \bSigma_k \Vert\right] \\ 
    &+ \left[\frac{1}{n_d} \sum_{k=1}^{n_d}\Vert \bSigma_k \Vert\right] \left[ \left( \frac{1}{n_d}\sum_{k=1}^{n_d}\Vert\bSigma_k^{-1}\Vert\right)^{\nicefrac{1}{2}}  \right]\frac{1}{n_d} \sum_{k=1}^{n_d} \Vert \bSigma_k^{-\nicefrac{1}{2}} \Vert \Vert\widehat{\bSigma}_k  - \bSigma_k \Vert.
\end{align*}
We now need to control the term $\Vert \widehat{\bSigma}_k \Vert$, to this end, we introduce the event
\begin{equation}
    \cE = \bigcap_{k=1}^{n_d}\left\{\Vert \widehat{\bSigma}_k - \bSigma_k \Vert \leq \frac{\Vert \bSigma_k \Vert}{2}\right\}\;.
\end{equation}
We have on $\cE$ that $\Vert \widehat{\bSigma}_k \Vert \leq \frac{3}{2}\Vert \bSigma_k \Vert
$, $\forall k\in [n_d]$ and consequently
% \begin{equation*}
%     \Vert \widehat{\bSigma}_k \Vert \leq \frac{3}{2}\Vert \bSigma_k \Vert
% ,\quad \forall k\in [n_d].
% \end{equation*}
%Hence we get on the event $\cE$
\begin{align*}
    \Vert \widehat{\overline{\bSigma}}_1 - \overline{\bSigma}_1 \Vert \lesssim &\left[\frac{1}{n_d} \sum_{k=1}^{n_d}  \Vert \bSigma_k  \Vert  \Vert \bSigma_k^{-\nicefrac{1}{2}} \Vert \right]\left[ \frac{1}{n_d} \sum_{k=1}^{n_d}\Vert\bSigma_k^{-1}\Vert\right] \left[\frac{1}{n_d} \sum_{k=1}^{n_d}  \Vert \bSigma_k^{\nicefrac{1}{2}}  \Vert  \right]  \left[\frac{1}{n_d} \sum_{k=1}^{n_d}\Vert \widehat{\bSigma}_k - \bSigma_k \Vert\right] \\
    &+ \left[\frac{1}{n_d} \sum_{k=1}^{n_d}\Vert \bSigma_k \Vert\right] \left[ \left( \frac{1}{n_d}\sum_{k=1}^{n_d}\Vert\bSigma_k^{-1}\Vert\right)^{\nicefrac{1}{2}}  \right]\frac{1}{n_d} \sum_{k=1}^{n_d} \Vert \bSigma_k^{-\nicefrac{1}{2}} \Vert \Vert\widehat{\bSigma}_k  - \bSigma_k \Vert \\
    % \lesssim & \frac{C}{n_d} \sum_{k=1}^{n_d}\Vert \widehat{\bSigma}_k - \bSigma_k \Vert 
    % +  \frac{C'}{n_d} \sum_{k=1}^{n_d} \Vert \bSigma_k^{-\nicefrac{1}{2}} \Vert \Vert\widehat{\bSigma}_k  - \bSigma_k \Vert  \\
    &\lesssim 
    \frac{1}{n_d}\sum_{k=1}^{n_d} \eta_k \Vert \widehat{\bSigma}_k - \bSigma_k \Vert,
\end{align*}
where 
\begin{multline*}
\eta_k = \left[\frac{1}{n_d} \sum_{l=1}^{n_d}  \Vert \bSigma_l  \Vert  \Vert \bSigma_l^{-\nicefrac{1}{2}} \Vert \right]\left[ \frac{1}{n_d} \sum_{l=1}^{n_d}\Vert\bSigma_l^{-1}\Vert\right] \left[\frac{1}{n_d} \sum_{l=1}^{n_d}  \Vert \bSigma_l^{\nicefrac{1}{2}}  \Vert  \right]\\ 
    + \left[\frac{1}{n_d} \sum_{l=1}^{n_d}\Vert \bSigma_l \Vert\right] \left[ \left( \frac{1}{n_d}\sum_{l=1}^{n_d}\Vert\bSigma_l^{-1}\Vert\right)^{\nicefrac{1}{2}}  \right] \Vert \bSigma_k^{-\nicefrac{1}{2}} \Vert.
\end{multline*}

From~\cite{flamary2020concentration}, and the bound for iteration of the barycenter, we have
\begin{align*}
    &\Vert \widehat{\bA} - \bA \Vert \lesssim \frac{\kappa(\bSigma_t)}{\lambda_\text{min}^{\nicefrac{1}{2}}(\bSigma_t^{\nicefrac{1}{2}} \overline{\bSigma} \bSigma_t^{\nicefrac{1}{2}})} \Vert \widehat{\overline{\bSigma}} - \overline{\bSigma} \Vert + \frac{\kappa(\overline{\bSigma}) \Vert \overline{\bSigma} \Vert \Vert \bSigma_t^{-1} \Vert}{\lambda_\text{min}^{\nicefrac{1}{2}}(\overline{\bSigma}^{\nicefrac{1}{2}} \bSigma_t\overline{\bSigma}^{\nicefrac{1}{2}})} \Vert \widehat{\bSigma}_t - \bSigma_t \Vert \\
    &\leq \kappa(\bSigma_t) \Vert\bSigma^{-\nicefrac{1}{2}}\Vert \Vert\overline{\bSigma}^{-\nicefrac{1}{2}}\Vert \left(\frac{1}{n_d}\sum_{k=1}^{n_d} \eta_k \Vert \widehat{\bSigma}_k - \bSigma_k \Vert \right) + \Vert\overline{\bSigma}\Vert^2 \Vert \Vert \overline{\bSigma}_0^{-1} \Vert \Vert \overline{\bSigma}_0^{-\nicefrac{1}{2}} \Vert\Vert \Vert\bSigma_t^{-1}\Vert \Vert \widehat{\bSigma}_t - \bSigma_t \Vert \\
    &\leq \kappa(\bSigma_t) \Vert\bSigma^{-\nicefrac{1}{2}}\Vert \left[\frac{1}{n_d}\sum_{k=1}^{n_d} \Vert\bSigma_k^{-\nicefrac{1}{2}}\Vert \right]\left(\frac{1}{n_d}\sum_{k=1}^{n_d} \eta_k \Vert \widehat{\bSigma}_k - \bSigma_k \Vert \right) \\ &+ \left[\frac{1}{n_d}\sum_{k=1}^{n_d} \Vert\bSigma_k\Vert \right]^2 \left[\frac{1}{n_d}\sum_{k=1}^{n_d} \Vert\bSigma_k^{-1}\Vert \right] \left[\frac{1}{n_d}\sum_{k=1}^{n_d} \Vert\bSigma_k^{-\nicefrac{1}{2}}\Vert \right] \Vert\bSigma_t^{-1}\Vert \Vert \widehat{\bSigma}_t - \bSigma_t \Vert
\end{align*}
With $c_k = \kappa(\bSigma) \Vert\bSigma^{-\nicefrac{1}{2}}\Vert \Vert\overline{\bSigma}^{-\nicefrac{1}{2}}\Vert \eta_k $ and $C =\Vert\overline{\bSigma}\Vert^2 \Vert\overline{\bSigma}^{-\nicefrac{3}{2}}\Vert \Vert\bSigma^{-1}\Vert $, we have
\begin{equation}
    \Vert \widehat{\bA} - \bA \Vert \leq \left(\frac{1}{n_d}\sum_{k=1}^{n_d} c_k \Vert \widehat{\bSigma}_k - \bSigma_k \Vert \right) + C\Vert \widehat{\bSigma}_t - \bSigma_t \Vert
    \label{eq:flam}
\end{equation}

\subsubsection{Upper-bound for covariance estimation using Welch method}
We now focus on the bound for $\Vert\widehat{\bSigma}  - \bSigma \Vert $. To do that, we use Corollary 1 from \cite{lamperski2023nonasymptotic} with Welch method case. First, we define some notation from \cite{lamperski2023nonasymptotic}.
We define the correlation and the continuous PSD by:
\begin{align*}
    % \bR[k] &= \bbE\left[ X[i+k]X[i]^\trans \right],\quad 
    \bQ(s) = \sum_{k=-\infty}^\infty e^{-j2\pi s k }\bR[k].
\end{align*}
We use the Welch method defined in Equation~\ref{eq:welch_stma} with a square window function, and we assume $n_o=f/2$, which is often the case \cite{lamperski2023nonasymptotic}. This assumption leads to $n_w=\frac{2n_\ell}{\filtersize}$:
% \begin{theorem}[4.2 from~\cite{lamperski2023nonasymptotic}]
%     \newline
%     \begin{enumerate}
%         \item If
%     \begin{equation*}
%         \frac{n_w}{1 + 2\frac{n_f}{n_o}} \geq \log\left(5 f^2 \left(\frac{4 \times 10^{2n_c}}{\delta}\right)^{32}\right) \max\left\{4\frac{\left\Vert\bQ\right\Vert_\infty^2}{\varepsilon^2}, 2\frac{\left\Vert \bQ\right\Vert_\infty}{\varepsilon}\right\}
%     \end{equation*}
%     then for $s \in [-\frac{1}{2},\frac{1}{2}]$
%     \begin{equation*}
%         \Pr\left(\sup_{s\in [-\frac{1}{2},\frac{1}{2}]} \left\Vert \bP(s) - \bbE\left[\bP(s)\right]\right\Vert_2 > \varepsilon \right) \leq \delta \;.
%     \end{equation*}
%         \item If $f \geq \widehat{f}(\epsilon)$ and for all $\vert k \vert < \widehat{f}(\epsilon)$ we have $\sum_{i=\vert k\vert}^{f-1}$
%     \end{enumerate}
% \end{theorem}

\begin{corollary}[4.2 from~\cite{lamperski2023nonasymptotic}]

    \begin{enumerate}
        \item If $\filtersize < n_\ell$ and 
    \begin{equation*}
        \frac{ 2\frac{n_\ell}{f}}{5} \geq \log\left(5 f^2 \left(\frac{4 \times 10^{2n_c}}{\delta}\right)^{32}\right) \max\left\{4\frac{\left\Vert\bQ\right\Vert_\infty^2}{\varepsilon^2}, 2\frac{\left\Vert \bQ\right\Vert_\infty}{\varepsilon}\right\}
    \end{equation*}
    then for all $\delta \in (0,1)$, the following bounds hold with probability at least $1 - \delta$:
    \begin{multline*}
        \left\Vert \widehat{\bQ}(s) - \bbE\left[\widehat{\bQ}(s)\right]\right\Vert_\infty \leq 2  \left\Vert\bQ\right\Vert_\infty \\
        \max\left\{ \frac{5}{2\frac{n_\ell}{f}} \log \left(5 f^2 \left(\frac{4 \times 10^{2n_c}}{\delta}\right)^{32}\right), \sqrt{\frac{5}{2\frac{n_\ell}{f}} \log \left(5 f^2 \left(\frac{4 \times 10^{2n_c}}{\delta}\right)^{32}\right)}\right\}  \;.
    \end{multline*}
        \item Assume that there are constants $\delta > 0$
 and $\rho \in [0, 1)$ such that $\Vert \bR[k]\Vert_2 \leq \gamma \rho^{\vert k \vert}$ for all $k \in \bbZ$
 % and assume that $\frac{n_\ell- \vert k \vert}{n_\ell}  = 0$ for all $\vert k \vert \qeg f$, where $f \leq n_\ell$. 
 Then
 \begin{equation*}
     \left\Vert \bQ - \bbE \left[\widehat{\bQ}\right] \right\Vert_\infty \leq 2\gamma \sum_{i=0}^{f-1}\frac{ i}{n_\ell} \rho^i + \frac{2\gamma \rho^f}{1-\rho}
 \end{equation*}
    \end{enumerate}
\label{cor:lamperski}
\end{corollary}

We have then
\begin{equation*}
    \left\Vert \widehat{\bSigma} - \bSigma \right\Vert =\max_i \left\Vert \widehat{\bQ}_{i\frac{n_\ell}{f}+1} - \bQ_{i\frac{n_\ell}{f}+1} \right\Vert \leq \sup_{s\in [0, 1]}\left\Vert \widehat{\bQ}(s)- \bQ(s) \right\Vert \leq \left\Vert \widehat{\bQ} - \bbE\left[\widehat{\bQ}\right]\right\Vert_\infty + \left\Vert \bQ - \bbE \left[\widehat{\bQ}\right] \right\Vert_\infty \;.
\end{equation*}
Combninb the previous display with Corollary \ref{cor:lamperski}, we get with probability at least $1 - \delta$
\begin{multline}
    \left\Vert \widehat{\bSigma} - \bSigma \right\Vert \leq \delta 2\sum_{i=0}^{f-1}\frac{i}{n_\ell} \rho^i + \frac{2\delta \rho^f}{1-\rho} \\ + 2  \left\Vert\bQ\right\Vert_\infty 
        \max\left\{ \frac{5}{2\frac{n_\ell}{f}} \log \left(5 f^2 \left(\frac{4 \times 10^{2n_c}}{\delta}\right)^{32}\right), \sqrt{\frac{5}{2\frac{n_\ell}{f}}\log \left(5 f^2 \left(\frac{4 \times 10^{2n_c}}{\delta}\right)^{32}\right)}\right\} 
        \label{eq:lamp}
\end{multline}
For the following we note $g = \frac{5}{2\frac{n_\ell}{f}} \log \left(5 f^2 \left(\frac{4 \times 10^{2n_c}}{\delta}\right)^{32}\right)$. With \ref{eq:flam} and \ref{eq:lamp} we get

\begin{align*}
     \Vert \widehat{\bA} - \bA \Vert \lesssim& \frac{1}{n_d}\sum_{k=1}^{n_d} c_k \Vert \widehat{\bSigma}_k - \bSigma_k \Vert  + C\Vert \widehat{\bSigma}_t - \bSigma_t \Vert \\
     \lesssim&  \frac{1}{n_d}\sum_{k=1}^{n_d} c_k \left( \delta 2\sum_{i=0}^{f-1}\frac{i}{n_\ell} \rho^i + \frac{2\delta \rho^f}{1-\rho}  + 2  \left\Vert\bQ_k\right\Vert_\infty 
        \max\left\{ g, \sqrt{g}\right\} \right) \\  &+ C\left( \delta 2\sum_{i=0}^{f-1}\frac{i}{n_\ell} \rho^i + \frac{2\delta \rho^f}{1-\rho}  + 2  \left\Vert\bQ_t\right\Vert_\infty 
        \max\left\{ g, \sqrt{g}\right\} \right) \\ 
    \lesssim&  \left(C + \frac{1}{n_d}\sum_{k=1}^{n_d} c_k \right)\left( \delta 2\sum_{i=0}^{f-1}\frac{i}{n_\ell} \rho^i + \frac{2\delta \rho^f}{1-\rho}  + 2  \widetilde{\bQ} 
        \max\left\{ g, \sqrt{g}\right\} \right) 
\end{align*}
where $\widetilde{\bQ} = \max_{d \in [1, \dots, n_d, t]} \Vert \bQ_d \Vert_\infty$

\subsection{Proof of \autoref{the:concentration_spatio}: \smethod{} concentration bound}

This bound starts the same way as the one for \stmethod{}. Using \cite{flamary2020concentration} we have 
 we have that
\begin{equation}
    \Vert \widehat{\bA} - \bA \Vert \lesssim 
    \left(\frac{1}{n_d}\sum_{k=1}^{n_d} c_k \Vert \widehat{\bSigma}_k - \bSigma_k \Vert \right) + C\Vert \widehat{\bSigma}_t - \bSigma_t \Vert\;.
    \label{eq:flam2}
\end{equation}
Here, the covariance matrices are not computed using Welch, the Lamperski Theorem does not apply.

\subsubsection{Bound of covariance matrices estimation}
We now apply Theorem 2 in \cite{koltchinskii2015asymptotics} to bound the estimation of covariance matrices. We obtain for any $\delta>0$, with probability at least $1 - \delta$,
\begin{equation*}
    \Vert  \widehat{\bSigma} - \bSigma \Vert \lesssim \Vert \bSigma \Vert \max\left(\sqrt{\frac{\br(\bSigma)}{n_\ell} }, \frac{\br(\bSigma)}{n_\ell}, \sqrt{\frac{-\ln{\delta}}{n_\ell} }, \frac{-\ln{\delta}}{n_\ell}\right)\;,
\end{equation*}

with $\br(\bSigma) = \frac{\text{tr}(\bSigma)}{\lambda_{\max}(\bSigma)}$.

Replacing the corresponding term in Equation \ref{eq:flam2}, we get
\begin{align*}
     \Vert \widehat{\bA} - \bA \Vert \lesssim& \frac{1}{n_d}\sum_{k=1}^{n_d} c_k \Vert \widehat{\bSigma}_k - \bSigma_k \Vert  + C\Vert \widehat{\bSigma}_t - \bSigma_t \Vert \\
     \lesssim&  \frac{1}{n_d}\sum_{k=1}^{n_d} c_k\Vert \bSigma_k \Vert \max\left(\sqrt{\frac{\br(\bSigma_k)}{n_\ell} }, \frac{\br(\bSigma_k)}{n_\ell}, \sqrt{\frac{-\ln{\delta}}{n_\ell} }, \frac{-\ln{\delta}}{n_c}\right) \\  &+ C\Vert \bSigma_t \Vert \max\left(\sqrt{\frac{\br(\bSigma_t)}{n_\ell} }, \frac{\br(\bSigma_t)}{n_\ell}, \sqrt{\frac{-\ln{\delta}}{n_\ell} }, \frac{-\ln{\delta}}{n_\ell}\right)\\ 
    % \lesssim&  \left(C'' + \frac{1}{n_d}\sum_{k=1}^{n_d} C_k' \right)\left( \delta 2\sum_{i=0}^{f-1}\frac{i}{n_\ell} \rho^i + \frac{2\delta \rho^f}{1-\rho}  + 2  \widetilde{\bQ} 
    %     \max\left\{ c, \sqrt{c}\right\} \right) \\
    %     \lesssim&  C'''\left( \delta 2\sum_{i=0}^{f-1}\frac{i}{n_\ell} \rho^i + \frac{2\delta \rho^f}{1-\rho}  + 2  \widetilde{\bQ}
    %     \max\left\{ c, \sqrt{c}\right\} \right)\;,
\end{align*}

\subsection{Proof of \autoref{the:concentration_temp}: \tmethod{} concentration bound}

We now focus on bounding the estimation error of the $\filtersize$-Monge mapping in the pure temporal case.
This mapping is given in Proposition~\ref{prop:temp_map}.
We have $\bA = \bF \diag \left(\bA_1, \dots, \bA_{n_c}\right) \bF\hermconj$. So the concentration bound is
\begin{equation*}
    \Vert \widehat{\bA} - \bA \Vert = \max_c \Vert \widehat{\bA}_c - \bA_c \Vert = \max_c \left\Vert \widehat{\overline\bp}_c^{\odot\frac{1}{2}} \odot \widehat{\bp}_{c, t}^{\odot-\frac{1}{2}} - \overline\bp_c^{\odot\frac{1}{2}} \odot \bp_{c, t}^{\odot-\frac{1}{2}} \right\Vert_\infty \;.
\end{equation*}

Thus, for every $c \in \intset{n_c}$, 
using the triangle inequality and the sub-multiplicativity of the $\infty$-norm, we get that
\begin{align*}
     \Vert \widehat{\bA}_c - \bA_c \Vert &\leq \Vert \widehat{\overline\bp}_c^{\odot\frac{1}{2}} \odot \widehat{\bp}_{c, t}^{\odot-\frac{1}{2}} - \overline\bp_c^{\odot\frac{1}{2}} \odot \bp_{c, t}^{\odot-\frac{1}{2}} \Vert_\infty \\
    &= \Vert \widehat{\overline\bp}_c^{\odot\frac{1}{2}} \odot \widehat{\bp}_{c, t}^{\odot-\frac{1}{2}} - \overline\bp_c^{\odot\frac{1}{2}} \odot \widehat{\bp}_{c, t}^{\odot-\frac{1}{2}} + \overline\bp_c^{\odot\frac{1}{2}} \odot \widehat{\bp}_{c, t}^{\odot-\frac{1}{2}} - \overline\bp_c^{\odot\frac{1}{2}} \odot \bp_{c, t}^{\odot-\frac{1}{2}} \Vert_\infty \\
    &\leq \Vert \widehat{\overline\bp}_c^{\odot\frac{1}{2}} \odot \widehat{\bp}_{c, t}^{\odot-\frac{1}{2}} - \overline\bp_c^{\odot\frac{1}{2}} \odot \widehat{\bp}_{c, t}^{\odot-\frac{1}{2}} \Vert_\infty + \Vert \overline\bp_c^{\odot\frac{1}{2}} \odot \widehat{\bp}_{c, t}^{\odot-\frac{1}{2}} - \overline\bp_c^{\odot\frac{1}{2}} \odot \bp_{c, t}^{\odot-\frac{1}{2}} \Vert_\infty \\
    &\leq \Vert \widehat{\bp}_{c, t}^{\odot-\frac{1}{2}} \Vert_\infty \Vert \widehat{\overline\bp}_c^{\odot\frac{1}{2}} - \overline\bp_c^{\odot\frac{1}{2}} \Vert_\infty + \Vert \overline\bp_c^{\odot\frac{1}{2}} \Vert_\infty \Vert \widehat{\bp}_{c, t}^{\odot-\frac{1}{2}} - \bp_{c, t}^{\odot-\frac{1}{2}} \Vert_\infty \;.
\end{align*}
We now need to deal with two terms $\Vert \widehat{\overline\bp}_c^{\odot\frac{1}{2}} - \overline\bp_c^{\odot\frac{1}{2}} \Vert_\infty $ and $\Vert \widehat{\bp}_{c, t}^{\odot-\frac{1}{2}} - \bp_{c, t}^{\odot-\frac{1}{2}} \Vert_\infty$. We focus on the first term. The barycenter is 
\begin{equation*}
    \overline{\bp}_c^{\odot\frac{1}{2}} = \frac{1}{n_d} \sum_{k=1}^{n_d} \bp_{c, k}^{\odot\frac{1}{2}}\;.
\end{equation*}
Thus, we get an upper-bound of the barycenter estimation error in terms of source PSD estimation errors, 
\begin{align*}
    \Vert \widehat{\overline\bp}_c^{\odot\frac{1}{2}} - \overline\bp_c^{\odot\frac{1}{2}} \Vert_\infty &\leq \frac{1}{n_d} \sum_{k=1}^{n_d} \Vert \widehat{\bp}_{c,k}^{\odot\frac{1}{2}} - \bp_{c,k}^{\odot\frac{1}{2}} \Vert_\infty \leq \frac{1}{n_d} \sum_{k=1}^{n_d} \Vert \diag{\widehat{\bp}_{c,k}}^{\frac{1}{2}} - \diag{\bp_{c,k}}^{\frac{1}{2}} \Vert
    \;.
\end{align*}
We apply Lemma 2.1 in \cite{SCHMITT1992215} to get
\begin{align}
    \Vert \widehat{\overline\bp}_c^{\odot\frac{1}{2}} - \overline\bp_c^{\odot\frac{1}{2}} \Vert_\infty &\leq \frac{1}{n_d} \sum^{n_d}_{k=1} \frac{1}{\Vert \bp_{c,k}^{\odot\frac{1}{2}}  \Vert_\infty} \Vert \diag{\widehat{\bp}_{c,k}} - \diag{\bp_{c,k}} \Vert \;.
\end{align}
We now focus on second term $\Vert \widehat{\bp}_{c, t}^{\odot-\frac{1}{2}} - \bp_{c, t}^{\odot-\frac{1}{2}} \Vert_\infty$.
\begin{align*}
   \Vert \widehat{\bp}_{c, t}^{\odot-\frac{1}{2}} - \bp_{c, t}^{\odot-\frac{1}{2}} \Vert_\infty &= \max_{j\in \intset{F}} \left\vert \frac{1}{(\widehat{\bp}_{c, t})_j^\frac{1}{2}} - \frac{1}{(\bp_{c, t})_j^\frac{1}{2}} \right\vert = \max_{j\in \intset{F}} \left\vert \frac{(\widehat{\bp}_{c, t})_j^\frac{1}{2} - (\bp_{c, t})_j^\frac{1}{2}}{(\widehat{\bp}_{c, t})_j^\frac{1}{2} (\bp_{c, t})_j^\frac{1}{2}} \right\vert \\
   & \leq \Vert \widehat{\bp}_{c, t}^{\odot-\frac{1}{2}} \odot \bp_{c, t}^{\odot-\frac{1}{2}} \Vert_\infty \max_{j\in \intset{F}} \left\vert(\widehat{\bp}_{c, t})_j^\frac{1}{2} - (\bp_{c, t})_j^\frac{1}{2} \right\vert \\
   &\leq \Vert \widehat{\bp}_{c, t}^{\odot-\frac{1}{2}} \Vert_\infty \Vert \bp_{c, t}^{\odot-\frac{1}{2}} \Vert_\infty \left\Vert \diag\widehat{\bp}_{c, t}^\frac{1}{2}-  \diag\bp_{c, t}^\frac{1}{2}\right\Vert.
\end{align*}

We apply Lemma 2.1 in \cite{SCHMITT1992215} to get
\begin{align}
    \Vert \widehat{\bp}_{c, t}^{\odot-\frac{1}{2}} - \bp_{c, t}^{\odot-\frac{1}{2}} \Vert_\infty &\leq\Vert \widehat{\bp}_{c, t}^{\odot-\frac{1}{2}} \Vert_\infty \Vert \bp_{c, t}^{\odot-1} \Vert_\infty \left\Vert \diag\widehat{\bp}_{c, t}-  \diag\bp_{c, t}\right\Vert\;.
\end{align}

Combining the two bounds, we get
\begin{multline*}
     \Vert \widehat{\bA}_c - \bA_c \Vert 
     \leq \Vert \widehat{\bp}_{c, t}^{\odot-\frac{1}{2}} \Vert_\infty \frac{1}{n_d} \sum^{n_d}_{k=1} \frac{1}{\Vert \bp_{c,k}^{\odot\frac{1}{2}}  \Vert_\infty} \Vert \diag{\widehat{\bp}_{c,k}} - \diag{\bp_{c,k}} \Vert \\+ \Vert \overline\bp_c^{\odot\frac{1}{2}} \Vert_\infty \Vert \Vert \widehat{\bp}_{c, t}^{\odot-\frac{1}{2}} \Vert_\infty \Vert \bp_{c, t}^{\odot-1} \Vert_\infty \left\Vert \diag\widehat{\bp}_{c, t}-  \diag\bp_{c, t}\right\Vert\;.
\end{multline*}

We now need to control the term $\Vert \widehat{\bp}_{c, t}^{\odot-\frac{1}{2}} \Vert_\infty$, to this end, we introduce the event
\begin{equation}
    \cE_2 = \bigcap_{c=1}^{n_c}\left\{\Vert 
    \widehat{\bp}_{c, t}^{\odot-\frac{1}{2}} - \bp_{c, t}^{\odot-\frac{1}
    {2}}\Vert_\infty \leq \frac{\Vert \bp_{c, t}^{\odot-\frac{1}{2}}\Vert}{2}\right\}\;.
\end{equation}
We have on $\cE_2$ that
\begin{equation*}
    \Vert \widehat{\bp}_{c, t}^{\odot-\frac{1}{2}} \Vert_\infty \leq \frac{3}{2}\Vert \bp_{c, t}^{\odot-\frac{1}{2}} \Vert \quad \text{and}\quad \Vert \overline\bp_c^{\odot\frac{1}{2}} \Vert_\infty \leq \frac{1}{n_d} \sum^{n_d}_{k=1} \Vert \bp_{c,k}^{\odot\frac{1}{2}} \Vert_\infty.
\end{equation*}
% And we have also
% \begin{equation*}
%     \Vert \overline\bp_c^{\odot\frac{1}{2}} \Vert_\infty \leq \frac{1}{n_d} \sum^{n_d}_{k=1} \Vert \bp_{c,k}^{\odot\frac{1}{2}} \Vert_\infty
% \end{equation*}
We then get
\begin{multline*}
     \Vert \widehat{\bA}_c - \bA_c \Vert 
     \lesssim \Vert \bp_{c,t}^{\odot\frac{1}{2}} \Vert_\infty \frac{1}{n_d} \sum^{n_d}_{k=1} \frac{1}{\Vert \bp_{c,k}^{\odot\frac{1}{2}}  \Vert_\infty} \Vert \diag{\widehat{\bp}_{c,k}} - \diag{\bp_{c,k}} \Vert \\+ \left[\frac{1}{n_d} \sum^{n_d}_{k=1} \Vert \bp_{c,k}^{\odot\frac{1}{2}}\Vert \Vert_\infty\right] \Vert\Vert \bp_{c,t}^{\odot\frac{1}{2}} \Vert_\infty\Vert \bp_{c, t}^{\odot-1} \Vert_\infty \left\Vert \diag\widehat{\bp}_{c, t}-  \diag\bp_{c, t}\right\Vert \;.
\end{multline*}

\subsubsection[Concentration bound]{Concentration bound using \cite{lamperski2023nonasymptotic}}
We know that $\Vert \bSigma_c \Vert = \Vert \diag \bp_{c,k} \Vert$. Using the last equation, we can bound the last term with Corrallary \ref{cor:lamperski} using the special case of $n_c=1$
\begin{multline*}
     \Vert \widehat{\bA}_c - \bA_c \Vert 
     \lesssim \Vert \bp_{c,t}^{\odot\frac{1}{2}} \Vert_\infty \frac{1}{n_d} \sum^{n_d}_{k=1} \frac{1}{\Vert \bp_{c,k}^{\odot\frac{1}{2}}  \Vert_\infty} \left( \delta 2\sum_{i=0}^{f-1}\frac{i}{n_\ell} \rho^i + \frac{2\delta \rho^f}{1-\rho}  + 2  \left\Vert\bp_{c,k}\right\Vert_\infty 
        \max\left\{ g, \sqrt{g}\right\} \right) \\+ \left[\frac{1}{n_d} \sum^{n_d}_{k=1} \Vert \bp_{c,k}^{\odot\frac{1}{2}} \Vert_\infty\right] \Vert \bp_{c,k}^{\odot\frac{1}{2}} \Vert_\infty\Vert \bp_{c, t}^{\odot-1} \Vert_\infty \left( \delta 2\sum_{i=0}^{f-1}\frac{i}{n_\ell} \rho^i + \frac{2\delta \rho^f}{1-\rho}  + 2  \left\Vert\bp_{c,t}\right\Vert_\infty 
        \max\left\{ g, \sqrt{g}\right\} \right)\;.
\end{multline*}
% \begin{align*}
%      \Vert \widehat{\bA} - \bA \Vert \lesssim& \frac{1}{n_d}\sum_{k=1}^{n_d} C_k' \Vert \widehat{\bSigma}_k - \bSigma_k \Vert  + C''\Vert \widehat{\bSigma}_t - \bSigma_t \Vert \\
%      \lesssim&  \frac{1}{n_d}\sum_{k=1}^{n_d} C_k' \left( \delta 2\sum_{i=0}^{f-1}\frac{i}{n_\ell} \rho^i + \frac{2\delta \rho^f}{1-\rho}  + 2  \left\Vert\bQ_k\right\Vert_\infty 
%         \max\left\{ c, \sqrt{c}\right\} \right) \\  &+ C''\left( \delta 2\sum_{i=0}^{f-1}\frac{i}{n_\ell} \rho^i + \frac{2\delta \rho^f}{1-\rho}  + 2  \left\Vert\bQ_t\right\Vert_\infty 
%         \max\left\{ c, \sqrt{c}\right\} \right) \\ 
%     \lesssim&  \left(C'' + \frac{1}{n_d}\sum_{k=1}^{n_d} C_k' \right)\left( \delta 2\sum_{i=0}^{f-1}\frac{i}{n_\ell} \rho^i + \frac{2\delta \rho^f}{1-\rho}  + 2  \widetilde{\bQ} 
%         \max\left\{ c, \sqrt{c}\right\} \right) 
% \end{align*}
where $g = \frac{5}{2\frac{n_\ell}{f}} \log \left(5 f^2 \left(\frac{4 \times 10^{2}}{\delta}\right)^{32}\right)$
Let be $c_{c, k}' = \Vert \bp_{c,k}^{\odot\frac{1}{2}} \Vert_\infty \frac{1}{\Vert \bp_{c,k}^{\odot\frac{1}{2}}  \Vert_\infty}  $ $C_c' = \left[\frac{1}{n_d} \sum^{n_d}_{k=1} \Vert \bp_{c,k}^{\odot\frac{1}{2}}\Vert \Vert_\infty\right] \Vert\Vert \bp_{c,k}^{\odot\frac{1}{2}} \Vert_\infty\Vert \bp_{c, t}^{\odot-1} \Vert_\infty$  and  $\widetilde{\bp}_c = \max_{k \in [1, \dots, n_d, t]} \Vert \bp_{c, k} \Vert_\infty$. Then, with $\delta > 0$ we have with probability at least $1 - \delta$

\begin{multline*}
     \Vert \widehat{\bA} - \bA \Vert 
     \lesssim \max_c\left( C_c' + \sum_{k=1}^{n_d} c_{c, k}' \right)\left( \delta 2\sum_{i=0}^{f-1}\frac{i}{n_\ell} \rho^i + \frac{2\delta \rho^f}{1-\rho}  \right. \\+  \left.2  \widetilde{\bp}_c 
        \max\left\{ \frac{5}{2\frac{n_\ell}{f}} \log \left(5 f^2 \left(\frac{4 \times 10^{2}}{\delta}\right)^{32}\right), \sqrt{\frac{5}{2\frac{n_\ell}{f}} \log \left(5 f^2 \left(\frac{4 \times 10^{2}}{\delta}\right)^{32}\right)}\right\} \right) \;.
\end{multline*}

% \subsection{Interpretation}
% \begin{figure}[t]
%     \begin{subfigure}{0.33\textwidth}
%         \centering
%         \includegraphics[width=\linewidth]{images/psd_reg_10-3.pdf}
%         \caption{$\lambda=0.001$}
%     \end{subfigure}
%     \begin{subfigure}{0.33\textwidth}
%         \centering
%         \includegraphics[width=\linewidth]{images/psd_reg_0_5.pdf}
%         \caption{$\lambda=0.5$}
%     \end{subfigure}
%     \begin{subfigure}{0.33\textwidth}
%         \centering
%         \includegraphics[width=\linewidth]{images/psd_reg_1.pdf}
%         \caption{$\lambda=1$}
%     \end{subfigure}
%     \label{fig:reg}
% \end{figure}
% \subsection{Interpretation}
% \begin{figure}[t]
%     \begin{minipage}{0.45\linewidth}
%     \centering
%     \includegraphics[width=\linewidth]{images/electrodes.pdf}
%     \end{minipage}
%     \hfill
%     \begin{minipage}{0.45\linewidth}
%     \centering
%     \includegraphics[width=\linewidth]{images/H_MASS.pdf}
%     \end{minipage}
%     \caption{Caption}
%     \label{fig:enter-label}
% \end{figure}

% \bibliographystyle{utphys}
% \bibliography{biblio}

\end{document}